\PassOptionsToPackage{table}{xcolor}
\documentclass[]{fairmeta}

\usepackage{amsmath}
\usepackage{amssymb}
\usepackage{amsthm}
\usepackage{wrapfig}
\usepackage{enumitem}
\usepackage{url}
\usepackage{algorithm}
\usepackage{algpseudocode}
\tcbuselibrary{breakable,skins}
\microtypesetup{expansion=false}
\newcommand{\MetaLayout}{}

\DeclareRobustCommand{\frozen}{\hspace{0.18em}\textcolor{blue!70!black}{\ensuremath{\ast}}\hspace{0.22em}}
\DeclareRobustCommand{\trainable}{\hspace{0.18em}\smash{\raisebox{-0.42ex}{\includegraphics[height=1.95ex]{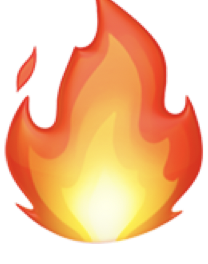}}}\hspace{0.22em}}

\definecolor{paperskyblue}{HTML}{559BC8}
\definecolor{citationblue}{HTML}{2980B9}
\newtheorem{theorem}{Theorem}

\newtheorem{appendixproposition}{Proposition}[section]
\newtcolorbox{thmbox}{
  colback=paperskyblue!7,
  colframe=citationblue!85!black,
  boxrule=1.0pt,
  arc=2pt,
  left=7pt,
  right=7pt,
  top=2.4pt,
  bottom=2.4pt,
  before skip=3.6pt,
  after skip=3.6pt
}

\definecolor{beliefbg}{HTML}{FDFBF0}      
\definecolor{progressbg}{HTML}{FBE3EF}    
\definecolor{experiencebg}{HTML}{E2F3E2}  

\newcommand{\statebox}[2]{%
  \begingroup\setlength{\fboxsep}{1.2pt}\colorbox{#1}{#2}\endgroup%
}

\newcommand{\Bstate}{\statebox{beliefbg}{\ensuremath{B_t}}}
\newcommand{\Pstate}{\statebox{progressbg}{\ensuremath{P_t}}}
\newcommand{\Estate}{\statebox{experiencebg}{\ensuremath{E_t}}}

\newcommand{\acttrack}{\statebox{beliefbg}{track}}
\newcommand{\actcommit}{\statebox{progressbg}{commit}}
\newcommand{\actrecall}{\statebox{experiencebg}{recall}}
\newcommand{\actnote}{\statebox{experiencebg}{note}}

\definecolor{darkred}{RGB}{120, 20, 20}
\newcommand{\deltagain}[1]{\textcolor{darkred}{#1}}
\newcommand{\oursdelta}[1]{\textcolor{darkred}{\textbf{#1}}}
\newcommand{\methodname}{\textsc{EvoHarness-RL}}

\title{\methodname{}: Learning Runtime Harness Coordination for Self-Evolving Agents}
\renewcommand{\titlelist}{{\LARGE\sffamily\bfseries\setstretch{1.0}%
  \methodname{}: Learning Runtime Harness Coordination\\for Self-Evolving Agents\par}}

\author[1]{Xuying Ning}
\author[2]{Dongqi Fu}
\author[1]{Tianxin Wei}
\author[1]{Yuanchen Bei}
\author[1]{Xiyuan Yang}
\author[3]{Wujiang Xu}
\author[4]{Yueqi Song}
\author[1]{\\Bingxuan Li}
\author[1]{Zihao Li}
\author[2]{Hanqing Zeng}
\author[2]{Xiang Shen}
\author[2]{Yajuan Wang}
\author[2]{Yifan Wu}
\author[2]{Qifan Wang}
\author[2]{\\Jiayi Liu}
\author[2]{Hong Li}
\author[2]{Yinglong Xia}
\author[2]{Xiangjun Fan}
\author[1]{Hanghang Tong}
\author[1]{Jingrui He}

\affiliation[1]{University of Illinois Urbana-Champaign}
\affiliation[2]{Meta AI}
\affiliation[3]{Rutgers University}
\affiliation[4]{Carnegie Mellon University}
\renewcommand{\affiliationlist}{{\normalsize $^{1}$University of Illinois Urbana-Champaign, $^{2}$Meta AI, $^{3}$Rutgers University,\\$^{4}$Carnegie Mellon University}}

\abstract{
Long-horizon LLM agents increasingly rely on external execution support to
maintain state, track progress, recover from failures, and reuse experience
across extended interactions. Yet existing harnesses and their use are often
tailored to environments and controlled through prompts, heuristics,
or system-specific rules, making agent and harness coordination difficult to
jointly optimize. We introduce \textbf{\methodname{}}, a unified framework that separates
environment-specific harness implementations from a shared policy-facing
interface. \methodname{} organizes external support into a Belief, Progress,
and Experience (BPE) workspace and exposes four compact harness actions for
accessing and updating this state. We first instantiate BPE as an inference-time
scaffold and then make harness coordination learnable through supervised
initialization followed by cost-aware GRPO. Across heterogeneous long-horizon
tasks, \textsc{EvoHarness-Base} improves the average success rate of frontier models
by 10.0 percentage points, while \methodname{} outperforms the strongest
open-source baseline by 8.5 percentage points, together with higher
RL rollout efficiency and stronger generalization to unseen tasks. Our analyses
show that training gradually shifts agents from frequent scaffold use toward
selective, environment-dependent harness access as the policy becomes more
capable, while the external workspace continues to evolve and refine itself
to better support task execution and generalization. Together, these results
show that long-horizon agents benefit not only from external scaffolding
itself, but also from learning how and when to coordinate with external
support as part of a cost-aware policy.
}

\date{\today}
\metadata[Project page]{\url{https://evoharnessrl.github.io}}
\metadata[arXiv]{\url{https://arxiv.org/abs/2608.05446}}

\begin{document}

\maketitle

\vspace{-5mm}
\section{Introduction}

LLM-based agents are increasingly deployed in long-horizon interactive settings,
where reliable task execution requires reasoning and acting over extended interactions.
In embodied interaction, web navigation, software engineering, and workflow
automation~\citep{alfworld,hong2026embodied,sweagent,zhou2024webarena}, agents
need to maintain environment state, track task progress, recover from failed
actions, and reuse knowledge from prior experience. These demands have motivated
a growing ecosystem of external execution support, which we broadly refer to as
the agent \emph{harness}. Such support can take many forms, ranging from memories
and state trackers to tools, verifiers, execution logs, and other mechanisms that
augment the agent during interaction~\citep{suzgun2026dynamic,toolformer,ning2026code,wei2025evo,harness1}.

Despite this common role, harnesses are typically tailored to individual
environments, leading to increasingly specialized and complex designs that are
difficult to transfer across environments or agent models~\citep{anthropic2025longrunning,lee2026meta,lou2026autoharness,harness1,harnessx}.
For example, embodied agents rely on world-state tracking, while web agents
rely on page-state tracking and element grounding~\citep{alfworld,zhou2024webarena}. Beyond what
information is maintained, how the agent interacts with the harness is also
largely manually designed, often through prompts, heuristics, or system-specific
rules rather than jointly learned with the agent. Yet even with the same harness, different models and tasks need different levels
of harness support, and these needs change as the policy improves. Learning when and
how to use the harness can therefore improve task execution while avoiding
unnecessary interaction costs.
However, the current fragmented harness designs and interaction rules make such
coordination difficult to formulate as a common learning problem.
This raises a central question: \textit{How can heterogeneous, task-specific
harness designs and interaction rules be unified into a learnable policy
for long-horizon execution?}

\begin{figure}[t]
\centering
\includegraphics[width=\textwidth]{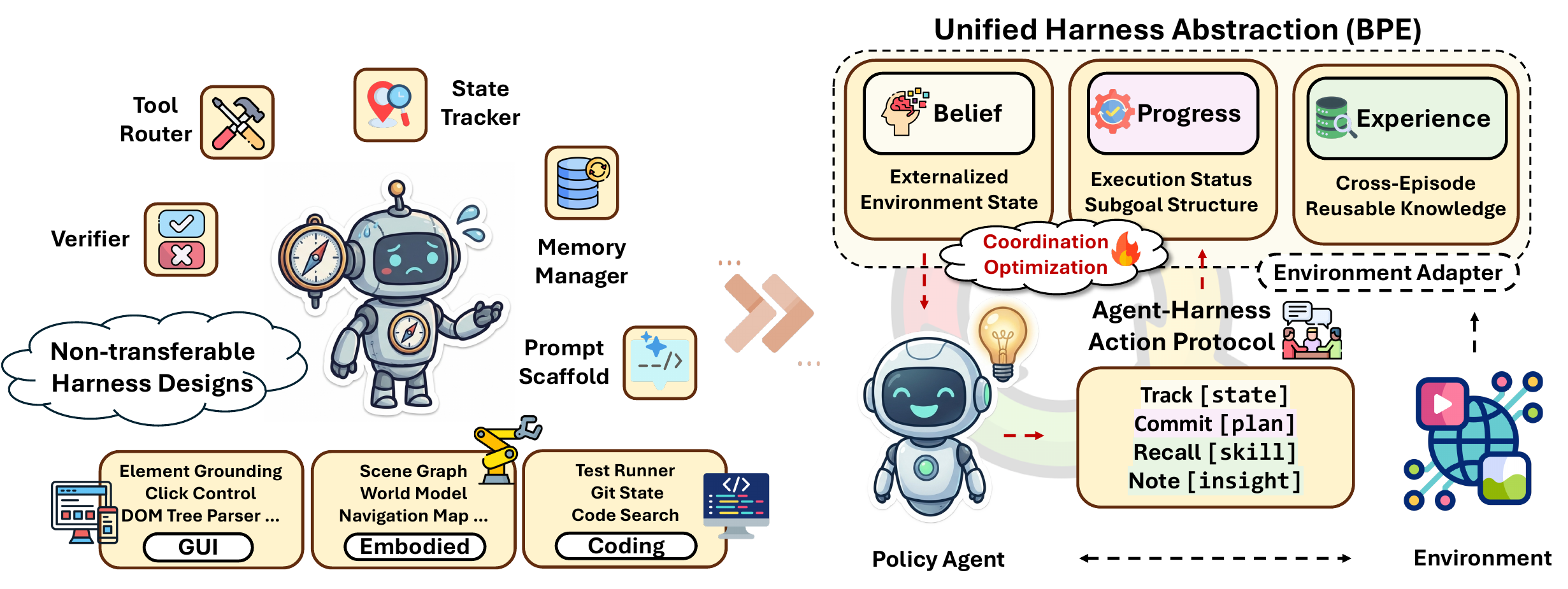}
\caption{
Overview of \textsc{EvoHarness-RL}.
Long-horizon agents rely on heterogeneous external execution support that is often manually designed and controlled.
\textsc{EvoHarness-RL} organizes this support into a shared BPE workspace: \emph{Belief} for environment state, \emph{Progress} for execution progress, and \emph{Experience} for reusable knowledge.
The agent coordinates with this workspace through four compact actions, track, commit, recall, and note, and learns when these operations are useful during task execution.
}
\label{fig}
\vspace{-0.5cm}
\end{figure}

To address this challenge, we propose \textbf{\methodname{}}, a unified framework for learning coordination between an agent and its evolving harness through reinforcement learning. \methodname{} separates environment-specific harness implementations from a shared policy-facing interface, organizing external support into a BPE workspace with three roles: \emph{Belief} for environment state, \emph{Progress} for execution progress and subgoals, and \emph{Experience} for reusable knowledge across episodes~\citep{singh2026agent,wang2026subgoal,reflexion}. We first expose this workspace as an inference-time scaffold in \textsc{EvoHarness-Base}, and then make harness coordination learnable through four compact actions: track, commit, recall, and note. We bootstrap this interface from successful trajectories and further optimize harness use with cost-aware GRPO~\citep{grpo}. A transient diversity reward encourages exploration of harness operations early in training, while a success-gated efficiency reward promotes increasingly selective coordination.

We evaluate \methodname{} across multiple model backbones and heterogeneous
long-horizon agent tasks~\citep{alfworld,yao2022webshop,zhou2024webarena}. The shared BPE scaffold
improves average success rate by 10.0 percentage points in inference time
execution across multiple frontier models, while learned harness coordination
improves average success rate by 8.5 percentage points over the strongest open-source baseline.
By externalizing state, progress, and reusable
experience, the scaffold supports early exploration, improves RL sample
efficiency, and reduces reliance on environment-specific behaviors, leading to stronger generalization on unseen tasks, with a cross-environment
transfer experiment further suggesting the potential for faster adaptation to a new environment. Beyond aggregate performance, our
empirical analyses reveal three characteristic dynamics. \emph{Harness
annealing:} training progressively internalizes recurring harness-use patterns
into the policy, shifting from frequent scaffold-like calls toward selective
external state access. \emph{Coordination specialization:} different
environments preserve different harness operations, suggesting that training
learns an environment-dependent boundary between what should be internalized by
the policy and what should remain externalized. \emph{Harness evolution:}
online progress updates keep the workspace aligned with ongoing execution,
while cross-episode experience consolidation refines knowledge to
support generalization to unseen tasks.

Our contributions are three-fold:
\begin{itemize}[itemsep=0pt,topsep=2pt,leftmargin=*]

\item \textbf{Unified Harness Abstraction.}
We introduce BPE as a shared policy-facing abstraction for heterogeneous harness
designs, separating environment-specific implementations from agent-side
coordination. The same abstraction supports both inference-time scaffolding with
\textsc{EvoHarness-Base} and trainable harness coordination with \methodname{}.

\item \textbf{Learnable Coordination.}
We formulate harness use as a learnable coordination problem through
a compact agent-harness action protocol, supervised initialization,
and cost-aware reinforcement learning. Our theoretically grounded
learning paradigm guides the policy from broad exploration of harness
operations toward selective, task-adaptive specialization.

\item \textbf{Empirical Insights.}
Across three heterogeneous agent environments and multiple model
families, our framework delivers substantial gains in task success
through inference-time scaffolding and learned harness coordination. Experiments
also show that \methodname{} improves RL rollout efficiency and strengthens
generalization to unseen tasks. We also identify agent and harness co-evolution
dynamics, including harness annealing, coordination specialization, and harness
evolution, which further show how training
shapes the policy to coordinate more effectively with its external workspace.

\end{itemize}

\section{Method}
\label{sec:method}

We introduce \methodname{}, a trainable framework for coordinating an agent with
its external harness. Section~\ref{sec:bpe_abstraction} defines a unified BPE
abstraction and its environment adapters; Section~\ref{sec:action_protocol}
presents a compact protocol for accessing and updating external state; and
Section~\ref{sec:training} describes the two-stage coordination training recipe.

\subsection{Unified Harness Abstraction: Belief, Progress, and Experience}
\label{sec:bpe_abstraction}

Long-horizon agents interact over extended contexts and often operate in
partially observable environments. As trajectories grow, they may lose track of
environment state, forget task progress, and repeat mistakes that could have
been avoided using prior experience
~\citep{reflexion,wang2026subgoal,singh2026agent}. Existing systems often address
these failures through domain-specific harness components, such as state
trackers, task planners, or episodic memory banks. However, these components
typically expose heterogeneous interfaces and rely on manually designed
invocation rules, making them difficult to transfer across domains or optimize
jointly with the agent policy.

We therefore abstract and unify the policy-facing harness state into three complementary
functional roles: \textbf{Belief}, \textbf{Progress}, and \textbf{Experience}
(BPE), corresponding to what the agent believes, how the task is progressing,
and what can be reused from prior experience. BPE defines shared functional
roles rather than a fixed implementation. An \textbf{environment adapter}
grounds these roles and their corresponding harness operations in each domain
while preserving the same policy-facing interface. Details of the BPE
instantiations, including their corresponding low-level tools and state
representations, are provided in Appendix~\ref{app:environment_grounding}.
Formally, at each step $t$, the harness maintains:
\begin{equation}
\mathcal{H}_t = (\Bstate, \Pstate, \Estate).
\end{equation}

\paragraph{Belief (\Bstate).}
Belief maintains persistent within-episode state derived from agent-visible
interaction. Its grounding depends on the environment: embodied agents may use
a world-state tracker over objects and locations, shopping agents maintain
product and constraint state, and general web agents preserve application state
across page transitions. Thus, Belief provides a shared persistent-state
abstraction over different domain-specific representations.

\paragraph{Progress (\Pstate).}
Progress maintains task decomposition and execution status, recording what has
been completed and what remains unresolved. Depending on the domain, this may
represent manipulation subgoals, unmet product requirements, or pending stages
of a multi-step workflow.

\paragraph{Experience (\Estate).}
Experience maintains reusable cross-episode knowledge, organized into
\textit{general experience} shared across tasks and \textit{task-specific
experience} specialized to particular task families. We maintain Experience
through a \textit{solver-evolver} structure. During rollouts, the solver
interacts with the environment using the current Experience and produces new
trajectory evidence through execution. The Experience store is
kept fixed within each rollout batch. At consolidation epochs, the evolver
processes the accumulated trajectories, outcomes, and notes, abstracts reusable
general and task-specific experience, and reconciles it with the existing store
through \textsc{Add}, \textsc{Update}, \textsc{Archive}, and \textsc{Skip}
operations.

We first instantiate BPE as \textbf{\textsc{EvoHarness-Base}}, an always-available
external scaffold for long-horizon execution. Throughout an episode, the harness
continuously maintains and exposes the latest BPE state, while the policy itself
selects only environment actions. In this way, task state, execution progress,
and reusable experience are explicitly externalized rather than requiring the
policy to repeatedly reconstruct or retain them within its own context. This
provides persistent support for long-horizon reasoning without requiring any
policy training.
However, always-on BPE fixes the boundary between what is maintained externally
and what can be handled internally by the policy. As execution progresses and
the policy becomes more capable, some external support may become redundant,
while other information remains useful to retain outside the model. This
motivates learning when to use the harness rather than exposing all BPE support
by default.

\subsection{Agent-Harness Action Protocol}
\label{sec:action_protocol}

To learn when external support should be used, we expose BPE through a compact
set of policy-facing meta actions. Belief maintains environment state,
and \acttrack{} retrieves task relevant information from $B_t$.
Progress maintains execution status, and \actcommit{} records subgoals
or progress updates in $P_t$. Experience maintains reusable cross
episode knowledge, where \actrecall{} retrieves relevant knowledge from $E_t$
and \actnote{} records new evidence for later consolidation. Together, these
operations define the harness action space:
\begin{equation}
\mathcal{A}_{\mathrm{bpe}}
=
\{\acttrack, \actcommit, \actrecall, \actnote\}.
\end{equation}

This protocol preserves distinct BPE roles through a shared interface across environments.

In \textbf{\methodname{}}, harness access becomes part of the policy decision.
Let $\mathcal{A}_{\mathrm{env}}$ denote the environment action space. The policy
acts over the joint action space:
\begin{equation}
\mathcal{A}
=
\mathcal{A}_{\mathrm{env}}
\cup
\mathcal{A}_{\mathrm{bpe}}.
\end{equation}

We distinguish the internal harness state $\mathcal{H}_t$ from the harness
context currently visible to the policy in \methodname{}, denoted by $\mathcal{V}_t$. At step
$t$, the policy receives the environment observation $o_t$, visible harness
context $\mathcal{V}_t$, and task context $c_t$, and samples:
\begin{equation}
a_t \sim \pi_\theta(\cdot \mid o_t, \mathcal{V}_t, c_t).
\end{equation}

If $a_t \in \mathcal{A}_{\mathrm{env}}$, the action advances the environment.
If $a_t \in \mathcal{A}_{\mathrm{bpe}}$, it accesses or updates the
corresponding BPE component and refreshes the visible harness context.
Because both types of actions share the same interaction budget, the policy
must learn when external support is worth its cost and when the corresponding
computation can be handled internally.

\begin{figure}[t]
\centering
\includegraphics[width=\textwidth]{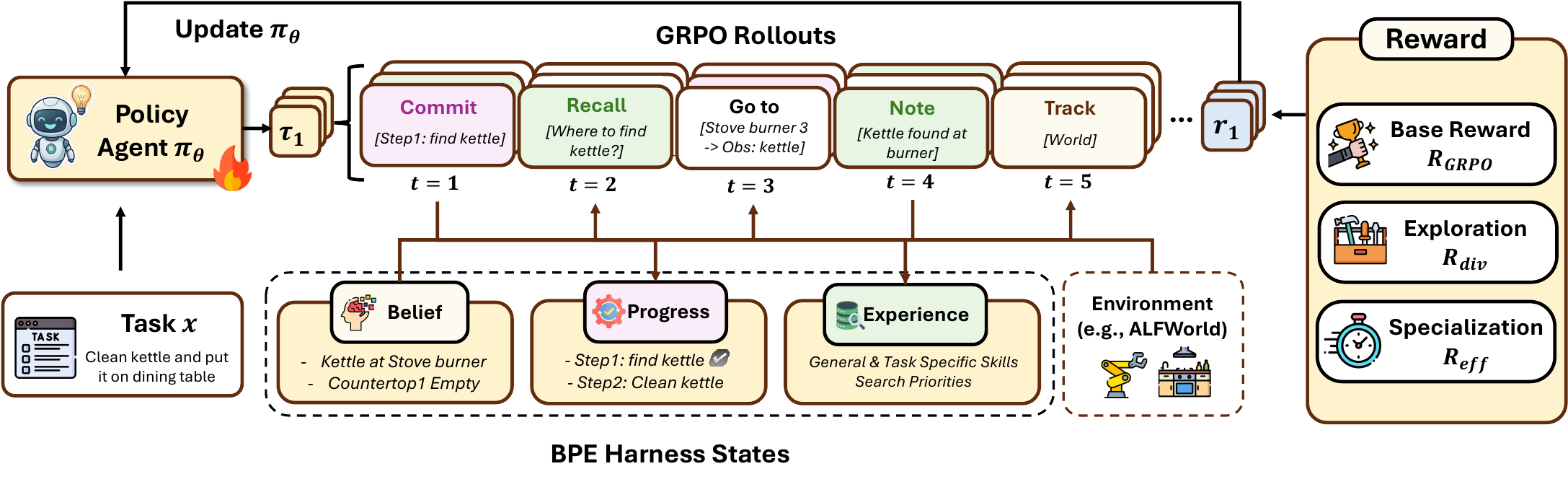}
\caption{
Training pipeline of \methodname{}. Given a task, the policy generates rollouts that interleave environment actions with BPE meta-actions, accessing and updating belief, progress, and experience states. GRPO updates the policy using a composite reward that combines the base objective with an annealed exploration bonus and an efficiency incentive, encouraging a transition from broad harness exploration to selective, task-adaptive coordination.
}
\label{fig:evoharness_training}
\vspace{-0.5cm}
\end{figure}

\subsection{Optimizing Agent and Harness Coordination}
\label{sec:training}

Introducing BPE actions expands the policy from environment interaction to
joint agent and harness control. This creates a structured optimization
problem. The policy must first learn how the new harness actions interact with
external state, and then learn when using them is worth the additional
interaction cost. We therefore separate these two roles. Supervised learning
provides an initial coordination scaffold, while reinforcement learning
refines this scaffold into a selective policy.

\paragraph{Supervised initialization.}
Directly optimizing over
$\mathcal{A}_{\mathrm{env}} \cup \mathcal{A}_{\mathrm{bpe}}$
from sparse task feedback is inefficient. Before the policy can learn
\emph{when} a harness operation is useful, it must first understand \emph{how}
each operation accesses or updates the external BPE state. We therefore use
successful teacher trajectories to bootstrap the joint action interface.

The teacher is provided with the same BPE action definitions exposed to the
policy and generates trajectories that interleave environment actions with
track, commit, recall, and note. We retain
successful trajectories and fine-tune the open-source policy on these
demonstrations. This stage is not intended to determine the final harness-use
strategy. Instead, it initializes a policy that can meaningfully operate in
the joint action space, providing a stronger starting point for subsequent RL
to explore and refine coordination strategies.

\paragraph{Reinforcement learning for harness coordination.}
The supervised policy inherits the teacher's harness-use pattern, which may
externalize state even when the same decision could be handled internally.
The remaining problem is therefore not whether the policy can use the
harness, but whether a particular harness interaction is worth its cost in
the current trajectory.

Starting GRPO from this initialization changes the role of exploration.
Rather than discovering the BPE interface from scratch, the policy can use
its rollout budget to compare alternative coordination strategies and learn
which external-state operations improve task execution. Meanwhile, this BPE
scaffold improves rollout efficiency for success discovery during GRPO training,
as analyzed in Appendix~\ref{app:theory-a}. For a rollout
trajectory $\tau$ collected at RL epoch $u$, we augment the reward used by
standard GRPO with two complementary terms:
\begin{equation}
R(\tau;u)
=
R_{\mathrm{GRPO}}(\tau)
+
\underbrace{\lambda_{\mathrm{div}}(u)R_{\mathrm{div}}(\tau)}
_{\text{Exploration}}
+
\underbrace{\lambda_{\mathrm{eff}}R_{\mathrm{eff}}(\tau)}
_{\text{Specialization}} .
\label{eq:reward}
\end{equation}
Here, $R_{\mathrm{GRPO}}$ denotes the base reward used by our standard GRPO
training, including task success and fixed penalties for invalid or
degenerate actions. Since environment actions and harness actions consume
the same interaction budget, we use
$R_{\mathrm{eff}}(\tau)
=
\mathbf{1}[\mathrm{succ}(\tau)]
\frac{T_{\max}-|\tau|}{T_{\max}}$.
This rewards successful trajectories that complete the task with fewer total
interactions, so unnecessary harness calls reduce the return through the same
budget as redundant environment actions.

\paragraph{Annealing exploration into specialization.}
Efficiency pressure alone can suppress newly introduced BPE actions before
the policy has collected enough evidence to discover their utility. We
therefore use a transient exploration bonus that encourages coverage of the
BPE interface early in training. Specifically,
$R_{\mathrm{div}}(\tau)
=
|\{a \in \mathcal{A}_{\mathrm{bpe}}:
a \text{ occurs in } \tau\}|/
|\mathcal{A}_{\mathrm{bpe}}|$,
which measures the fraction of BPE action types exercised in a trajectory.
Its weight follows
$\lambda_{\mathrm{div}}(u)
=
\frac{\lambda_{\mathrm{div}}^{\max}}{2}
\left(1+\cos\left(\pi\min(u,U)/U\right)\right)$,
where $U$ is the annealing horizon.

\paragraph{Convergence with diversity annealing.}
Let $J$ be the final cost-aware objective with maximum $J^*$, and let
$b\sim\mu_u$ be a coordination rule at epoch $u$, with scheduled-objective
residual $\varepsilon_{\mathrm{opt}}(u)$ and diversity weight
$\lambda_u=\lambda_{\mathrm{div}}(u)$.

\begin{thmbox}
\begin{theorem}[Convergence with diversity annealing]
\label{prop:annealed-diversity}
Let $\mathcal B^*$ be the set of coordination rules attaining $J^*$, and let
$\Delta_J>0$ be the reward gap to the best non-optimal rule. Then
\begin{equation}
J^*-J(\mu_u)\le \varepsilon_{\mathrm{opt}}(u)+\lambda_u,
\label{eq:objective_gap_bound}
\end{equation}
and
\begin{equation}
\Pr(b\in\mathcal B^*)
\ge
1-\frac{1}{\Delta_J}
\bigl(\varepsilon_{\mathrm{opt}}(u)+\lambda_u\bigr).
\label{eq:optimal-calling-probability}
\end{equation}
In particular, $\lambda_u=0$ for all $u\ge U$, so if
$\varepsilon_{\mathrm{opt}}(u)\to0$, then $J(\mu_u)\to J^*$ and
$\Pr(b\in\mathcal B^*)\to1$.
Full definitions and proofs are provided in
Appendix~\ref{app:theory-b}.
\end{theorem}
\end{thmbox}

Since $R_{\mathrm{div}}(\tau)\in[0,1]$, the temporary exploration bonus changes
the final objective by at most $\lambda_u$ beyond the current optimization
residual. Under our cosine schedule, $\lambda_u=0$ after annealing, so the
diversity term introduces no further objective-level distortion and retained
harness use is determined only by task reward, interaction efficiency, and the
remaining optimization error.

\FloatBarrier

\section{Experiments}
\label{sec:experiments}

We evaluate \textsc{EvoHarness-Base} for inference time scaffolding and
\methodname{} for learned harness coordination across three heterogeneous long
horizon agent environments. We compare against strong inference time and
trainable baselines, study the contributions of individual components and
generalization to unseen tasks, and analyze agent and harness coevolution
dynamics.

\begingroup
\renewcommand{\deltagain}[1]{{\color{darkred}#1}}
\renewcommand{\oursdelta}[1]{{\color{darkred}\bfseries\boldmath #1}}
\newcommand{\deltadown}[1]{{\color{darkred}#1}}
\newcommand{\bestcell}[1]{\cellcolor{yellow!22}#1}
\newcommand{\secondcell}[1]{#1}
\begin{table*}[t]
\centering
\small
\setlength{\tabcolsep}{4.2pt}
\renewcommand{\arraystretch}{1.08}
\caption{\small
Main results on ALFWorld, WebShop, and WebArena; \colorbox{yellow!22}{\textbf{best}}
marks the column best for each backbone. Qwen3-4B results and three-seed
means and standard deviations for \textsc{EvoHarness-RL} are in
Appendices~\ref{app:cross_backbone_adaptability}
and~\ref{app:seed_robustness}. \frozen/\trainable denote frozen/trainable
methods; \deltagain{$\Delta$} averages the four aggregate-metric differences
from paired ReAct. \textsuperscript{\dag} marks methods whose ALFWorld and
WebShop results are transcribed from SkillOS~\citep{ouyang2026skillos}. All
values are percentages.
}
\label{tab:main_results}

\resizebox{\textwidth}{!}{
\begin{tabular}{l|*{6}{r}|>{\columncolor{gray!10}}r|*{2}{r}|*{4}{r}|>{\columncolor{gray!10}}r|r}
\toprule
\multirow{2}{*}{\textbf{Approach}}
& \multicolumn{7}{c}{\textbf{ALFWorld}}
& \multicolumn{2}{|c|}{\textbf{WebShop}}
& \multicolumn{5}{c|}{\textbf{WebArena}}
& \\
\cmidrule(lr){2-8}
\cmidrule(lr){9-10}
\cmidrule(lr){11-15}
& \textbf{Pick}
& \textbf{Look}
& \textbf{Clean}
& \textbf{Heat}
& \textbf{Cool}
& \textbf{Pick2}
& \textbf{Avg.}
& \textbf{SR}
& \textbf{Score}
& \textbf{SP}
& \textbf{CMS}
& \textbf{Maps}
& \textbf{G\&R}
& \textbf{Avg.}
& \deltagain{\textbf{$\Delta$}} \\
\midrule

\rowcolor{gray!16}
\multicolumn{16}{l}{\textbf{\textit{Frontier Models}}} \\

ReAct (Claude)\frozen
& 100.0 & 92.3 & 100.0 & 100.0 & 88.0 & 95.8
& 96.4
& 40.2 & 57.3
& 26.7 & 13.3 & 30.0 & 20.0 & 22.0
& \deltagain{--} \\

\hspace{0.7em}+ \textbf{\textsc{EvoHarness-Base}}\frozen
& 100.0 & \bestcell{\textbf{100.0}} & 100.0 & 100.0 & \bestcell{\textbf{96.0}} & 95.8
& \bestcell{\textbf{98.6}}
& \bestcell{\textbf{46.4}} & \bestcell{\textbf{61.5}}
& \bestcell{\textbf{33.3}}
& \bestcell{\textbf{20.0}}
& \bestcell{\textbf{40.0}}
& \bestcell{\textbf{50.0}}
& \bestcell{\textbf{34.0}}
& \oursdelta{$+6.2$} \\

ReAct (GLM)\frozen
& 77.1 & 69.2 & 48.1 & 50.0 & 20.0 & 45.8
& 52.1
& 46.2 & 62.8
& 26.7 & 13.3 & \bestcell{\textbf{20.0}} & 20.0 & 20.0
& \deltagain{--} \\

\hspace{0.7em}+ \textbf{\textsc{EvoHarness-Base}}\frozen
& \bestcell{\textbf{94.3}} & \bestcell{\textbf{84.6}} & \bestcell{\textbf{74.1}} & \bestcell{\textbf{68.8}} & \bestcell{\textbf{52.0}} & \bestcell{\textbf{87.5}}
& \bestcell{\textbf{77.9}}
& \bestcell{\textbf{49.6}} & \bestcell{\textbf{70.0}}
& 26.7
& \bestcell{\textbf{20.0}}
& 10.0
& \bestcell{\textbf{30.0}}
& \bestcell{\textbf{22.0}}
& \oursdelta{$+9.6$} \\

ReAct (GPT)\frozen
& 74.3 & 53.8 & 48.1 & 62.5 & 60.0 & 58.3
& 60.7
& 37.8 & 56.1
& 26.7 & 26.7 & \bestcell{\textbf{10.0}} & 10.0 & 20.0
& \deltagain{--} \\

\hspace{0.7em}+ \textbf{\textsc{EvoHarness-Base}}\frozen
& \bestcell{\textbf{97.1}} & \bestcell{\textbf{76.9}} & \bestcell{\textbf{85.2}} & \bestcell{\textbf{93.8}} & \bestcell{\textbf{88.0}} & \bestcell{\textbf{62.5}}
& \bestcell{\textbf{85.0}}
& \bestcell{\textbf{46.2}} & \bestcell{\textbf{63.9}}
& \bestcell{\textbf{40.0}} & \bestcell{\textbf{33.3}} & 0.0 & \bestcell{\textbf{20.0}} & \bestcell{\textbf{26.0}}
& \oursdelta{$+11.6$} \\

\midrule
\rowcolor{gray!16}
\multicolumn{16}{l}{\textbf{\textit{Open-Source Baselines (Qwen3-8B)}}} \\

ReAct\frozen
& 78.1 & 46.2 & 33.3 & 37.5 & 29.3 & 47.2
& 47.8
& 12.0 & 42.9
& 20.0 & 13.3 & 0.0 & \secondcell{20.0} & 14.0
& \deltagain{--} \\

ExpeL\frozen
& 91.4 & 76.9 & 14.8 & 43.8 & 28.0 & 45.8
& 50.7
& 7.0 & 25.7
& 20.0 & \secondcell{26.7} & 0.0 & \secondcell{20.0} & \secondcell{18.0}
& \deltadown{$-3.8$} \\

ReasoningBank\textsuperscript{\dag}\frozen
& 83.8 & 48.7 & 49.4 & 39.6 & 41.3 & 54.2
& 56.2
& 11.4 & 35.4
& 20.0 & 13.3 & \secondcell{10.0} & 10.0 & 14.0
& \deltagain{$+0.1$} \\

MemP\textsuperscript{\dag}\frozen
& 80.0 & 43.6 & 24.7 & 33.3 & 38.7 & 48.6
& 47.9
& 12.0 & 35.7
& 20.0 & 20.0 & 0.0 & 10.0 & 14.0
& \deltadown{$-1.8$} \\

Dynamic Cheatsheet\frozen
& 88.6 & 53.8 & 29.6 & 37.5 & 20.0 & 66.7
& 52.1
& 5.2 & 20.4
& 20.0 & \bestcell{\textbf{33.3}} & 0.0 & 10.0 & \secondcell{18.0}
& \deltadown{$-5.3$} \\

ACE\frozen
& 85.7 & 38.5 & 29.6 & 37.5 & 36.0 & 58.3
& 51.4
& 3.2 & 13.2
& 20.0 & \secondcell{26.7} & 0.0 & 10.0 & 16.0
& \deltadown{$-8.2$} \\

SkillOS-base\textsuperscript{\dag}\frozen
& 79.0 & 41.0 & 45.7 & 37.5 & 38.7 & 55.6
& 53.1
& 13.6 & 38.6
& 20.0 & 20.0 & 0.0 & \secondcell{20.0} & 16.0
& \deltagain{$+1.2$} \\

GRPO\trainable
& 88.6 & 69.2 & 74.1 & 68.8 & 48.0 & 41.7
& 66.4
& 73.0 & 82.4
& 33.3 & 13.3 & \bestcell{\textbf{20.0}} & 0.0 & \secondcell{18.0}
& \deltagain{$+30.8$} \\

SkillOS\textsuperscript{\dag}\trainable
& 85.7 & 56.4 & 54.3 & 43.8 & 46.7 & 62.5
& 61.2
& 16.5 & 40.6
& \secondcell{26.7} & 20.0 & \secondcell{10.0} & 10.0 & \secondcell{18.0}
& \deltagain{$+4.9$} \\

SkillRL\trainable
& 94.3
& \secondcell{84.6}
& \secondcell{92.6}
& \secondcell{87.5}
& 72.0
& 75.0
& \secondcell{85.0}
& \secondcell{73.2} & \secondcell{85.5}
& 20.0 & 20.0 & \secondcell{10.0} & \secondcell{20.0} & \secondcell{18.0}
& \deltagain{$+36.3$} \\

\midrule
\rowcolor{gray!16}
\multicolumn{16}{l}{\textbf{\textit{Ours (Qwen3-8B)}}} \\

\textbf{\textsc{EvoHarness-Base}}\frozen
& 71.4 & 53.8 & 63.0 & 50.0 & 48.0 & 41.7
& 56.4
& 18.6 & 43.7
& 20.0 & \secondcell{26.7} & \secondcell{10.0} & 10.0 & \secondcell{18.0}
& \oursdelta{$+5.0$} \\

\textbf{\textsc{EvoHarness-SFT}}\trainable
& 80.0 & 53.8 & 88.9 & 75.0 & 40.0 & 62.5
& 68.6
& 43.8 & 48.1
& 33.3 & 6.7 & 0.0 & \secondcell{20.0} & 16.0
& \oursdelta{$+15.0$} \\

\textbf{\textsc{EvoHarness-RL}}\trainable
& \bestcell{\textbf{100.0}}
& \bestcell{\textbf{92.3}}
& \bestcell{\textbf{96.3}}
& \bestcell{\textbf{100.0}}
& \bestcell{\textbf{84.0}}
& \bestcell{\textbf{95.8}}
& \bestcell{\textbf{95.0}}
& \bestcell{\textbf{80.6}}
& \bestcell{\textbf{90.8}}
& \bestcell{\textbf{40.0}}
& 20.0
& \secondcell{10.0}
& \bestcell{\textbf{30.0}}
& \bestcell{\textbf{26.0}}
& \oursdelta{$+43.9$} \\

\bottomrule
\end{tabular}
}
\end{table*}
\endgroup

\vspace{-3mm}
\subsection{Experiment Setup}
\vspace{-0.35em}
\paragraph{Environments.}
\textbf{ALFWorld}~\citep{alfworld} is a text-based household environment with
six task families: pick-and-place (\textsc{Pick}), inspection under light
(\textsc{Look}), cleaning (\textsc{Clean}), heating (\textsc{Heat}), cooling
(\textsc{Cool}), and placing two objects (\textsc{Pick2}).
\textbf{WebShop}~\citep{yao2022webshop} evaluates product search and purchase
under compositional constraints. For both benchmarks, we use the SkillRL
training and validation splits~\citep{xia2026skillrl}.
\textbf{WebArena}~\citep{zhou2024webarena} covers realistic multi-site tasks in
shopping (\textsc{SP}), content management (\textsc{CMS}), maps, and GitLab/Reddit
(\textsc{G\&R}). Following AgentGym-RL~\citep{xi2025agentgymrl}, we exclude
backend-edit tasks to avoid persistent state changes across on-policy
rollouts. Together, these distinct environments test whether our method
generalizes across domains and interaction structures in long-horizon agentic
tasks.

\paragraph{Baselines.}
We compare \methodname{} against frontier and open-source models. First, we
evaluate frontier models from three families: Claude Opus 4.6, GLM-5, and GPT-5. For each model, we report standard ReAct and
then apply our inference-time \textsc{EvoHarness-Base}, as described in
Section~\ref{sec:action_protocol}. Second, we evaluate open-source baselines,
all using the same Qwen3-8B backbone. The \textbf{frozen inference-time
baselines} are ReAct~\citep{react}, ExpeL~\citep{zhao2024expel},
ReasoningBank~\citep{ouyang2025reasoningbank}, MemP~\citep{fang2026memp},
Dynamic Cheatsheet~\citep{suzgun2026dynamic}, ACE~\citep{zhang2025agentic},
and SkillOS-base~\citep{ouyang2026skillos}; these methods use prompting or
external memory without updating model parameters. The \textbf{training-time
baselines} are standard GRPO~\citep{grpo}, SkillOS~\citep{ouyang2026skillos},
and SkillRL~\citep{xia2026skillrl}, which optimize the policy through
reinforcement learning or memory-augmented training. We compare them with our
variants: \textsc{EvoHarness-Base},
\textsc{EvoHarness-SFT}, and
\textsc{EvoHarness-RL}, instantiated with
Qwen3-8B. Detailed baseline descriptions are provided in
Appendix~\ref{app:baseline_details}, and implementation details in
Appendix~\ref{app:implementation_details}.

\subsection{Main Results}
\vspace{-0.35em}
Table~\ref{tab:main_results} shows that \methodname{} consistently benefits both
inference time scaffolding and learned harness coordination across heterogeneous
agent environments. On frontier models, \textsc{EvoHarness-Base} yields a
\textbf{10.0 percentage-point} improvement in average success rate over
the corresponding ReAct agents, demonstrating that
the shared BPE scaffold provides effective long horizon support without parameter
updates. In contrast, existing frozen memory and harness baselines on Qwen3-8B
show mixed or even negative gains across environments, suggesting that manually
designed external support does not necessarily generalize across different
agentic settings. Learning the coordination policy further amplifies the benefit:
\methodname{} yields an \textbf{8.5 percentage-point} improvement in
average success rate over the strongest trainable baseline. Supervised
initialization is less consistent, suggesting that imitation inherits teacher
specific harness use patterns, whereas coordination RL learns when and how to
use external support.

\begin{table}[t]
    \centering
    \begin{minipage}[t]{0.47\textwidth}
        \vspace{0pt}
        \centering
        \begin{minipage}[t]{\linewidth}
\vspace{0pt}
\centering
\scriptsize
\setlength{\tabcolsep}{3.0pt}
\renewcommand{\arraystretch}{1.08}
\caption{\small
Ablation results on ALFWorld.
}
\label{tab:ablation}

\resizebox{\linewidth}{!}{
\begin{tabular}{lrrrrrrr@{\hspace{4pt}}r}
\toprule
\multirow{2}{*}{\textbf{Variant}}
& \multicolumn{7}{c}{\textbf{ALFWorld}}
& \multirow{2}{*}{\deltagain{\textbf{$\Delta$}}} \\
\cmidrule(lr){2-8}
& \textbf{Pick}
& \textbf{Look}
& \textbf{Clean}
& \textbf{Heat}
& \textbf{Cool}
& \textbf{Pick2}
& \textbf{Avg.}
& \\
\midrule

\rowcolor{yellow!15}
\multicolumn{9}{l}{\textbf{\textit{Training-time BPE ablation for \textsc{EvoHarness-RL}}}} \\

Full BPE
& 100.0
& 92.3
& 96.3
& 100.0
& 84.0
& 95.8
& 95.0
& \deltagain{--} \\

w/o Belief
& 97.1
& 76.9
& 88.9
& 87.5
& 84.0
& 91.7
& 89.3
& \deltagain{$\blacktriangledown$\,5.7} \\

w/o Progress
& 100.0
& 84.6
& 100.0
& 100.0
& 84.0
& 70.8
& 90.7
& \deltagain{$\blacktriangledown$\,4.3} \\

w/o Experience
& 97.1
& 76.9
& 88.9
& 81.2
& 84.0
& 66.7
& 84.3
& \deltagain{$\blacktriangledown$\,10.7} \\

Always-on BPE
& 97.1
& 92.3
& 85.2
& 100.0
& 68.0
& 91.7
& 88.6
& \deltagain{$\blacktriangledown$\,6.4} \\

w/o SFT
& 68.6
& 53.8
& 55.6
& 50.0
& 52.0
& 12.5
& 50.0
& \deltagain{$\blacktriangledown$\,45.0} \\

SFT + GRPO
& 100.0
& 61.5
& 92.6
& 81.3
& 80.0
& 91.7
& 87.9
& \deltagain{$\blacktriangledown$\,7.1} \\

\midrule

\rowcolor{yellow!15}
\multicolumn{9}{l}{\textbf{\textit{Reward ablation for \textsc{EvoHarness-RL}}}} \\

Full Reward
& 100.0
& 92.3
& 96.3
& 100.0
& 84.0
& 95.8
& 95.0
& \deltagain{--} \\

w/o Diversity
& 100.0
& 100.0
& 96.3
& 93.8
& 84.0
& 83.3
& 92.9
& \deltagain{$\blacktriangledown$\,2.1} \\

w/o Efficiency
& 100.0
& 53.8
& 96.3
& 75.0
& 84.0
& 83.3
& 86.4
& \deltagain{$\blacktriangledown$\,8.6} \\

\bottomrule
\end{tabular}
}
\vspace{-4pt}
\end{minipage}

    \end{minipage}
    \hfill
    \begin{minipage}[t]{0.50\textwidth}
        \vspace{0pt}
        \centering
        \begin{minipage}[t]{\linewidth}
\vspace{0pt}
\centering
\scriptsize
\setlength{\tabcolsep}{2.8pt}
\renewcommand{\arraystretch}{1.18}
\captionof{table}{\small
Generalization results on the ALFWorld unseen split for ReAct, vanilla
SFT + GRPO, and \methodname{} variants.
}
\label{tab:alfworld_unseen_results}

\resizebox{\linewidth}{!}{
\begin{tabular}{lrrrrrrr}
\toprule
\multirow{2}{*}{\textbf{Approach}}
& \multicolumn{7}{c}{\textbf{ALFWorld Unseen SR (\%)}} \\
\cmidrule(lr){2-8}
& \textbf{Pick}
& \textbf{Look}
& \textbf{Clean}
& \textbf{Heat}
& \textbf{Cool}
& \textbf{Pick2}
& \textbf{Avg.} \\
\midrule

ReAct (Opus 4.6)\frozen
& 100.0
& 83.3
& 96.8
& 78.3
& 90.5
& 94.1
& 91.0 \\

+ \textsc{EvoHarness-Base}\frozen
& 100.0
& 94.4
& 100.0
& 95.7
& 100.0
& 94.1
& \textbf{97.8} \\

ReAct (Qwen3-8B)\frozen
& 66.7
& 50.0
& 48.4
& 47.8
& 52.4
& 29.4
& 50.0 \\

+ \textsc{EvoHarness-Base}\frozen
& 83.3
& 38.9
& 87.1
& 91.3
& 90.5
& 58.8
& 77.6 \\

+ \textsc{EvoHarness-SFT}\trainable
& 70.8
& 66.7
& 80.6
& 69.6
& 47.6
& 76.5
& 69.4 \\

+ SFT + GRPO\trainable
& 75.0
& 66.7
& 83.9
& 69.6
& 90.5
& 76.5
& 77.6 \\

+ Always-on BPE\trainable
& 91.7
& 83.3
& 96.8
& 100.0
& 81.0
& 82.4
& 90.3 \\

\rowcolor{yellow!12}
+ \textsc{EvoHarness-RL}\trainable
& 95.8
& 100.0
& 93.5
& 91.3
& 100.0
& 88.2
& \textbf{94.8} \\

\bottomrule
\end{tabular}
}
\vspace{-0.8em}
\end{minipage}

    \end{minipage}
    \vspace{-0.8em}
\end{table}

\subsection{Ablation Results}
\vspace{-0.35em}

Table~\ref{tab:ablation} evaluates the contributions of BPE state, the training
pipeline, and the reward design. Removing any BPE component reduces average
performance, consistent with the inference-time ablations in
Appendix~\ref{app:more_bpe_ablations}, with \emph{Experience} causing the largest
drop due to the recurring procedures and reusable object-location knowledge in
ALFWorld.
The training ablations further highlight the importance of learned harness
coordination. Without SFT initialization, the policy struggles to explore the
new BPE action space from sparse rewards. A vanilla SFT + GRPO baseline with the
same supervised-data budget also remains below the full method, showing that the
gain is not simply due to additional post-training. Beyond final performance,
the BPE scaffold also improves rollout efficiency, allowing subsequent GRPO to
reach strong validation performance substantially earlier than vanilla
SFT + GRPO (Appendix~\ref{app:rollout_efficiency}). Similarly, Always-on BPE
underperforms \methodname{}: without control over when to access or update the
harness, the policy receives redundant external context rather than selectively
using the most task-relevant information. Together, these results support learning
when and how to coordinate with the harness rather than merely exposing it to
the agent.
Finally, removing either the diversity or efficiency reward lowers performance;
Appendix~\ref{app:reward_analysis} further analyzes their effects on harness-use
dynamics.

\subsection{Unseen-Task Evaluation}
\vspace{-0.35em}

Table~\ref{tab:alfworld_unseen_results} evaluates whether the benefits of BPE
and learned harness coordination persist on unseen tasks. At inference time,
\textsc{EvoHarness-Base} continues to improve both frontier and open-source
agents without parameter updates, suggesting that the unified BPE scaffold
remains useful beyond the seen split. Learned coordination provides a further
advantage: \textsc{EvoHarness-RL} achieves $94.8\%$ success, corresponding to a
$22.2\%$ relative improvement over vanilla SFT + GRPO and closely matching its
performance on seen tasks. One possible explanation is a benefit of
\emph{cognitive externalization}:
by keeping task state, progress, and reusable experience outside the model,
BPE reduces the need for the policy to encode environment-specific and
task-specific regularities directly in its parameters. This separation may
mitigate overfitting to the training instances and preserve more flexible
decision making when the agent encounters unseen tasks.

\section{Analysis of Agent-Harness Evolution Dynamics}
\label{sec:analysis}
We study how the policy learns selective harness coordination, how this
coordination specializes across environments, and how the Experience bank
evolves. Together, these dynamics reveal an evolving division of labor between
the model and its external harness. Additional results are provided in
Appendix~\ref{app:training_dynamics}.

\ifdefined\MetaLayout
\subsection{Harness Annealing and Internalization}
\else
\subsection{Learning Selective Coordination}
\fi
\vspace{-0.35em}
\ifdefined\MetaLayout
\begin{wrapfigure}{r}{0.43\textwidth}
    \vspace{-1.1em}
    \centering
    \includegraphics[width=0.96\linewidth]{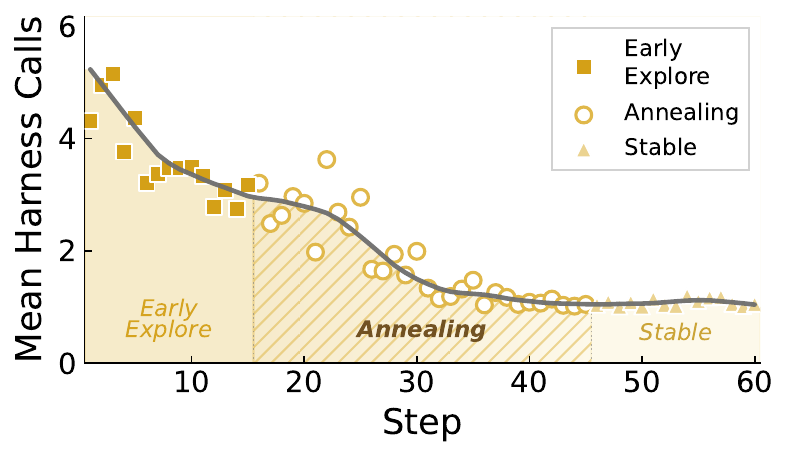}
    \vspace{-0.8em}
    \caption{\small Harness annealing phases. Mean harness calls during ALFWorld
    GRPO training move from early exploration through annealing to a stable regime.}
    \label{fig:harness_calls_by_phase}
    \vspace{-0.8em}
\end{wrapfigure}
\else
\begin{wrapfigure}{r}{0.43\textwidth}
    \vspace{-1.1em}
    \centering
    \includegraphics[width=0.96\linewidth]{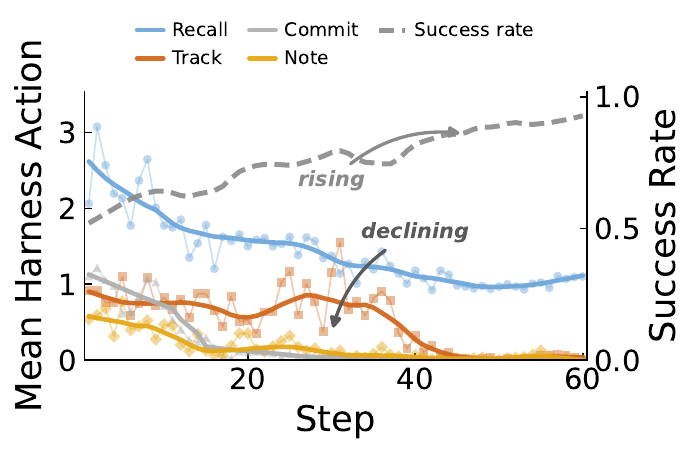}
    \vspace{-0.8em}
    \caption{\small Action-specific harness annealing. Different BPE actions
    decay at different rates, while the success rate is still rising.}
    \label{fig:harness_action_distribution}
    \vspace{-0.8em}
\end{wrapfigure}
\fi
We study how harness coordination evolves as the policy improves.
\ifdefined\MetaLayout
Figure~\ref{fig:harness_calls_by_phase} shows that harness use declines through
early-exploration, annealing, and stable phases, and
Figure~\ref{fig:harness_action_distribution} shows that this decline occurs
while task success continues to rise.
\else
Figure~\ref{fig:harness_action_distribution} shows that harness use declines
while task success continues to rise.
\fi Early in training, the policy relies more
heavily on external support to stabilize long-horizon execution; as recurring
interaction patterns are learned, routine decisions are increasingly handled
internally, leaving the harness for information that remains difficult or
useful to externalize. This shift reflects a transition from broad scaffold
dependence to selective externalization. Different BPE actions stabilize at
different levels, suggesting that training learns a boundary between what can
be internalized by the policy and what remains useful to externalize.
Cross-environment results in Appendix~\ref{app:more_bpe_ablations}
(Figure~\ref{fig:cross_env_annealing}) further show that this boundary varies
with the interaction demands of each environment. Finally, a coordination
perturbation ablation (Table~\ref{tab:schedule_ablation}) shows that harness
frequency alone does not explain the gain; the learned choice and timing of BPE
operations are also important.

\subsection{Learning Environment-Dependent Coordination}
\label{sec:environment_dependent_retention}
\vspace{-0.35em}
\ifdefined\MetaLayout
\begin{figure}[t]
    \centering
    \begin{minipage}[b]{0.52\textwidth}
        \centering
        \includegraphics[height=5.3cm,width=\linewidth,keepaspectratio]{figures/harness_action_distribution.pdf}
        \caption{\small Action-specific harness annealing. Different BPE actions
        decay at different rates, while the success rate is still rising.}
        \label{fig:harness_action_distribution}
    \end{minipage}\hfill
    \begin{minipage}[b]{0.44\textwidth}
        \centering
        \includegraphics[height=5.3cm,width=\linewidth,keepaspectratio]{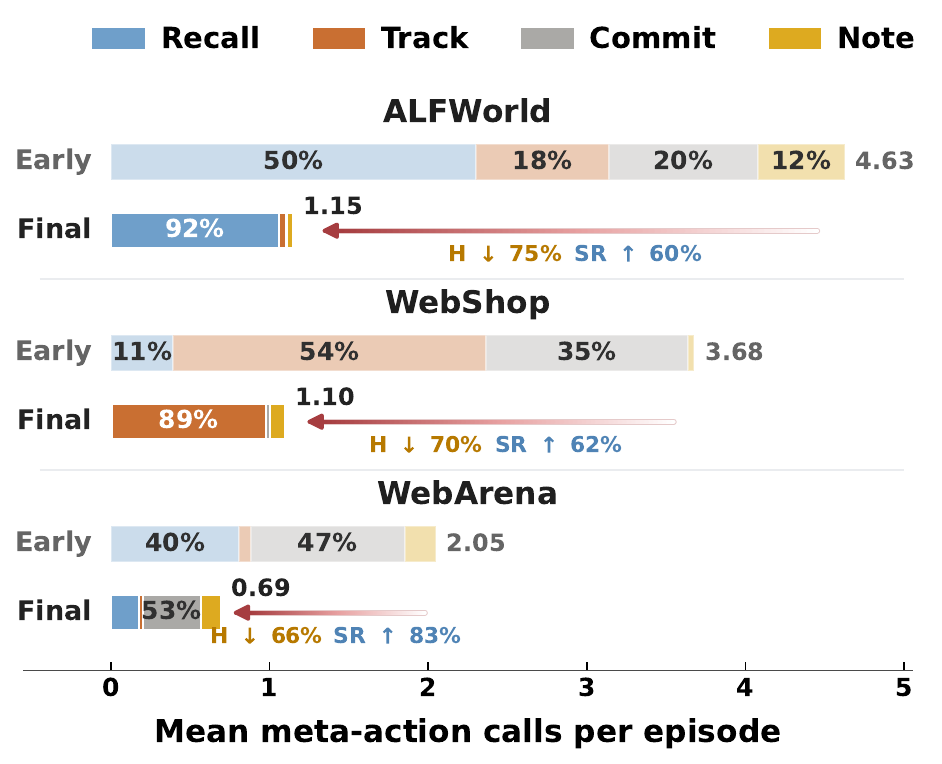}
        \caption{\small Environment-dependent retention of meta-actions across
        benchmarks. $H$ denotes the number of harness actions, and $\mathrm{SR}$
        denotes training success rate.}
        \label{fig:cross_benchmark_annealing}
    \end{minipage}
\end{figure}
\else
\begin{wrapfigure}{r}{0.43\textwidth}
    \vspace{-0.8em}
    \centering
    \includegraphics[width=0.96\linewidth]{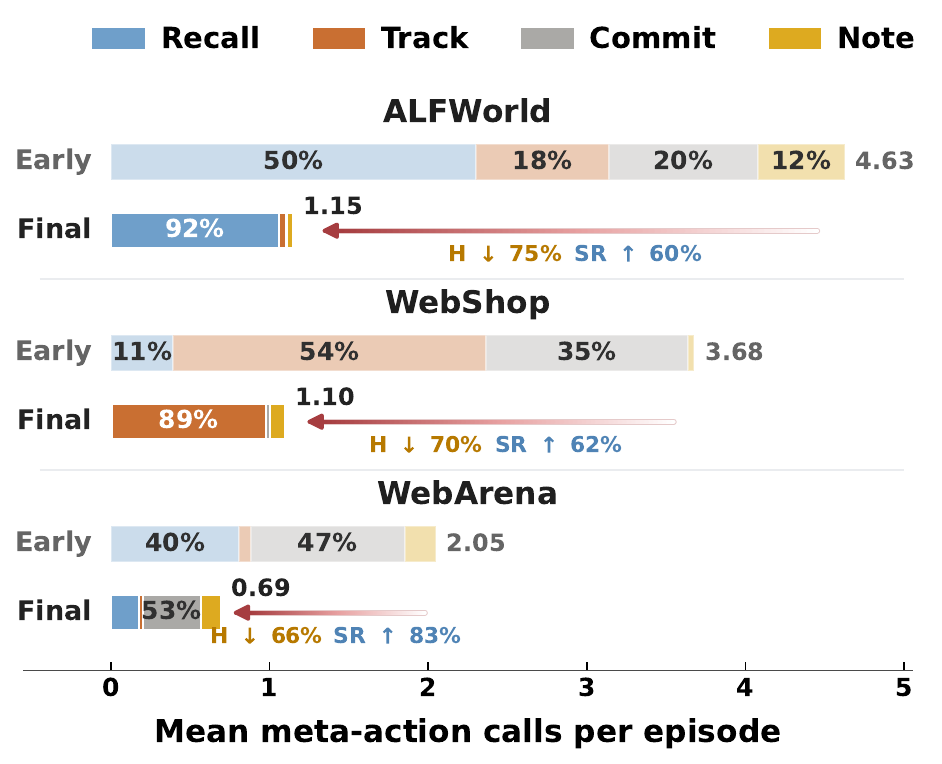}
    \caption{\small Environment-dependent retention of meta-actions across
    benchmarks. $H$ denotes the number of harness actions, and $\mathrm{SR}$
    denotes training success rate.}
    \label{fig:cross_benchmark_annealing}
    \vspace{-0.6em}
\end{wrapfigure}
\fi
We compare harness retention across the three environments.
Figure~\ref{fig:cross_benchmark_annealing} shows that harness use decreases in
all three, but the retained meta-actions differ with the task
structure. ALFWorld consists of recurring household routines in partially
observed scenes, where object locations and manipulation procedures are often
reusable across episodes. This structure makes recalled experience especially
valuable. WebShop instead presents a new combination of product attributes and
purchase constraints in each episode. General shopping procedures transfer,
but the policy must primarily preserve the current request and intermediate
search state, which favors tracking. WebArena spans heterogeneous websites and
longer workflows with changing subgoals, so no single state type is sufficient;
its retained actions remain more mixed, with progress management playing a
larger role. The key result is therefore not a uniform reduction in harness
use. Rather, the same training recipe learns different patterns of harness use
depending on the environment. Appendix~\ref{app:cross_domain_transfer} further
shows that these different usage patterns can still share a transferable
underlying coordination strategy: a WebShop-trained \methodname{} adapts to
ALFWorld substantially faster than standard GRPO and quickly approaches the
in-domain training trajectory. This suggests that the transferable structure
lies in the underlying coordination strategy for deciding when external
support is useful, rather than in a fixed source-domain pattern of harness
actions.

\subsection{Harness Evolution}
\vspace{-0.7em}
Table~\ref{tab:skill_evolution} (right) shows a clear transition from
acquisition to refinement. Early consolidation cycles are dominated by
\textsc{Add} operations that accumulate general strategies, task-specific
procedures, common mistakes, and search priorities. As the bank fills, additions
saturate and consolidation shifts toward \textsc{Update}, which incorporates
later evidence, corrects stale entries, and reconciles redundant or conflicting
experience. Experience evolution is therefore not append-only accumulation,
but continual maintenance of reusable knowledge as training progresses.
Table~\ref{tab:skill_evolution} (left) studies whether Experience should remain
fixed or co-evolve with the policy during RL. Each variant is trained separately
under its corresponding Experience mechanism, so the comparison reflects how
Experience evolution shapes policy learning rather than an inference-time bank
swap. A fixed bank provides useful prior knowledge but cannot adapt to the
changing errors and behaviors of the policy. Continual consolidation instead
forms a feedback loop in which rollout evidence refines Experience and the
updated bank guides subsequent policy learning. The larger gain on unseen tasks
suggests that this co-evolution improves not only knowledge accumulation but
also abstraction and transferability. Self-consolidation with Qwen3-8B remains
effective without a frontier consolidator, showing that the pipeline is robust
even when Experience evolution is performed by the policy model itself.
Appendix~\ref{app:case_study} provides qualitative examples of this evolution
process.

\begin{table}[t]
    \vspace{-0.5em}
    \centering
    \begin{minipage}[t]{0.49\textwidth}
        \vspace{0pt}
        \centering
        \begin{minipage}[t]{\linewidth}
\vspace{0pt}
\centering
\scriptsize
\setlength{\tabcolsep}{2.5pt}
\renewcommand{\arraystretch}{1.18}
\resizebox{\linewidth}{!}{
\begin{tabular}{clrrrrrrr}
\toprule
\textbf{Setting}
& \textbf{Experience Variant}
& \textbf{Pick}
& \textbf{Look}
& \textbf{Clean}
& \textbf{Heat}
& \textbf{Cool}
& \textbf{Pick2}
& \textbf{Avg.} \\
\midrule

\multirow{4}{*}{\rotatebox{90}{\textbf{Seen}}}
& w/o Experience
& 97.1
& 76.9
& 88.9
& 81.2
& 84.0
& 66.7
& 84.3 \\

& SFT-init bank
& 100.0
& 92.3
& 92.6
& 100.0
& 84.0
& 91.7
& 93.6 \\

& Self-Consolidate
& 100.0
& 100.0
& 96.3
& 100.0
& 84.0
& 87.5
& 94.3 \\

& Evolved bank
& \cellcolor{yellow!8}\textbf{100.0}
& \cellcolor{yellow!8}\textbf{92.3}
& \cellcolor{yellow!8}\textbf{96.3}
& \cellcolor{yellow!8}\textbf{100.0}
& \cellcolor{yellow!8}\textbf{84.0}
& \cellcolor{yellow!8}\textbf{95.8}
& \cellcolor{yellow!8}\textbf{95.0} \\

\midrule

\multirow{4}{*}{\rotatebox{90}{\textbf{Unseen}}}
& w/o Experience
& 83.3
& 61.1
& 93.5
& 82.6
& 95.2
& 82.4
& 84.3 \\

& SFT-init bank
& 70.8
& 94.4
& 87.1
& 87.0
& 95.2
& 88.2
& 86.6 \\

& Self-Consolidate
& 95.8
& 94.4
& 90.3
& 82.6
& 81.0
& 82.4
& 88.1 \\

& Evolved bank
& \cellcolor{yellow!8}\textbf{95.8}
& \cellcolor{yellow!8}\textbf{100.0}
& \cellcolor{yellow!8}\textbf{93.5}
& \cellcolor{yellow!8}\textbf{91.3}
& \cellcolor{yellow!8}\textbf{100.0}
& \cellcolor{yellow!8}\textbf{88.2}
& \cellcolor{yellow!8}\textbf{94.8} \\

\bottomrule
\end{tabular}
}
\vspace{-6pt}
\end{minipage}

    \end{minipage}
    \hfill
    \begin{minipage}[t]{0.47\textwidth}
        \vspace{-0.2em}
        \centering
        \includegraphics[width=0.86\linewidth,height=0.175\textheight,keepaspectratio]{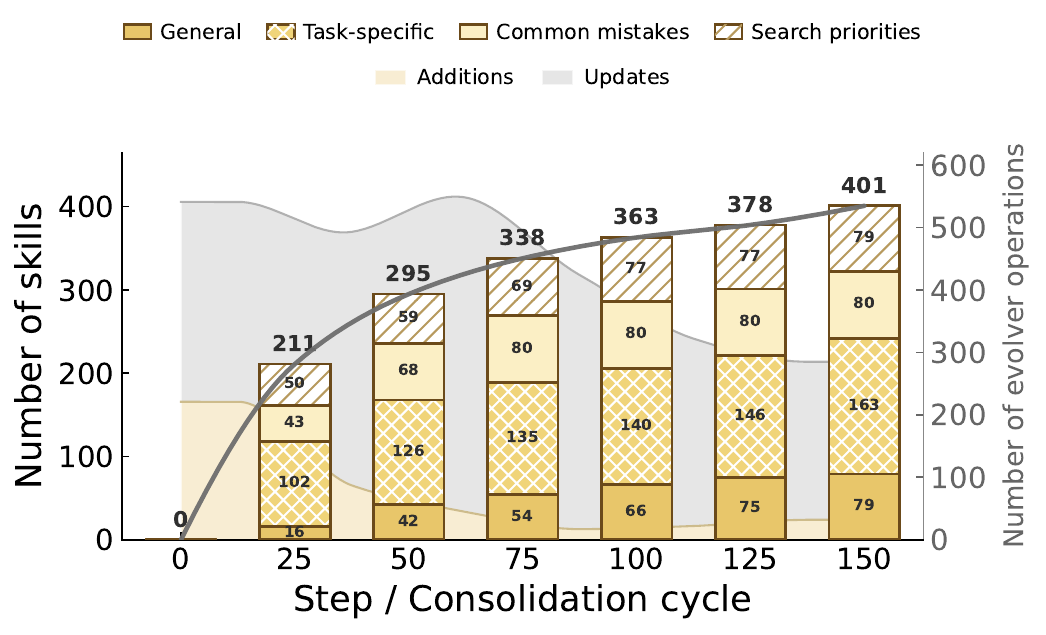}
    \end{minipage}
    \vspace{-0.3em}
    \caption{\small Experience evolution during \methodname{} training.
    Left: ALFWorld seen and unseen success rates of separately trained variants
    under Experience mechanisms. Right: Experience-bank composition
    and consolidation operations over training; task-specific counts aggregate
    entries across task-family categories.}
    \label{tab:skill_evolution}
    \vspace{-0.8em}
\end{table}

\vspace{-0.3em}


\section{Related Work}
\label{sec:related_work}

\vspace{-0.3em}
We review harness engineering and self-evolving agents; see
Appendix~\ref{app:related_work} for details.

\vspace{-3mm}
\paragraph{Harness engineering.}
LLM agents increasingly rely on external harnesses that structure observations,
actions, tools, and execution feedback~\citep{react,toolformer,sweagent,
zhou2024webarena,ning2026code}. Recent work further optimizes or evolves these
systems through state externalization, harness search, trace driven adaptation,
and automated harness construction~\citep{harness1,lee2026meta,harnessx,
zhang2026self,shao2026harness,huang2026memoharness,
ke2026evoharnessbench}. Existing approaches largely study \emph{what harness
should be provided or constructed}. In contrast, \methodname{} unifies
heterogeneous harness functions behind a shared policy facing interface and
learns \emph{how the acting policy coordinates with them at runtime}, including
when external support should be accessed, updated, or bypassed.

\vspace{-3mm}
\paragraph{Memory and self-evolving agents.}
External memory provides a natural mechanism for context externalization in
long horizon agents. Prior work progresses from reusing trajectories and
reflections~\citep{reflexion,zhao2024expel}, to organizing experience into
reusable procedural knowledge~\citep{voyager,wang2025agent,
ouyang2025reasoningbank,fang2026memp}, and further to continuously evolving
external memories and skill repositories~\citep{suzgun2026dynamic,
zhang2025agentic,ouyang2026skillos,xia2026skillrl}. These methods primarily
focus on how memory is represented, retrieved, and evolved within particular
settings. \methodname{} instead places cross episode Experience together with
within episode Belief and Progress under a shared interface, and studies how
the policy learns to coordinate with these forms of externalized context during
execution.

\section{Conclusion}
We introduced \methodname{}, a shared BPE interface that turns coordination
with environment-specific harnesses into a learnable decision problem. Across
multiple benchmarks and model families, BPE improves inference-time execution,
while SFT-initialized, cost-aware reinforcement learning further improves task
success, rollout efficiency, and generalization to unseen tasks. Our analyses show that, as the policy improves, harness use becomes more selective and environment-dependent, revealing an evolving division of labor between the policy and external support. Overall,
\methodname{} provides a step toward jointly optimizing what an agent should
internalize and what should remain external during long-horizon execution.

\newpage

\bibliographystyle{assets/plainnat}
\bibliography{paper}

\begin{thebibliography}{38}
\providecommand{\natexlab}[1]{#1}
\providecommand{\url}[1]{\texttt{#1}}
\expandafter\ifx\csname urlstyle\endcsname\relax
  \providecommand{\doi}[1]{doi: #1}\else
  \providecommand{\doi}{doi: \begingroup \urlstyle{rm}\Url}\fi

\bibitem[Agarwal et~al.(2021)Agarwal, Kakade, Lee, and Mahajan]{agarwal2021policygradient}
Alekh Agarwal, Sham~M. Kakade, Jason~D. Lee, and Gaurav Mahajan.
\newblock On the theory of policy gradient methods: Optimality, approximation, and distribution shift.
\newblock \emph{Journal of Machine Learning Research}, 22\penalty0 (98):\penalty0 1--76, 2021.
\newblock \url{https://jmlr.org/papers/v22/19-736.html}.

\bibitem[Cao et~al.(2026)Cao, Deng, Yu, Zhou, Liu, Ding, and Zhao]{cao2026reme}
Zouying Cao, Jiaji Deng, Li~Yu, Weikang Zhou, Zhaoyang Liu, Bolin Ding, and Hai Zhao.
\newblock Remember me, refine me: A dynamic procedural memory framework for experience-driven agent evolution.
\newblock In \emph{Findings of the Association for Computational Linguistics: ACL 2026}, pages 16803--16822. Association for Computational Linguistics, 2026.
\newblock \doi{10.18653/v1/2026.findings-acl.829}.
\newblock \url{https://aclanthology.org/2026.findings-acl.829/}.

\bibitem[Chen et~al.(2026)Chen, Lu, Zhao, Meng, Teng, Li, Li, Liu, Liang, Zhang, et~al.]{harnessx}
Tingyang Chen, Shuo Lu, Kang Zhao, Weicheng Meng, Hanlin Teng, Tianhao Li, Chao Li, Xule Liu, Jian Liang, Zhizhong Zhang, et~al.
\newblock Harnessx: A composable, adaptive, and evolvable agent harness foundry.
\newblock \emph{arXiv preprint arXiv:2606.14249}, 2026.

\bibitem[Fang et~al.(2026)Fang, Liang, Wang, Wu, Qiao, Xie, Huang, Chen, and Zhang]{fang2026memp}
Runnan Fang, Yuan Liang, Xiaobin Wang, Jialong Wu, Shuofei Qiao, Pengjun Xie, Fei Huang, Huajun Chen, and Ningyu Zhang.
\newblock Memp: Exploring agent procedural memory.
\newblock In \emph{Findings of the Association for Computational Linguistics: ACL 2026}, pages 17490--17502, 2026.

\bibitem[Hong et~al.(2026)Hong, Sun, Li, Yao, Wu, Chien, Yin, Wu, Wang, and Chang]{hong2026embodied}
Yining Hong, Rui Sun, Bingxuan Li, Xingcheng Yao, Maxine Wu, Alexander Chien, Da~Yin, Ying~Nian Wu, Zhecan Wang, and Kai-Wei Chang.
\newblock Embodied web agents: Bridging physical-digital realms for integrated agent intelligence.
\newblock \emph{Advances in Neural Information Processing Systems}, 38, 2026.

\bibitem[Huang et~al.(2026)Huang, Wang, Bao, Ma, Luo, Nian, Zhuang, Liu, Zhao, and Zhang]{huang2026memoharness}
Yue Huang, Wenjie Wang, Han Bao, Yuchen Ma, Xiaonan Luo, Yi~Nian, Haomin Zhuang, Zheyuan Liu, Yue Zhao, and Xiangliang Zhang.
\newblock Memoharness: Agent harnesses that learn from experience.
\newblock \emph{arXiv preprint arXiv:2607.14159}, 2026.

\bibitem[Jiang et~al.(2026)Jiang, Shi, Hong, Xu, Sun, Sun, Bashir, and Han]{harness1}
Pengcheng Jiang, Zhiyi Shi, Kelly Hong, Xueqiang Xu, Jiashuo Sun, Jimeng Sun, Hammad Bashir, and Jiawei Han.
\newblock Harness-1: Reinforcement learning for search agents with state-externalizing harnesses.
\newblock \emph{arXiv preprint arXiv:2606.02373}, 2026.

\bibitem[Ke et~al.(2026)Ke, Patil, Shi, Li, Liu, Shekkizhar, Koul, Wang, Nguyen, Yavuz, et~al.]{ke2026evoharnessbench}
Zixuan Ke, Vaidehi Patil, Haizhou Shi, Yang Li, Ye~Liu, Sarath Shekkizhar, Anurag Koul, Jiayu Wang, Xuan~Phi Nguyen, Semih Yavuz, et~al.
\newblock Evoharnessbench: Can your agents keep pace with an evolving harness?
\newblock \emph{arXiv preprint arXiv:2609.04280}, 2026.

\bibitem[Lee et~al.(2026)Lee, Nair, Zhang, Lee, Khattab, and Finn]{lee2026meta}
Yoonho Lee, Roshen Nair, Qizheng Zhang, Kangwook Lee, Omar Khattab, and Chelsea Finn.
\newblock Meta-harness: End-to-end optimization of model harnesses.
\newblock \emph{arXiv preprint arXiv:2603.28052}, 2026.

\bibitem[Lou et~al.(2026)Lou, L{\'a}zaro-Gredilla, Dedieu, Wendelken, Lehrach, and Murphy]{lou2026autoharness}
Xinghua Lou, Miguel L{\'a}zaro-Gredilla, Antoine Dedieu, Carter Wendelken, Wolfgang Lehrach, and Kevin~P Murphy.
\newblock Autoharness: improving llm agents by automatically synthesizing a code harness.
\newblock \emph{arXiv preprint arXiv:2603.03329}, 2026.

\bibitem[Ning et~al.(2026)Ning, Tieu, Fu, Wei, Li, Bei, Zou, Ai, Liu, Li, et~al.]{ning2026code}
Xuying Ning, Katherine Tieu, Dongqi Fu, Tianxin Wei, Zihao Li, Yuanchen Bei, Jiaru Zou, Mengting Ai, Zhining Liu, Ting-Wei Li, et~al.
\newblock Code as agent harness.
\newblock \emph{arXiv preprint arXiv:2605.18747}, 2026.

\bibitem[Ouyang et~al.(2025)Ouyang, Yan, Hsu, Chen, Jiang, Wang, Han, Le, Daruki, Tang, et~al.]{ouyang2025reasoningbank}
Siru Ouyang, Jun Yan, I~Hsu, Yanfei Chen, Ke~Jiang, Zifeng Wang, Rujun Han, Long~T Le, Samira Daruki, Xiangru Tang, et~al.
\newblock Reasoningbank: Scaling agent self-evolving with reasoning memory.
\newblock \emph{arXiv preprint arXiv:2509.25140}, 2025.

\bibitem[Ouyang et~al.(2026)Ouyang, Yan, Chen, Han, Wang, Mishra, Meng, Li, Jiao, Zha, et~al.]{ouyang2026skillos}
Siru Ouyang, Jun Yan, Yanfei Chen, Rujun Han, Zifeng Wang, Bhavana~Dalvi Mishra, Rui Meng, Chun-Liang Li, Yizhu Jiao, Kaiwen Zha, et~al.
\newblock Skillos: Learning skill curation for self-evolving agents.
\newblock \emph{arXiv preprint arXiv:2605.06614}, 2026.

\bibitem[Schick et~al.(2023)Schick, Dwivedi-Yu, Dess{\`\i}, Raileanu, Lomeli, Hambro, Zettlemoyer, Cancedda, and Scialom]{toolformer}
Timo Schick, Jane Dwivedi-Yu, Roberto Dess{\`\i}, Roberta Raileanu, Maria Lomeli, Eric Hambro, Luke Zettlemoyer, Nicola Cancedda, and Thomas Scialom.
\newblock Toolformer: Language models can teach themselves to use tools.
\newblock \emph{Advances in neural information processing systems}, 36:\penalty0 68539--68551, 2023.

\bibitem[Shao et~al.(2026)Shao, Zhang, Li, Wang, Wang, Jiao, Lu, Guo, Liu, and Zhang]{shao2026harness}
Shuai Shao, Kangning Zhang, Qingyao Li, Shijian Wang, Hao Wang, Wenxiang Jiao, Yuan Lu, Yi~Guo, Weiwen Liu, and Weinan Zhang.
\newblock Harness-r1: Learning to edit executable runtime harnesses from agent failure trajectories.
\newblock \emph{arXiv preprint arXiv:2608.02276}, 2026.

\bibitem[Shao et~al.(2024)Shao, Wang, Zhu, Xu, Song, Bi, Zhang, Zhang, Li, Wu, et~al.]{grpo}
Zhihong Shao, Peiyi Wang, Qihao Zhu, Runxin Xu, Junxiao Song, Xiao Bi, Haowei Zhang, Mingchuan Zhang, YK~Li, Yang Wu, et~al.
\newblock Deepseekmath: Pushing the limits of mathematical reasoning in open language models.
\newblock \emph{arXiv preprint arXiv:2402.03300}, 2024.

\bibitem[Shinn et~al.(2023)Shinn, Cassano, Gopinath, Narasimhan, and Yao]{reflexion}
Noah Shinn, Federico Cassano, Ashwin Gopinath, Karthik Narasimhan, and Shunyu Yao.
\newblock Reflexion: Language agents with verbal reinforcement learning.
\newblock \emph{Advances in neural information processing systems}, 36:\penalty0 8634--8652, 2023.

\bibitem[Shridhar et~al.(2021)Shridhar, Yuan, C{\^o}t{\'e}, Bisk, Trischler, and Hausknecht]{alfworld}
Mohit Shridhar, Xingdi Yuan, Marc-Alexandre C{\^o}t{\'e}, Yonatan Bisk, Adam Trischler, and Matthew Hausknecht.
\newblock {ALFWorld}: Aligning text and embodied environments for interactive learning.
\newblock In \emph{International Conference on Learning Representations}, 2021.
\newblock \url{https://openreview.net/forum?id=0IOX0YcCdTn}.

\bibitem[Singh et~al.(2026)Singh, Khan, Prasad, Chen, Nambi, Lee, Stengel-Eskin, and Bansal]{singh2026agent}
Joykirat Singh, Zaid Khan, Archiki Prasad, Justin Chih-Yao Chen, Akshay Nambi, Hyunji Lee, Elias Stengel-Eskin, and Mohit Bansal.
\newblock Agent-brace: Decoupling beliefs from actions in long-horizon tasks via verbalized state uncertainty.
\newblock \emph{arXiv preprint arXiv:2605.11436}, 2026.

\bibitem[Suzgun et~al.(2026)Suzgun, Yuksekgonul, Bianchi, Jurafsky, and Zou]{suzgun2026dynamic}
Mirac Suzgun, Mert Yuksekgonul, Federico Bianchi, Dan Jurafsky, and James Zou.
\newblock Dynamic cheatsheet: Test-time learning with adaptive memory.
\newblock In \emph{Proceedings of the 19th Conference of the European Chapter of the Association for Computational Linguistics (Volume 1: Long Papers)}, pages 7080--7106, 2026.

\bibitem[Wang et~al.(2024)Wang, Xie, Jiang, Mandlekar, Xiao, Zhu, Fan, and Anandkumar]{voyager}
Guanzhi Wang, Yuqi Xie, Yunfan Jiang, Ajay Mandlekar, Chaowei Xiao, Yuke Zhu, Linxi Fan, and Anima Anandkumar.
\newblock Voyager: An open-ended embodied agent with large language models.
\newblock \emph{Transactions on Machine Learning Research}, 2024.
\newblock ISSN 2835-8856.
\newblock \url{https://openreview.net/forum?id=ehfRiF0R3a}.

\bibitem[Wang et~al.(2026)Wang, Gooding, Hartmann, Riva, and Grefenstette]{wang2026subgoal}
Taiyi Wang, Sian Gooding, Florian Hartmann, Oriana Riva, and Edward Grefenstette.
\newblock A subgoal-driven framework for improving long-horizon llm agents.
\newblock \emph{arXiv preprint arXiv:2603.19685}, 2026.

\bibitem[Wang et~al.(2025)Wang, Mao, Fried, and Neubig]{wang2025agent}
Zora~Zhiruo Wang, Jiayuan Mao, Daniel Fried, and Graham Neubig.
\newblock Agent workflow memory.
\newblock In \emph{International Conference on Machine Learning}, pages 63897--63911. PMLR, 2025.

\bibitem[Wei et~al.(2025)Wei, Sachdeva, Coleman, He, Bei, Ning, Ai, Li, He, Chi, et~al.]{wei2025evo}
Tianxin Wei, Noveen Sachdeva, Benjamin Coleman, Zhankui He, Yuanchen Bei, Xuying Ning, Mengting Ai, Yunzhe Li, Jingrui He, Ed~H Chi, et~al.
\newblock Evo-memory: Benchmarking llm agent test-time learning with self-evolving memory.
\newblock \emph{arXiv preprint arXiv:2511.20857}, 2025.

\bibitem[Wu et~al.(2026)Wu, Zhang, Shi, Lei, Gu, Zhang, Wang, He, Huang, Song, et~al.]{wu2026harnessdev}
Yuhao Wu, Jingyuan Zhang, Jiajun Shi, Xinping Lei, Qingshui Gu, Yuxuan Zhang, Zexuan Wang, Chen He, Chen Huang, Maojia Song, et~al.
\newblock Harnessdev: Can llms create and evolve their own agent harness?
\newblock \emph{arXiv preprint arXiv:2609.01437}, 2026.

\bibitem[Xi et~al.(2025)Xi, Huang, Liao, Huang, Guo, Liu, Zheng, Ye, Zhang, Chen, et~al.]{xi2025agentgymrl}
Zhiheng Xi, Jixuan Huang, Chenyang Liao, Baodai Huang, Honglin Guo, Jiaqi Liu, Rui Zheng, Junjie Ye, Jiazheng Zhang, Wenxiang Chen, et~al.
\newblock Agentgym-rl: Training llm agents for long-horizon decision making through multi-turn reinforcement learning.
\newblock \emph{arXiv preprint arXiv:2509.08755}, 2025.

\bibitem[Xia et~al.(2026)Xia, Chen, Wang, Liu, Zeng, Wang, Han, Zhou, Zhao, Chen, et~al.]{xia2026skillrl}
Peng Xia, Jianwen Chen, Hanyang Wang, Jiaqi Liu, Kaide Zeng, Yu~Wang, Siwei Han, Yiyang Zhou, Xujiang Zhao, Haifeng Chen, et~al.
\newblock Skillrl: Evolving agents via recursive skill-augmented reinforcement learning.
\newblock \emph{arXiv preprint arXiv:2602.08234}, 2026.

\bibitem[Xiao(2022)]{xiao2022policygradient}
Lin Xiao.
\newblock On the convergence rates of policy gradient methods.
\newblock \emph{Journal of Machine Learning Research}, 23\penalty0 (282):\penalty0 1--36, 2022.
\newblock \url{https://jmlr.org/papers/v23/22-0056.html}.

\bibitem[Xie et~al.(2022)Xie, Raghunathan, Liang, and Ma]{xie2022implicitbayes}
Sang~Michael Xie, Aditi Raghunathan, Percy Liang, and Tengyu Ma.
\newblock An explanation of in-context learning as implicit {Bayesian} inference.
\newblock In \emph{International Conference on Learning Representations}, 2022.
\newblock \url{https://arxiv.org/abs/2111.02080}.

\bibitem[Yang et~al.(2024)Yang, Jimenez, Wettig, Lieret, Yao, Narasimhan, and Press]{sweagent}
John Yang, Carlos Jimenez, Alexander Wettig, Kilian Lieret, Shunyu Yao, Karthik Narasimhan, and Ofir Press.
\newblock Swe-agent: Agent-computer interfaces enable automated software engineering.
\newblock \emph{Advances in Neural Information Processing Systems}, 37:\penalty0 50528--50652, 2024.

\bibitem[Yao et~al.(2022{\natexlab{a}})Yao, Chen, Yang, and Narasimhan]{yao2022webshop}
Shunyu Yao, Howard Chen, John Yang, and Karthik Narasimhan.
\newblock {WebShop}: Towards scalable real-world web interaction with grounded language agents.
\newblock In \emph{Advances in Neural Information Processing Systems}, 2022{\natexlab{a}}.

\bibitem[Yao et~al.(2022{\natexlab{b}})Yao, Zhao, Yu, Du, Shafran, Narasimhan, and Cao]{react}
Shunyu Yao, Jeffrey Zhao, Dian Yu, Nan Du, Izhak Shafran, Karthik Narasimhan, and Yuan Cao.
\newblock React: Synergizing reasoning and acting in language models.
\newblock \emph{arXiv preprint arXiv:2210.03629}, 2022{\natexlab{b}}.

\bibitem[Young(2025)]{anthropic2025longrunning}
Justin Young.
\newblock Effective harnesses for long-running agents.
\newblock Anthropic Engineering Blog, November 2025.
\newblock \url{https://www.anthropic.com/engineering/effective-harnesses-for-long-running-agents}.
\newblock Accessed: 2026-05-11.

\bibitem[Zhang et~al.(2026)Zhang, Zhang, Li, Zhang, Chen, Zhang, Bai, and Hu]{zhang2026self}
Hangfan Zhang, Shao Zhang, Kangcong Li, Chen Zhang, Yang Chen, Yiqun Zhang, Lei Bai, and Shuyue Hu.
\newblock Self-harness: Harnesses that improve themselves.
\newblock \emph{arXiv preprint arXiv:2606.09498}, 2026.

\bibitem[Zhang et~al.(2025{\natexlab{a}})Zhang, Hu, Upasani, Ma, Hong, Kamanuru, Rainton, Wu, Ji, Li, et~al.]{zhang2025agentic}
Qizheng Zhang, Changran Hu, Shubhangi Upasani, Boyuan Ma, Fenglu Hong, Vamsidhar Kamanuru, Jay Rainton, Chen Wu, Mengmeng Ji, Hanchen Li, et~al.
\newblock Agentic context engineering: Evolving contexts for self-improving language models.
\newblock \emph{arXiv preprint arXiv:2510.04618}, 2025{\natexlab{a}}.

\bibitem[Zhang et~al.(2025{\natexlab{b}})Zhang, Zhang, Yang, and Wang]{zhang2025icltheory}
Yufeng Zhang, Fengzhuo Zhang, Zhuoran Yang, and Zhaoran Wang.
\newblock What and how does in-context learning learn? {Bayesian} model averaging, parameterization, and generalization.
\newblock In \emph{Proceedings of the 28th International Conference on Artificial Intelligence and Statistics}, volume 258 of \emph{Proceedings of Machine Learning Research}, pages 1684--1692. PMLR, 2025{\natexlab{b}}.
\newblock \url{https://proceedings.mlr.press/v258/zhang25d.html}.

\bibitem[Zhao et~al.(2024)Zhao, Huang, Xu, Lin, Liu, and Huang]{zhao2024expel}
Andrew Zhao, Daniel Huang, Quentin Xu, Matthieu Lin, Yong-Jin Liu, and Gao Huang.
\newblock Expel: Llm agents are experiential learners.
\newblock In \emph{Proceedings of the AAAI Conference on Artificial Intelligence}, volume~38, pages 19632--19642, 2024.

\bibitem[Zhou et~al.(2024)Zhou, Xu, Zhu, Zhou, Lo, Sridhar, Cheng, Ou, Bisk, Fried, et~al.]{zhou2024webarena}
Shuyan Zhou, Frank~F Xu, Hao Zhu, Xuhui Zhou, Robert Lo, Abishek Sridhar, Xianyi Cheng, Tianyue Ou, Yonatan Bisk, Daniel Fried, et~al.
\newblock Webarena: A realistic web environment for building autonomous agents.
\newblock In \emph{International Conference on Learning Representations}, volume 2024, pages 15585--15606, 2024.

\end{thebibliography}

\newpage
\clearpage
\beginappendix

\section*{Appendix Contents}
\phantomsection
\addcontentsline{toc}{section}{Appendix Contents}

\begingroup
\hypersetup{linkcolor=paperskyblue}
\setlength{\parindent}{0pt}
\newcommand{\appcontentsentry}[3]{%
  \vspace{0.55em}%
  \noindent\hyperref[#1]{\textbf{\ref*{#1}\quad #2}}%
  \dotfill\hyperref[#1]{\pageref*{#1}}\par
  \vspace{0.12em}%
  {\small\color{black!65}\hspace*{2.15em}#3\par}%
}

\appcontentsentry{app:related_work}{Related Work}
{Extended discussion of harness engineering and self-evolving agents.}
\appcontentsentry{app:training_dynamics}{Training Dynamics}
{Harness annealing and optimization behavior across benchmarks.}
\appcontentsentry{app:cross_domain_transfer}{Cross-Domain Transferability}
{Adaptation from WebShop to ALFWorld under different initialization strategies.}
\appcontentsentry{app:cross_backbone_adaptability}{Cross-Backbone Adaptability}
{Results with Qwen3-4B and Qwen3-8B policy backbones.}
\appcontentsentry{app:seed_robustness}{Method Robustness Across Random Seeds}
{Three-seed performance on ALFWorld, WebShop, and WebArena.}
\appcontentsentry{app:theory}{Theoretical Analysis}
{Rollout efficiency from BPE information, convergence under diversity
annealing, and the scope of both results.}
\appcontentsentry{app:more_bpe_ablations}{More Ablation Analysis}
{Harness-schedule, component-removal, and prompt-stripping studies.}
\appcontentsentry{app:case_study}{Qualitative Case Study}
{Within-episode coordination and cross-episode Experience evolution.}
\appcontentsentry{app:environment_grounding}{Environment-Specific BPE Grounding}
{Concrete Belief, Progress, and Experience representations.}
\appcontentsentry{app:baseline_details}{Baseline Details}
{Implementation notes for the compared agent baselines.}
\appcontentsentry{app:implementation_details}{Implementation Details}
{Models, training settings, horizons, and reward hyperparameters.}
\appcontentsentry{app:prompts}{Prompts}
{Runtime harness, message-format, and Experience-consolidation prompts.}
\endgroup
\hypersetup{linkcolor=red}

\clearpage

\section{Related Work}
\label{app:related_work}

Our work is closely related to two lines of research: harness engineering, which studies the external systems that support agent execution, and memory and self-evolving agents, which externalize and adapt task-relevant context across interactions.

\paragraph{Harness engineering.}
The execution capability of an LLM agent is strongly shaped by its external
harness, which determines how observations are represented, how actions and
tools are exposed, and what execution feedback is returned to the policy. Early
agent paradigms established explicit reasoning and action loops and tool
augmented interaction~\citep{react, toolformer}, while subsequent systems
introduced increasingly structured execution interfaces, including specialized
observation renderers, file system and code access, environment specific action
spaces, and execution feedback~\citep{sweagent, zhou2024webarena,
ning2026code}. As agent tasks become longer and more complex, recent work has
increasingly studied how these surrounding systems can themselves be optimized
or evolved. Harness-1 externalizes intermediate search state into environment
side memory~\citep{harness1}, while Meta-Harness searches over harness code and
configurations using execution traces~\citep{lee2026meta}. HarnessX studies
composable and trace driven harness adaptation~\citep{harnessx}, and recent
approaches such as Self-harness and Harness-R1 improve executable harnesses from
model specific weaknesses or failure trajectories~\citep{zhang2026self,
shao2026harness}. MemoHarness further studies experience conditioned harness
adaptation~\citep{huang2026memoharness}, while \textsc{EvoHarnessBench} and
HarnessDev study how agents cope with, construct, and evolve changing
harnesses~\citep{ke2026evoharnessbench, wu2026harnessdev}. These approaches
primarily optimize, synthesize, or evaluate \emph{what harness is provided} to
an agent. In contrast, \methodname{} abstracts heterogeneous harness mechanisms
into a shared policy facing interface and focuses on \emph{runtime agent and
harness coordination}, allowing the acting policy to learn when different forms
of external support should be accessed, updated, or bypassed jointly with
environment execution.

\paragraph{Memory and self-evolving agents.}
External memory provides a natural form of context externalization for long horizon agents, allowing information that would otherwise need to be retained or reconstructed within the model context to persist across interactions. Early memory augmented agents reuse trajectories, reflections, and induced insights across episodes, as in Reflexion and ExpeL~\citep{reflexion,zhao2024expel}. Later work increasingly structures such experience into reusable procedural knowledge, including executable skills, workflows, reasoning items, and procedural scripts~\citep{voyager,wang2025agent,ouyang2025reasoningbank,fang2026memp}. More recent systems treat the external context itself as an evolving resource: Dynamic Cheatsheet and ACE continually refine compact memories or structured playbooks~\citep{suzgun2026dynamic,zhang2025agentic}, while SkillOS and SkillRL learn to curate or co-evolve external skill repositories together with agent execution~\citep{ouyang2026skillos,xia2026skillrl}. These methods demonstrate the value of externalizing and evolving context, but primarily focus on how memory is represented, retrieved, updated, or curated within particular settings. In contrast, \methodname{} abstracts heterogeneous forms of externalized context into a shared policy facing interface, organizing within episode Belief and Progress together with cross episode Experience while leaving their concrete implementations environment specific. Our focus is on learning how the acting policy coordinates with these external states at runtime, and on analyzing how this coordination changes during training as different forms of support are increasingly handled internally or remain useful to access externally.

\section{Training Dynamics}
\label{app:training_dynamics}

We provide additional analyses of learned agent--harness coordination by
comparing training-success dynamics and harness use across environments.

\subsection{Environment-Dependent Harness Dynamics}
\begin{figure}[H]
    \centering
    \includegraphics[width=\textwidth]{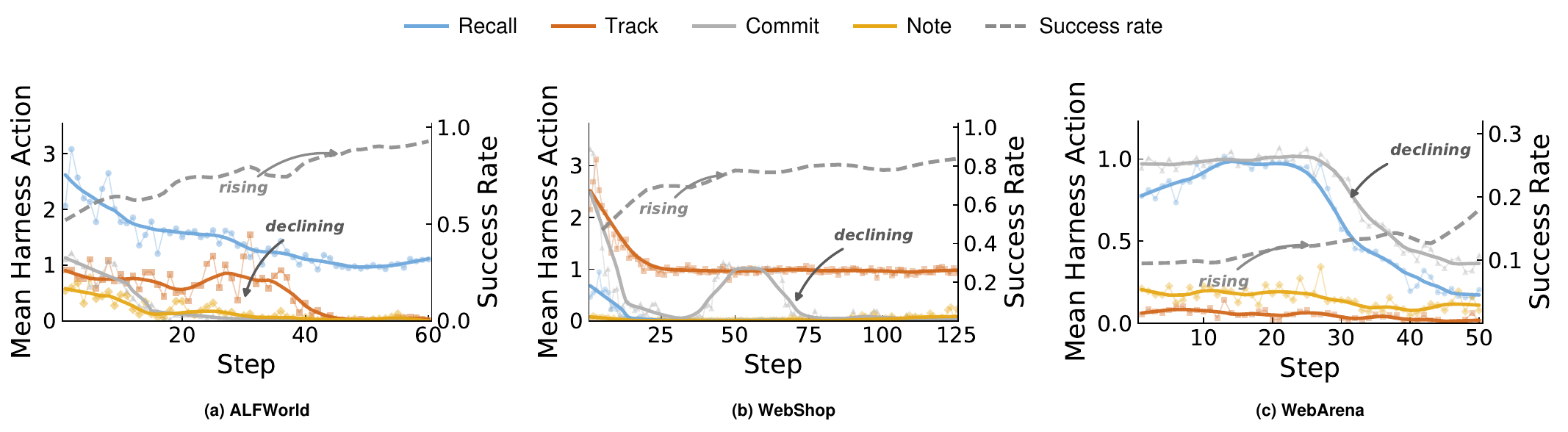}
    \caption{\small Environment-dependent harness dynamics across benchmarks.}
    \label{fig:cross_env_annealing}
    \vspace{-0.3cm}
\end{figure}

Figure~\ref{fig:cross_env_annealing} shows that harness use decreases across
all three environments, but different operations are retained because the
environments impose different information demands. In ALFWorld, tasks
repeatedly reuse household procedures and object-location priors across
episodes. This makes recalled Experience persist throughout training, while
explicit Progress and newly written notes become less necessary once recurring
manipulation patterns are absorbed by the policy. WebShop has a different
structure: each episode introduces a new combination of product attributes and
purchase constraints, so cross-episode retrieval is less directly useful,
whereas maintaining the current request, candidate attributes, and verification
status remains important. Consequently, Track is retained more strongly as an
episode-specific state mechanism. WebArena is both more heterogeneous and more
workflow dependent. Tasks span different websites and multi-stage interactions
whose relevant state and subgoals change across pages, producing a more mixed
pattern rather than a single dominant harness operation. Recall is useful early
for reusable interaction routines but declines substantially as common
navigation behaviors are learned, while state- and progress-related operations
remain useful for coordinating instance-specific workflows.

These differences suggest that annealing is not simply uniform sparsification
of harness calls. Instead, training appears to preserve external access when
information remains difficult to encode parametrically or varies substantially
across episodes, while recurring and predictable coordination patterns are
increasingly handled by the policy itself. The resulting retention pattern
therefore reflects the interaction between task regularity, cross-episode
reusability, and within-episode state dependence.

\subsection{Training and Rollout Efficiency}
\label{app:rollout_efficiency}

\ifdefined\MetaLayout
\suppressfloats[t]
\begin{figure}[t]
    \centering
    \begin{minipage}[t]{0.47\textwidth}
        \vspace{0pt}\centering
        \includegraphics[width=0.86\linewidth]{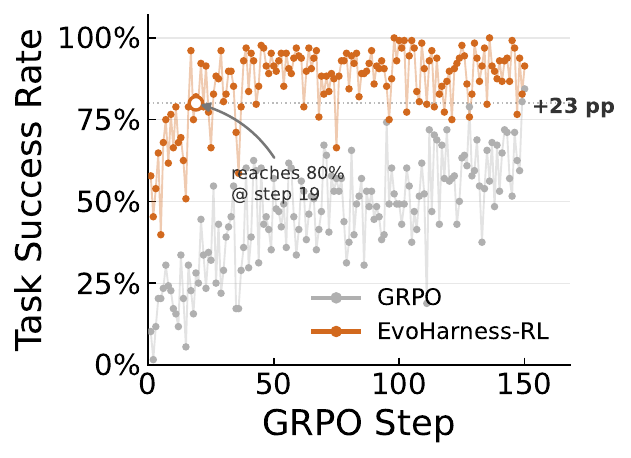}
        \caption{\small ALFWorld training-success dynamics for GRPO and \methodname.}
        \label{fig:training_dynamics_comparison}
    \end{minipage}\hfill
    \begin{minipage}[t]{0.49\textwidth}
        \vspace{0pt}\centering
        \includegraphics[width=0.94\linewidth]{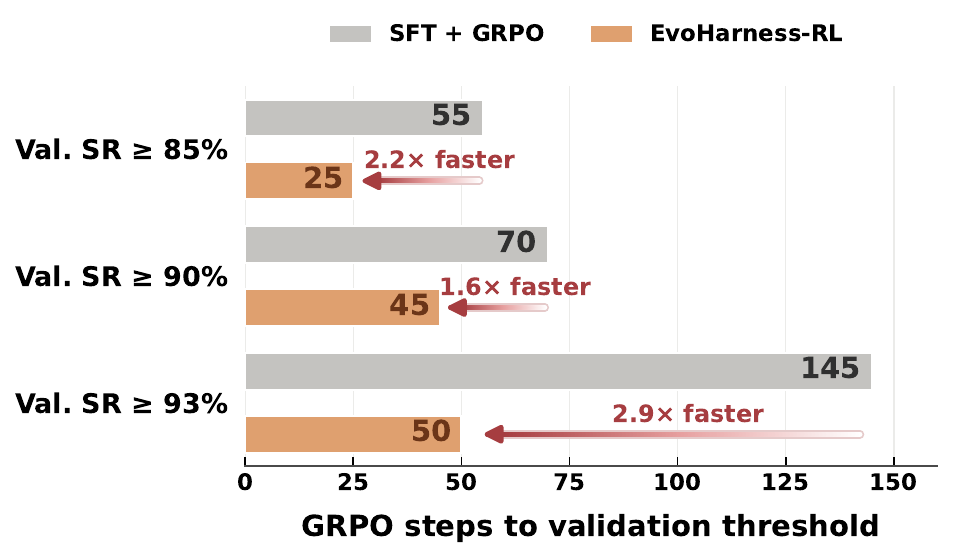}
        \caption{\small GRPO steps to reach ALFWorld validation-set success
        thresholds, evaluated every five training steps.}
        \label{fig:validation_sample_efficiency}
    \end{minipage}
\end{figure}
\else
\begin{wrapfigure}[11]{r}{0.40\textwidth}
    \vspace{-1.2em}
    \centering
    \includegraphics[width=0.94\linewidth]{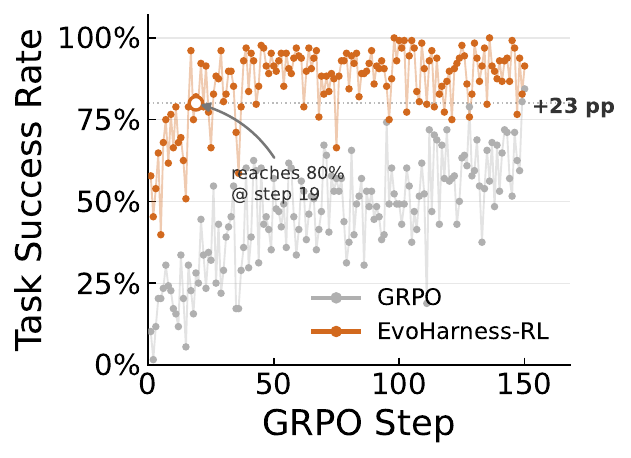}
    \vspace{-0.8em}
    \caption{
    \small ALFWorld training-success dynamics for GRPO and \methodname.
    }
    \label{fig:training_dynamics_comparison}
    \vspace{-0.8em}
\end{wrapfigure}
\fi

Figure~\ref{fig:training_dynamics_comparison} compares GRPO with the full
\methodname{} training pipeline on ALFWorld. \methodname{} enters the
high-success regime substantially earlier, with its smoothed success rate first
reaching 80\% at step 19, while GRPO improves more slowly and remains more
variable. Over the final ten training steps, \methodname{} achieves an average
success rate approximately 23 percentage points higher than GRPO. The earlier
rise and stronger tail performance indicate that the harness-aware pipeline
extracts useful learning signal more efficiently from the rollout budget.

\ifdefined\MetaLayout\else
\begin{wrapfigure}[13]{r}{0.43\textwidth}
    \vspace{-1.0em}
    \centering
    \includegraphics[width=0.98\linewidth]{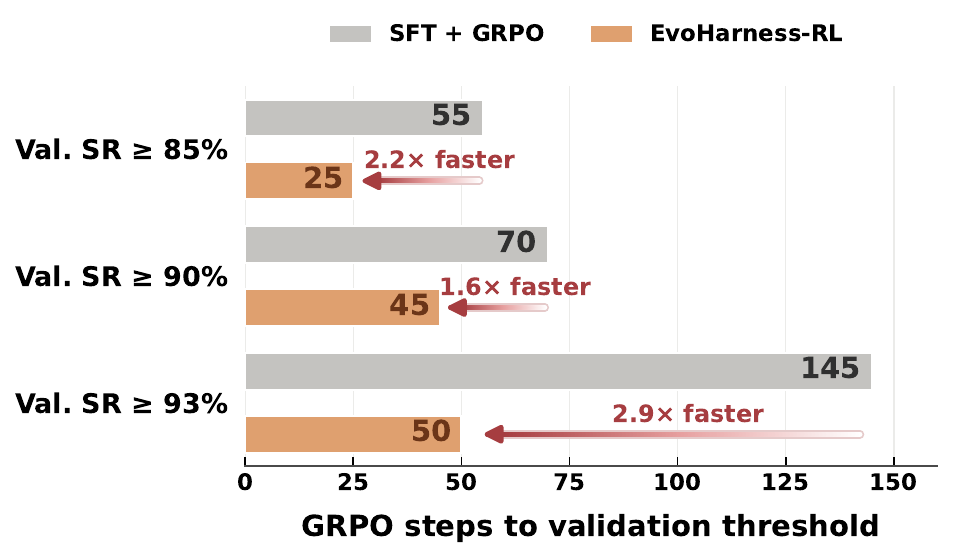}
    \vspace{-0.8em}
    \caption{\small GRPO steps to reach ALFWorld validation-set success
    thresholds, evaluated every five training steps.}
    \label{fig:validation_sample_efficiency}
    \vspace{-0.2em}
\end{wrapfigure}
\fi

To separate this gain from SFT initialization alone,
Figure~\ref{fig:validation_sample_efficiency} compares against SFT + GRPO under
the same supervised-data budget. The figure reports performance on the
validation set during training, evaluated every five GRPO steps. At the first
evaluation, SFT + GRPO has the higher validation success rate (79.7\% versus
57.8\%). This initial gap reflects
a harder supervised-learning problem for \methodname{}: with the same number of
examples, SFT + GRPO only needs to imitate environment-action patterns, whereas
our policy must jointly learn those task-solving patterns, the semantics and
syntax of four additional harness actions, and when each action is worth
invoking. The resulting policy therefore begins RL with less capacity devoted
to direct task execution.

The ordering reverses as optimization proceeds. Once the harness-action
interface is established, the BPE scaffold gives GRPO structured access to
belief, progress, and reusable experience. This reduces repeated state
reconstruction and unproductive exploration, while making successful
coordination patterns easier to reinforce. Consequently, \methodname{} reaches
85\%, 90\%, and 93\% validation success with 2.2$\times$, 1.6$\times$, and
2.9$\times$ speedups in GRPO steps, respectively. The lower initial success but
faster subsequent improvement shows that the gain is not a consequence of a
stronger SFT checkpoint alone; rather, the learned scaffold makes downstream
GRPO substantially more sample efficient.

Proposition~\ref{prop:conditional-efficiency} provides a conditional mechanism
for this empirical trend: useful BPE information can reduce the rollout
budget needed to discover a successful trajectory at a fixed policy
checkpoint. The training curves additionally capture policy updates and
experience-bank evolution.

\subsection{Reward-Shaping Dynamics}
\label{app:reward_analysis}

\begin{figure}[H]
    \centering
    \includegraphics[width=0.94\textwidth]{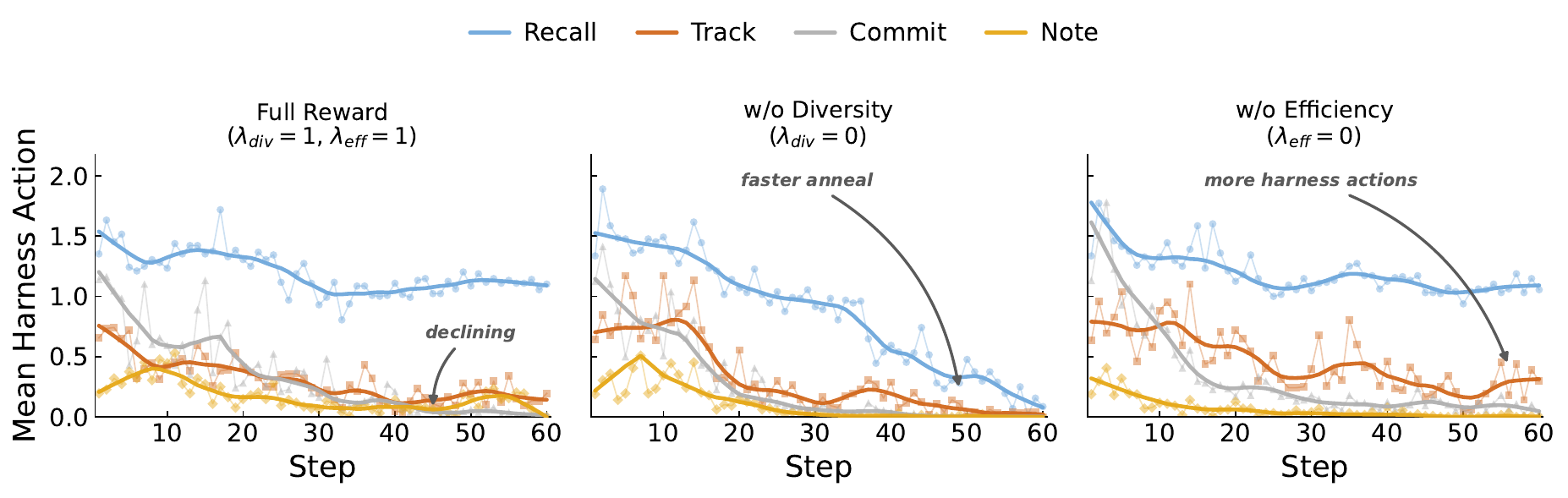}
    \caption{\small
    Harness-action dynamics under different reward designs. Without diversity,
    harness use anneals more quickly; without efficiency, it remains higher
    later in training.
    }
    \label{fig:harness_action_analysis}
    \vspace{-0.5em}
\end{figure}

Figure~\ref{fig:harness_action_analysis} explains how the two auxiliary rewards
produce the performance differences in Table~\ref{tab:ablation}. Without the
diversity reward, aggregate BPE use falls earlier and more rapidly. We interpret
this decline as premature pruning rather than successful specialization: the
policy receives less incentive to try newly introduced harness operations
before it has gathered enough evidence about their utility. The diversity term
therefore acts as a temporary exploration mechanism that keeps multiple
coordination strategies available early in training.

Without the efficiency reward, harness use remains elevated later in training.
This indicates that task reward alone can preserve useful external operations,
but provides little pressure to distinguish necessary calls from redundant
ones. The efficiency term supplies this pressure by rewarding successful
trajectories that solve the task with fewer total interactions. Together, the
two rewards play complementary temporal roles: diversity broadens early
exploration, whereas efficiency converts that exploration into selective
harness use as training progresses. This interpretation also complements
Theorem~\ref{prop:annealed-diversity}: the bounded diversity bonus adds at
most $\lambda_u$ to the final cost-aware objective gap and vanishes from the
objective after annealing. This does not rule out optimization-path effects
from early exploration.

\paragraph{The early role of \textsc{Note}.}
Although the evolver receives complete trajectories, useful discoveries and
corrections can be difficult to identify in long interaction histories.
\textsc{Note} lets the acting policy mark this evidence when it is most salient,
giving consolidation a sparse signal about what should be added or revised.
Its relatively frequent use early in training therefore helps the initially
empty Experience bank accumulate useful knowledge quickly. As the bank matures
and efficiency pressure increases, \textsc{Note} calls decline, while the
Experience distilled from earlier notes remains available for later retrieval.
Thus, even though \textsc{Note} is invoked less often later in training, its
more frequent early use still provides important support for bootstrapping the
Experience bank used in subsequent coordination.

\subsection{From Interface Acquisition to Selective Coordination}
\begin{figure}[H]
    \centering
    \includegraphics[width=0.75\textwidth]{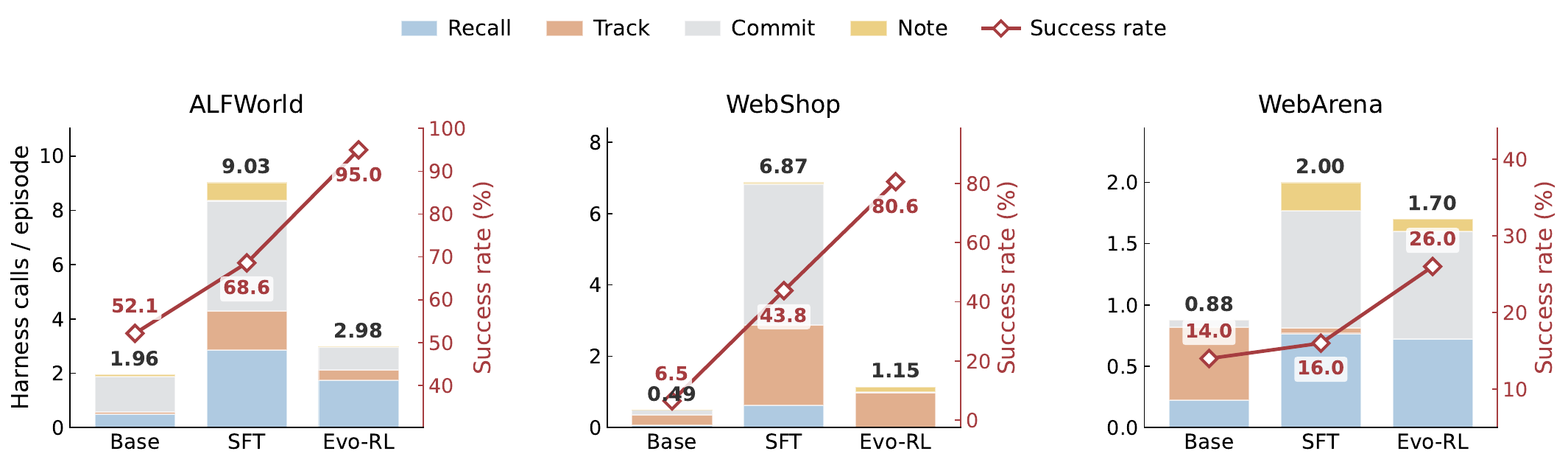}
    \caption{\small Average harness calls per test episode and test-set success
    rates across the three benchmarks.}
    \label{fig:harness_usage_stage}
    \vspace{-0.3cm}
\end{figure}

Figure~\ref{fig:harness_usage_stage} compares harness usage before and after
coordination training. Here, Qwen3-8B \textsc{Base} denotes the pretrained
policy placed directly in the adaptive BPE action interface, rather than the
always-on \textsc{EvoHarness-Base} scaffold. Despite having access to the interface, the
base policy invokes harness actions only sparsely, suggesting that prompting
alone is insufficient for reliably operating the newly introduced meta-action
space. SFT addresses this cold-start problem by imitating successful teacher
trajectories in which the BPE operations are explicitly demonstrated,
substantially increasing harness use together with task success. After this
interface has been acquired, reinforcement learning reverses the trend:
\textsc{Evoharness-RL} (Evo-RL) reduces the number of harness calls while success continues to
improve across all three environments. The resulting pattern suggests that SFT
primarily teaches the policy \emph{how to use} the harness, whereas RL learns
\emph{when its use is worth the interaction cost}, pruning redundant calls
while retaining task-relevant external support.

\section{Cross-Domain Transferability}
\label{app:cross_domain_transfer}

\begin{figure}[H]
    \centering
    \includegraphics[width=0.75\textwidth]{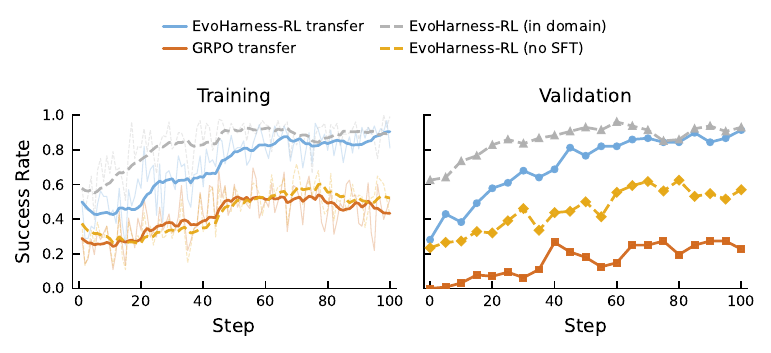}
    \caption{Cross-domain adaptation from WebShop to ALFWorld over 100
    target-domain training steps. Left: training success. Right: validation
    success.}
    \label{fig:transfer_curve}
\end{figure}

We further test whether the coordination learned by \methodname{} transfers
across environments by initializing ALFWorld training from a WebShop-trained
checkpoint and adapting for 100 target-domain steps. As shown in
Figure~\ref{fig:transfer_curve}, \methodname{} transfer adapts rapidly and
progressively approaches the in-domain \methodname{} policy on both training
and validation performance. In contrast, transferring a standard GRPO
checkpoint exhibits strong negative transfer: training performance partially
recovers, but validation success remains substantially lower throughout
adaptation. Removing target-side SFT also degrades transfer, indicating that
interface initialization still facilitates adaptation. Together, these results
suggest that \methodname{} learns coordination patterns that are partially
reusable across domains, providing a better initialization than transferring
environment-specific policy behavior alone.

Interestingly, cross-domain transfer does not imply that the same harness-use
pattern should be preserved across environments. Our earlier analysis shows
that WebShop and ALFWorld converge to different mixtures of BPE operations.
What appears to transfer is instead the underlying coordination policy: how the
agent decides whether external support is useful under the current execution
state. The observed action frequency is environment dependent, since different
domains induce different state and information demands, while the conditional
decision rule for when to externalize state can remain reusable. This helps
explain why \methodname{} exhibits substantially less negative transfer than
standard GRPO: it transfers a reusable coordination prior rather than a fixed
source-domain action pattern.

\section{Cross-Backbone Adaptability}
\label{app:cross_backbone_adaptability}

We evaluate whether our framework remains effective across backbone scales.
Table~\ref{tab:cross_backbone_adaptability} reports selected results with
Qwen3-4B on ALFWorld, WebShop, and WebArena, alongside Qwen3-8B
\textsc{EvoHarness-RL} as a reference. This comparison tests the adaptability
of our harness-learning framework beyond the 8B backbone used in the main
experiments. Unreported results are marked with an em dash.

\begin{table}[H]
\centering
\small
\setlength{\tabcolsep}{4.2pt}
\renewcommand{\arraystretch}{1.08}
\caption{\small Cross-backbone adaptability across ALFWorld, WebShop, and WebArena.}
\label{tab:cross_backbone_adaptability}
\resizebox{\textwidth}{!}{
\begin{tabular}{c l*{14}{r}}
\toprule
\multirow{2}{*}{\textbf{Backbone}}
& \multirow{2}{*}{\textbf{Approach}}
& \multicolumn{7}{c}{\textbf{ALFWorld}}
& \multicolumn{2}{c}{\textbf{WebShop}}
& \multicolumn{5}{c}{\textbf{WebArena}} \\
\cmidrule(lr){3-9}
\cmidrule(lr){10-11}
\cmidrule(lr){12-16}
& & \textbf{Pick}
& \textbf{Look}
& \textbf{Clean}
& \textbf{Heat}
& \textbf{Cool}
& \textbf{Pick2}
& \textbf{Avg.}
& \textbf{SR}
& \textbf{Score}
& \textbf{SP}
& \textbf{CMS}
& \textbf{Maps}
& \textbf{G\&R}
& \textbf{Overall} \\
\midrule
\multirow{5}{*}{\textbf{4B}}
& ReAct\frozen
& 62.9 & 61.5 & 11.1 & 62.5 & 8.0 & 8.3 & 33.6
& 3.2 & 39.5
& 13.3 & 6.7 & 10.0 & 0.0 & 8.0 \\

& SkillRL\trainable
& 94.3 & 69.2 & 85.2 & 68.8 & 92.0 & 87.5 & 85.7
& 74.4 & 84.0
& 26.7 & 20.0 & 0.0 & 10.0 & 16.0 \\

& \textsc{EvoHarness-Base}\frozen
& 85.7 & 30.8 & 51.9 & 50.0 & 36.0 & 33.3 & 52.1
& 12.6 & 28.7
& 20.0 & 6.7 & 20.0 & 30.0 & 18.0 \\

& \textsc{EvoHarness-SFT}\trainable
& 57.1 & 7.7 & 37.0 & 12.5 & 12.0 & 16.7 & 28.6
& 39.0 & 46.3
& 20.0 & 6.7 & 0.0 & 10.0 & 10.0 \\

& \textsc{EvoHarness-RL}\trainable
& 100.0 & 61.5 & 92.6 & 81.3 & 80.0 & 91.7 & 87.9
& 79.0 & 89.0
& 20.0 & 20.0 & 10.0 & 20.0 & 18.0 \\

\midrule
\textbf{8B}
& \textsc{EvoHarness-RL}\trainable
& \textbf{100.0} & \textbf{92.3} & \textbf{96.3} & \textbf{100.0}
& \textbf{84.0} & \textbf{95.8} & \textbf{95.0}
& \textbf{80.6} & \textbf{90.8}
& \textbf{40.0} & \textbf{20.0} & \textbf{10.0} & \textbf{30.0}
& \textbf{26.0} \\
\bottomrule
\end{tabular}
}
\end{table}

\section{Method Robustness Across Random Seeds}
\label{app:seed_robustness}

Table~\ref{tab:main_results} reports a single evaluation run, whereas here we
evaluate \textsc{EvoHarness-RL} across three random seeds and report the mean
and population standard deviation in Table~\ref{tab:seed_robustness}. Randomness
has little effect on ALFWorld, where every run solves 133 of 140 tasks, or on
WebShop, which achieves $80.5\pm0.7\%$ success and an average score of
$89.8\pm0.9$. WebArena exhibits greater variation partly because the
standardized AgentGym-RL split is comparatively small: it contains 372 training
tasks and only 50 evaluation tasks, divided into Shopping (15), CMS (15), Maps
(10), GitLab (5), and Reddit (5). A single changed outcome can therefore move a
domain score by 6.7--20 percentage points, making the aggregate result over all
50 tasks more informative than any individual category. The three overall
scores are 20.0\%, 22.0\%, and 26.0\%, yielding $22.7\pm2.5\%$. The gains on
WebArena relative to ReAct also vary across task categories: they are strongest
on Shopping and GitLab/Reddit, while CMS and Maps benefit less. The latter tasks depend more on
precise site-specific action execution and current-page grounding, leaving less
room for persistent state and reusable Experience to help. Because of time and
compute constraints, we run three seeds only for \textsc{EvoHarness-RL}; the
comparison baselines reproduce the point estimates from
Table~\ref{tab:main_results}. Nevertheless, even the mean minus one standard
deviation for our method exceeds the strongest listed baseline in 11 of the 14
reported columns.

\begin{table}[H]
\centering
\scriptsize
\setlength{\tabcolsep}{3.0pt}
\renewcommand{\arraystretch}{1.12}
\newcommand{\seedstat}[2]{\shortstack{\textbf{#1}\\[-0.25em]{\tiny $\pm\,#2$}}}
\newcommand{\seedstatplain}[2]{\shortstack{#1\\[-0.25em]{\tiny $\pm\,#2$}}}
\caption{Robustness across three random seeds. Baselines reproduce the point
estimates from Table~\ref{tab:main_results}; \textsc{EvoHarness-RL} reports
mean $\pm$ population standard deviation over three runs. All values are percentages
except WebShop Score. Bold indicates the best result in each column.}
\label{tab:seed_robustness}
\resizebox{\textwidth}{!}{%
\begin{tabular}{l*{14}{r}}
\toprule
\multirow{2}{*}{\textbf{Approach}}
& \multicolumn{7}{c}{\textbf{ALFWorld}}
& \multicolumn{2}{c}{\textbf{WebShop}}
& \multicolumn{5}{c}{\textbf{WebArena}} \\
\cmidrule(lr){2-8}
\cmidrule(lr){9-10}
\cmidrule(lr){11-15}
& \textbf{Pick} & \textbf{Look} & \textbf{Clean} & \textbf{Heat}
& \textbf{Cool} & \textbf{Pick2} & \textbf{Avg.}
& \textbf{SR} & \textbf{Score}
& \textbf{SP} & \textbf{CMS} & \textbf{Maps} & \textbf{G\&R}
& \textbf{Avg.} \\
\midrule
ReasoningBank\frozen
& 83.8 & 48.7 & 49.4 & 39.6 & 41.3 & 54.2 & 56.2
& 11.4 & 35.4
& 20.0 & 13.3 & \textbf{10.0} & 10.0 & 14.0 \\
SkillOS-base\frozen
& 79.0 & 41.0 & 45.7 & 37.5 & 38.7 & 55.6 & 53.1
& 13.6 & 38.6
& 20.0 & \textbf{20.0} & 0.0 & 20.0 & 16.0 \\
SkillOS\trainable
& 85.7 & 56.4 & 54.3 & 43.8 & 46.7 & 62.5 & 61.2
& 16.5 & 40.6
& 26.7 & \textbf{20.0} & \textbf{10.0} & 10.0 & 18.0 \\
SkillRL\trainable
& 94.3 & 84.6 & 92.6 & 87.5 & 72.0 & 75.0 & 85.0
& 73.2 & 85.5
& 20.0 & \textbf{20.0} & \textbf{10.0} & 20.0 & 18.0 \\
\midrule
\textbf{\textsc{EvoHarness-RL}}\trainable
& \seedstat{100.0}{0.0} & \seedstat{94.9}{3.6}
& \seedstat{96.3}{0.0} & \seedstat{100.0}{0.0}
& \seedstat{82.7}{1.9} & \seedstat{95.8}{0.0}
& \seedstat{95.0}{0.0}
& \seedstat{80.5}{0.7} & \seedstat{89.8}{0.9}
& \seedstat{31.1}{6.3} & \seedstat{20.0}{0.0}
& \seedstatplain{6.7}{4.7} & \seedstat{30.0}{0.0}
& \seedstat{22.7}{2.5} \\
\bottomrule
\end{tabular}
}
\end{table}

\section{Theoretical Analysis}
\label{app:theory}

This appendix analyzes what the two training stages in
Section~\ref{sec:training} can guarantee, in the order in which they act.
Supervised initialization yields a policy that can operate the BPE
interface, so Appendix~\ref{app:theory-a} first fixes such a policy and proves
Proposition~\ref{prop:conditional-efficiency}: when a BPE return carries
decision-relevant information that the policy uses with bounded error, fewer
independent rollouts suffice to discover a successful trajectory with high
probability. Reinforcement learning then updates the policy under the
scheduled reward of Equation~\eqref{eq:reward}, whose diversity bonus is
annealed. Appendix~\ref{app:theory-b} proves
Theorem~\ref{prop:annealed-diversity}: the bonus shifts the final cost-aware
objective by at most its annealed weight, so after annealing, the probability
of selecting an optimal coordination rule is limited only by the optimization
residual. Both results are conditional, and Appendix~\ref{app:theory-c}
states what they do not establish.

\subsection{Conditional Rollout Efficiency for Success Discovery}
\label{app:theory-a}

This subsection proves Proposition~\ref{prop:conditional-efficiency}, which
compares how many independent rollouts a baseline policy and a BPE policy
need to discover a successful trajectory from the same context. The proof has
three steps: BPE information raises the probability that the next search
finds its target (Step~1), a higher hit probability raises full-task success
(Step~2), and higher success reduces the required number of rollouts
(Step~3). A case from the paper first identifies the decision; we then fix the
comparison, define the quantities of each step, and prove the result.

\paragraph{The analyzed decision.}
The kettle-cleaning episode in Appendix~\ref{app:case_study} contains the
decision analyzed here. After committing to find the kettle, the agent calls
recall, which suggests a countertop, and must then choose where to search. The
suggestion is wrong, yet the agent checks the stoveburners and completes the
task. A faithful model must therefore allow a return to mislead and a search
miss to be followed by recovery; the episode illustrates the decision but
measures none of the parameters below.

\paragraph{Comparison and assumptions.}
Let $x$ be the policy-visible context at such a search decision, before the
analyzed BPE return, following the prompt notation $x_t$. We compare two fixed
checkpoints, a baseline policy $\pi_0$ and a BPE policy $\pi_1$; index $0$
refers to the baseline and index $1$ to the BPE policy throughout. Both use the
same experience bank, held fixed as within a rollout batch
(Section~\ref{sec:bpe_abstraction}), and the same remaining interaction budget
$H_x$, which BPE calls consume together with environment actions. A rollout
runs from $x$ to termination, and the comparison rests on two assumptions.
\begin{enumerate}[label=(A\arabic*),leftmargin=2.6em,itemsep=1pt,topsep=2pt]
\item \emph{Independent attempts.} Each rollout independently draws a task
state from the same conditional distribution given $x$ and samples its
actions independently; any hidden state held fixed across attempts is
included in $x$.
\item \emph{Common coupling.} Both policies are analyzed on the same joint
distribution of task state and BPE return $M$ given $x$, and reading $M$
changes neither the target state nor the legal search choices. The baseline
chooses its search action independently of $M$ given $x$, whereas the BPE
policy may condition on $M$.
\end{enumerate}
Under (A2), the same search choice has the same hit probability for both
policies, so any difference in discovery comes from how each policy uses the
return and from what happens after the search.

\paragraph{Step~1 quantities: search hits.}
Let $m$ denote a value of $M$, with $w_x(m)=\Pr(M=m\mid x)$; an empty return
covers failed retrieval or no call, and sums become integrals for continuous
returns. Let $\mathcal A_x$ be the finite nonempty set of legal search
choices. For $a\in\mathcal A_x$, $h_x(m,a)$ is the true probability that $a$
finds the target given $x$ and $m$, averaged over the remaining environment
uncertainty; it is an environment probability, not a confidence reported by
the model. With the BPE policy's choice probabilities $\pi_1(a\mid x,m)$,
define
\begin{align}
v(x,m)&=\max_{a\in\mathcal A_x}h_x(m,a), &
q(x,m)&=\sum_{a\in\mathcal A_x}\pi_1(a\mid x,m)h_x(m,a),\label{eq:theory-search-values}\\
e(x,m)&=v(x,m)-q(x,m)\in[0,1].\label{eq:theory-decision-error}
\end{align}
Here $v$ is the best hit probability achievable with the same information,
$q$ is the policy's actual hit probability, and $e$ is its decision error. The
two policies' hit probabilities are
\begin{equation}
s_0(x)=\sum_a\pi_0(a\mid x)\sum_m w_x(m)h_x(m,a),
\qquad
s_1(x)=\sum_m w_x(m)q(x,m).
\end{equation}
Their difference depends on how much the return can improve the search and on
how closely the policy realizes that improvement.

\paragraph{Information value and decision error.}
Without the return, the best hit probability is
$v_0(x)=\max_a\sum_m w_x(m)h_x(m,a)$, and the information value is
\begin{equation}
\gamma(x)=\sum_m w_x(m)v(x,m)-v_0(x)\ge0.
\label{eq:theory-information-value}
\end{equation}
It is nonnegative because a decision-maker holding the return can ignore it,
and it is positive only when some return changes the best search choice. How
closely the policy realizes this value is measured at a permitted failure
fraction $0\le\delta<1$ by the conditional quantile of its decision error,
\begin{equation}
F_x(z)=\sum_m w_x(m)\mathbf 1\{e(x,m)\le z\},\qquad
\varepsilon_\delta(x)=\inf\{z\in[0,1]:F_x(z)\ge1-\delta\},
\label{eq:theory-error-quantile}
\end{equation}
and set $b_\delta(x)=(1-\delta)\varepsilon_\delta(x)+\delta$, which Step~1 of
the proof shows to bound the mean decision error. The failure fraction
$\delta$ concerns returns, whereas the confidence $\rho$ in Step~3 concerns
rollouts. For finitely many returns, $\varepsilon_\delta(x)$ is read off by
sorting the errors and accumulating their probabilities until the mass
reaches $1-\delta$. The quantile guarantees
$\Pr(e(x,M)\le\varepsilon_\delta(x)\mid x)\ge1-\delta$ by construction, so the
substantive capability condition is that $\varepsilon_\delta(x)$ is small. This
approximate-use model follows the Bayesian view of in-context learning in
\citet{xie2022implicitbayes,zhang2025icltheory}, and experience-reuse results
with Qwen3-8B in ReMe and SkillOS motivate it empirically
\citep{cao2026reme,ouyang2026skillos}.

\paragraph{Step~2 quantities: from search hits to task success.}
A search hit matters only through its effect on completing the task. For
$i\in\{0,1\}$, let $r_{i,h}(x)$ and $r_{i,m}(x)$ be policy $i$'s full-task
completion probabilities given that its next search hits or misses, under the
same remaining budget; $r_{i,m}$ accounts for recovery after a miss, as in the
kettle episode. Policy $i$ then succeeds with probability
\begin{equation}
p_i(x)=s_i(x)r_{i,h}(x)+\bigl(1-s_i(x)\bigr)r_{i,m}(x).
\label{eq:theory-success-decomposition}
\end{equation}
Two quantities convert a hit into success:
\begin{align}
g(x)&=r_{0,h}(x)-r_{0,m}(x),\\
\kappa(x)&=s_1(x)\bigl[r_{0,h}(x)-r_{1,h}(x)\bigr]
 +\bigl(1-s_1(x)\bigr)\bigl[r_{0,m}(x)-r_{1,m}(x)\bigr],\\
c(x)&=\max\{\kappa(x),0\}.
\end{align}
The gain $g$ is the baseline's completion advantage of a hit over a miss. The
loss $c$ measures, in units of task-success probability, the budget spent on
calls and any change in subsequent execution; any specified upper bound on
$\kappa$ may replace it. The quantities of Steps~1 and~2 combine into the net
benefit and the condition of Proposition~\ref{prop:conditional-efficiency}:
\begin{equation}
\begin{gathered}
\alpha(x)=g(x)\bigl[\gamma(x)-b_\delta(x)\bigr]-c(x)>0,\\
g(x)>0,\qquad 0<s_i(x)<1,\qquad 0<p_0(x)<p_0(x)+\alpha(x)<1.
\end{gathered}
\label{eq:theory-net-gain}
\end{equation}
The condition requires that useful information, after decision error, raises
completion by more than the extra loss.

\begin{thmbox}
\begin{appendixproposition}[Conditional rollout efficiency for success discovery]
\label{prop:conditional-efficiency}
Let $N_{\mathrm{base}}(x)$ and $N_{\mathrm{BPE}}(x)$ be the minimum numbers
of independent rollouts needed to find at least one successful trajectory
conditioned on context $x$ with high probability, that is, with probability at
least $1-\rho$ for a common $0<\rho<1$. Under (A1), (A2), and
Condition~\eqref{eq:theory-net-gain}, the BPE policy needs at most a
$1/(1+\alpha(x)/p_0(x))$ fraction of the baseline's rollouts, rounded up, to
find a successful trajectory with high probability:
\begin{equation}
N_{\mathrm{BPE}}(x)
\le
\left\lceil
\frac{1}{1+\alpha(x)/p_0(x)}
\cdot N_{\mathrm{base}}(x)
\right\rceil,
\label{eq:conditional-rollout-efficiency}
\end{equation}
where $p_0(x)\in(0,1)$ is the baseline task completion probability and
$\alpha(x)>0$ is the conservative net benefit from BPE information defined
above.
\end{appendixproposition}
\end{thmbox}

\paragraph{Proof.}
Fix $x$ and suppress it in the notation.

\emph{Step~1 (search hits).} Let $\beta=\Pr(e>\varepsilon_\delta)\le\delta$.
Since $0\le e\le1$ and $\varepsilon_\delta\le1$,
$\mathbb E[e]\le(1-\beta)\varepsilon_\delta+\beta\le b_\delta$. By (A2), the
baseline's choice is independent of $M$, so
$s_0=\sum_a\pi_0(a)\sum_m w(m)h(m,a)\le v_0$. With $q=v-e$,
\begin{equation}
s_1-s_0\ge \mathbb E_M[v]-v_0-\mathbb E_M[e]
\ge\gamma-b_\delta.
\end{equation}

\emph{Step~2 (task success).} Adding and subtracting
$s_1r_{0,h}+(1-s_1)r_{0,m}$ in Equation~\eqref{eq:theory-success-decomposition}
and using $g>0$,
\begin{equation}
p_1-p_0=g(s_1-s_0)-\kappa
\ge g(\gamma-b_\delta)-c=\alpha>0.
\label{eq:theory-success-gain}
\end{equation}

\emph{Step~3 (rollout count).} By (A1), $n$ rollouts with success probability
$0<p<1$ all fail with probability $(1-p)^n$, so the minimum number that
discovers a success with probability at least $1-\rho$ is
\begin{equation}
N(p)=\left\lceil\frac{\log(1/\rho)}{f(p)}\right\rceil,
\qquad f(p)=-\log(1-p),
\end{equation}
and $N_{\mathrm{base}}=N(p_0)$. Let $k=1+\alpha/p_0$. Since $f$ is increasing
and convex with $f(0)=0$, and $kp_0=p_0+\alpha<1$,
$f(p_1)\ge f(p_0+\alpha)\ge kf(p_0)$. For
$n=\lceil N_{\mathrm{base}}/k\rceil$,
\begin{equation}
n f(p_1)\ge N_{\mathrm{base}} f(p_0)\ge\log(1/\rho),
\end{equation}
so $n$ BPE rollouts suffice, proving
Equation~\eqref{eq:conditional-rollout-efficiency}. If $p_1=1$, one rollout
suffices and the bound holds trivially.\hfill$\square$

\subsection{Conditional Convergence with Diversity Annealing}
\label{app:theory-b}

Proposition~\ref{prop:conditional-efficiency} holds the policy fixed, whereas
training updates it under a scheduled reward that adds a temporary diversity
bonus to the final objective. This subsection proves
Theorem~\ref{prop:annealed-diversity}, which bounds how far the epoch-$u$
policy is from the optimum of the final cost-aware objective and how likely it
is to select an optimal coordination rule. We define the rule class and the
two objectives, restate the theorem, prove it in two steps, and identify when
its residual vanishes.

\paragraph{Rules and objectives.}
Let $\mathcal B$ be a finite class of complete, context-dependent coordination
rules; each rule specifies harness calls together with the subsequent
execution decisions, rather than a single harness action. We hold the task
distribution, experience bank, interaction horizon, and efficiency weight
fixed, so the final objective does not depend on the epoch. For a rule $b$,
the final cost-aware objective and the diversity score are
\begin{align}
J(b)&=\mathbb E_b\left[R_{\mathrm{GRPO}}(\tau)
+\lambda_{\mathrm{eff}}R_{\mathrm{eff}}(\tau)\right],
&
d(b)&=\mathbb E_b[R_{\mathrm{div}}(\tau)]\in[0,1],
\end{align}
where $d(b)\le1$ because $R_{\mathrm{div}}$ is the fraction of the four BPE
action types used. The efficiency term is active throughout training; only
diversity is annealed, with weight $\lambda_u=\lambda_{\mathrm{div}}(u)\ge0$ at
epoch $u$. The policy at epoch $u$ induces a distribution $\mu_u$ over
$\mathcal B$, and for any distribution $\mu$ we write
$J(\mu)=\sum_b\mu(b)J(b)$ and $d(\mu)=\sum_b\mu(b)d(b)$. Training at epoch $u$
thus targets the scheduled objective $J(\mu)+\lambda_u d(\mu)$, whose
optimization residual is
\begin{equation}
\varepsilon_{\mathrm{opt}}(u)=
\max_{b\in\mathcal B}\bigl[J(b)+\lambda_u d(b)\bigr]
-\bigl[J(\mu_u)+\lambda_u d(\mu_u)\bigr]\ge0.
\label{eq:theory-opt-residual}
\end{equation}
The residual is defined at every finite epoch; it presupposes neither that the
optimizer converges nor that the neural policy can represent an optimal rule.
The final objective determines the optimal rules and their margin,
\begin{equation}
J^*=\max_{b\in\mathcal B}J(b),\qquad
\mathcal B^*=\{b:J(b)=J^*\},\qquad
\Delta_J=\min_{b\notin\mathcal B^*}\bigl[J^*-J(b)\bigr]>0,
\end{equation}
where finiteness makes $\Delta_J$ well defined unless every rule is optimal,
in which case the theorem is trivial. Since $b\sim\mu_u$ in the main text,
$\Pr(b\in\mathcal B^*)=\mu_u(\mathcal B^*)$, and the theorem reads as follows.

\begin{thmbox}
\noindent\textbf{Theorem~\ref{prop:annealed-diversity} (restated).}
\itshape For every epoch $u$,
\begin{equation}
J^*-J(\mu_u)\le\varepsilon_{\mathrm{opt}}(u)+\lambda_u
\quad\text{and}\quad
\mu_u(\mathcal B^*)\ge1-\frac{1}{\Delta_J}
\bigl(\varepsilon_{\mathrm{opt}}(u)+\lambda_u\bigr).
\tag*{(\ref{eq:objective_gap_bound}) and (\ref{eq:optimal-calling-probability})}
\end{equation}
In particular, $\lambda_u=0$ for all $u\ge U$, so if
$\varepsilon_{\mathrm{opt}}(u)\to0$, then $J(\mu_u)\to J^*$ and
$\mu_u(\mathcal B^*)\to1$.
\end{thmbox}

\paragraph{Proof.}
\emph{Step~1 (objective gap).} Choose $b^*\in\mathcal B^*$. The scheduled
objective of $\mu_u$ is within $\varepsilon_{\mathrm{opt}}(u)$ of its maximum,
and this maximum is at least the scheduled objective of $b^*$. Hence
\begin{align}
J^*+\lambda_u d(b^*)&\le
J(\mu_u)+\lambda_u d(\mu_u)+\varepsilon_{\mathrm{opt}}(u),\\
J^*-J(\mu_u)&\le\varepsilon_{\mathrm{opt}}(u)
+\lambda_u\bigl[d(\mu_u)-d(b^*)\bigr]
\le\varepsilon_{\mathrm{opt}}(u)+\lambda_u,
\end{align}
where the last inequality uses $d(\mu_u)\le1$ and $d(b^*)\ge0$. This proves
Equation~\eqref{eq:objective_gap_bound}.

\emph{Step~2 (optimal rules).} Every non-optimal rule loses at least
$\Delta_J$ in the final objective, so
\begin{equation}
\Delta_J\bigl[1-\mu_u(\mathcal B^*)\bigr]
\le\sum_b\mu_u(b)\bigl[J^*-J(b)\bigr]
=J^*-J(\mu_u).
\end{equation}
Dividing by $\Delta_J$ and applying Step~1 proves
Equation~\eqref{eq:optimal-calling-probability}. When $\lambda_u=0$ and
$\varepsilon_{\mathrm{opt}}(u)\to0$, both bounds give the limits stated in the
theorem.\hfill$\square$

\paragraph{When the residual vanishes.}
The cosine schedule in Section~\ref{sec:training} sets $\lambda_u=0$ for all
$u\ge U$, so after annealing the residual reduces to
$\varepsilon_{\mathrm{opt}}(u)=J^*-J(\mu_u)$. Its vanishing is therefore an
optimization-consistency condition on the fixed rule class,
$J(\mu_u)\to J^*$, which concerns the objective value; convergence to a
stationary point does not suffice. Exact-gradient results in finite
discounted MDPs motivate this condition:
\citet[Theorem~10]{agarwal2021policygradient} for tabular softmax policy
gradient with a full-support state distribution and a prescribed step size,
and \citet[Theorem~8]{xiao2022policygradient} for policy mirror descent under
direct parameterization with interior initialization and a constant step
size. Our training additionally regularizes toward the SFT reference and adds
an entropy bonus (Appendix~\ref{app:implementation_details}); any persistent
gap they create, together with optimization and representation error, remains
in $\varepsilon_{\mathrm{opt}}(u)$, and the finite-epoch bounds hold
regardless. After annealing, retained harness use is therefore determined only
by task reward, interaction efficiency, and the optimization residual, which is
the sense of convergence in Theorem~\ref{prop:annealed-diversity}.

\subsection{Scope of the Two Results}
\label{app:theory-c}

Proposition~\ref{prop:conditional-efficiency} and
Theorem~\ref{prop:annealed-diversity} each certify one link between the
training design and the empirical results, under the conditions stated above.
This subsection lists what they do not establish, so that the corresponding
claims in the main text and in Appendix~\ref{app:training_dynamics} are read
at their intended strength.

\paragraph{Fixed-policy discovery is not training sample complexity.}
Proposition~\ref{prop:conditional-efficiency} compares two fixed checkpoints
with a fixed bank at one context, so it does not bound the number of GRPO
steps. The speedups in Appendix~\ref{app:training_dynamics} are measured; the
proposition supplies a mechanism consistent with them, not a prediction of
their size. Its integer form has two edge cases: rounding may make both budgets equal,
and when $p_0=0$ the baseline has no finite discovery budget.

\paragraph{The gain requires decision-relevant information.}
Even at a fixed checkpoint, the condition $\alpha>0$ is an assumption about the return and the policy, not
a consequence of calling BPE. If $M$ deterministically rewrites information
already in $x$, then $\gamma=0$ and the condition cannot certify a gain; a
benefit from reorganizing the same information would instead require
comparing the two policies' processing errors. The loss $c$ is defined through
completion probabilities rather than derived from call counts. The analysis
assumes neither context truncation, perfect retrieval, nor failure after a
search miss.

\paragraph{Theorem~\ref{prop:annealed-diversity} is neither a rate nor a comparison with GRPO.}
On the training side, the theorem bounds the objective gap by the current
residual but does not show
that early diversity makes this residual shrink faster than standard GRPO, nor
that diversity leaves the optimization path unchanged; the benefit of early
exploration is supported by the reward ablations in
Appendix~\ref{app:reward_analysis}. It relies on a finite rule class with a
positive margin $\Delta_J$, and its probability refers to the current policy
selecting any optimal rule, not to a training run finding globally optimal
parameters.

\paragraph{Fixed banks and cited results.}
Both analyses hold the experience bank fixed, so neither credits a note for
improving future banks through consolidation. The cited works motivate the
modeling conditions without verifying them here: the high-probability bounds
of \citet{zhang2025icltheory} concern pretraining data rather than BPE
returns, ReMe and SkillOS do not estimate $\delta$, $\varepsilon_\delta$, or
$\alpha$, and the policy-gradient results cover exact gradients in tabular or
directly parameterized settings rather than neural GRPO.

\section{More Ablation Analysis}
\label{app:more_bpe_ablations}

\paragraph{Learned coordination beyond harness frequency.}
We further test whether the learned coordination policy captures more than the
overall amount of harness use. We evaluate the same final checkpoint with the
same frozen Experience bank and decoding protocol, changing only the schedule
of harness interactions. As shown in Table~\ref{tab:schedule_ablation}, the
learned schedule achieves the strongest overall performance. Shuffling
invocation timing causes a moderate drop, while shuffling harness-action types
leads to a substantially larger degradation. A frequency-matched control also
remains well below the learned policy despite preserving a comparable amount
of harness use. These results indicate that the gain does not come simply from
invoking the harness more often: the policy learns which external operation is
useful in the current execution state, with action selection appearing more
important than precise invocation timing on ALFWorld. The effect is strongest
on task families requiring state disambiguation or multi-stage coordination,
while simple pick-and-place tasks are largely insensitive to schedule
perturbations.

\begin{table}[H]
\centering
\begin{minipage}[t]{0.495\textwidth}
\vspace{0pt}
\centering
\scriptsize
\setlength{\tabcolsep}{3.0pt}
\renewcommand{\arraystretch}{1.08}
\captionof{table}{\small ALFWorld harness-schedule ablation under a shared
checkpoint, Experience bank, and decoding protocol. $\Delta$ is the drop from
Learned.}
\label{tab:schedule_ablation}
\resizebox{\linewidth}{!}{
\begin{tabular}{lrrrrrrr@{\hspace{4pt}}r}
\toprule
\multirow{2}{*}{\textbf{Variant}}
& \multicolumn{7}{c}{\textbf{ALFWorld}}
& \multirow{2}{*}{\deltagain{\textbf{$\Delta$}}} \\
\cmidrule(lr){2-8}
& \textbf{Pick} & \textbf{Look} & \textbf{Clean} & \textbf{Heat}
& \textbf{Cool} & \textbf{Pick2} & \textbf{Avg.} & \\
\midrule
Learned
& \cellcolor{yellow!8}100.0
& \cellcolor{yellow!8}92.3
& \cellcolor{yellow!8}96.3
& \cellcolor{yellow!8}100.0
& \cellcolor{yellow!8}84.0
& \cellcolor{yellow!8}95.8
& \cellcolor{yellow!8}\textbf{95.0}
& \cellcolor{yellow!8}\deltagain{--} \\
Timing Shuffle
& 100.0 & 100.0 & 96.3 & 87.5 & 84.0 & 91.7 & 93.6
& \deltagain{$\blacktriangledown$\,1.4} \\
Type Shuffle
& 100.0 & 61.5 & 92.6 & 100.0 & 76.0 & 87.5 & 88.6
& \deltagain{$\blacktriangledown$\,6.4} \\
Freq. Matched
& 100.0 & 76.9 & 96.3 & 93.8 & 76.0 & 91.7 & 90.7
& \deltagain{$\blacktriangledown$\,4.3} \\
\bottomrule
\end{tabular}
}
\end{minipage}
\hfill
\begin{minipage}[t]{0.495\textwidth}
\vspace{0pt}
\centering
\scriptsize
\setlength{\tabcolsep}{3.0pt}
\renewcommand{\arraystretch}{1.08}
\captionof{table}{\small ALFWorld BPE component ablation for
\textsc{EvoHarness-Base} under a shared evaluation protocol. $\Delta$ is the
drop from Full BPE.}
\label{tab:inference_ablation}
\resizebox{\linewidth}{!}{
\begin{tabular}{lrrrrrrr@{\hspace{4pt}}r}
\toprule
\multirow{2}{*}{\textbf{Variant}}
& \multicolumn{7}{c}{\textbf{ALFWorld}}
& \multirow{2}{*}{\deltagain{\textbf{$\Delta$}}} \\
\cmidrule(lr){2-8}
& \textbf{Pick} & \textbf{Look} & \textbf{Clean} & \textbf{Heat}
& \textbf{Cool} & \textbf{Pick2} & \textbf{Avg.} & \\
\midrule
Full BPE
& \cellcolor{yellow!8}71.4
& \cellcolor{yellow!8}53.8
& \cellcolor{yellow!8}63.0
& \cellcolor{yellow!8}50.0
& \cellcolor{yellow!8}48.0
& \cellcolor{yellow!8}41.7
& \cellcolor{yellow!8}56.4
& \cellcolor{yellow!8}\deltagain{--} \\
w/o Belief     & 74.3 & 53.8 & 40.7 & 75.0 & 20.0 & 37.5 & 50.0 & \deltagain{$\blacktriangledown$\,6.4} \\
w/o Progress   & 68.6 & 53.8 & 44.4 & 68.8 & 32.0 & 37.5 & 50.7 & \deltagain{$\blacktriangledown$\,5.7} \\
w/o Experience & 65.7 & 53.8 & 40.7 & 62.5 & 28.0 & 41.7 & 48.6 & \deltagain{$\blacktriangledown$\,7.8} \\
\bottomrule
\end{tabular}
}
\end{minipage}

\end{table}

\paragraph{Inference-time BPE components.}
Table~\ref{tab:inference_ablation} isolates the contribution of each component
in the prompt-time harness. Removing Belief most strongly affects tasks that
require localization and state verification, while removing Progress harms
dependent-subgoal tasks such as \textsc{Pick2}. Removing Experience yields the
lowest overall average, confirming that all three state types provide
complementary support at inference time.

\paragraph{RL internalizes harness-use patterns beyond explicit instructions.}
\begin{wraptable}{r}{0.48\textwidth}
\vspace{-8pt}
\centering
\scriptsize
\setlength{\tabcolsep}{4.0pt}
\renewcommand{\arraystretch}{1.08}
\caption{\small
Robustness to removing explicit harness-use instructions on the
140-task ALFWorld seen split.
}
\label{tab:prompt_stripping}

\resizebox{\linewidth}{!}{
\begin{tabular}{lrrr}
\toprule
\textbf{Policy}
& \textbf{Stripped SR}
& \textbf{Full-Prompt SR}
& \textbf{Drop} \\
\midrule

\textsc{EvoHarness-SFT}\trainable
& 59.29
& 68.57
& $\downarrow$ 9.29 \\

\textsc{EvoHarness-RL}\trainable
& \cellcolor{yellow!8}\textbf{92.86}
& \cellcolor{yellow!8}\textbf{95.00}
& \cellcolor{yellow!8}\textbf{$\downarrow$ 2.14} \\

\bottomrule
\end{tabular}
}
\vspace{-6pt}
\end{wraptable}
Table~\ref{tab:prompt_stripping} shows that sensitivity to explicit harness
instructions decreases substantially after RL optimization. Removing guidance
for when and how to invoke harness actions reduces \textsc{EvoHarness-SFT} from
68.57\% to 59.29\%, a drop of 9.29 percentage points. In contrast,
\textsc{EvoHarness-RL} remains largely robust, decreasing only from 95.00\% to
92.86\% ($-2.14$ points). This suggests that SFT still relies on explicit prompt
guidance to coordinate with the external harness, whereas RL increasingly
internalizes when external state should be accessed. Together with harness
annealing, where explicit harness calls decrease while success improves, this
provides complementary evidence that \methodname{} learns the coordination
strategy itself rather than simply following hand-written instructions.

\section{Qualitative Case Study}
\label{app:case_study}

We present two complementary ALFWorld cases that expose the self-evolving
harness at different timescales: policy--harness coordination within one
episode and Experience consolidation across episodes.

\paragraph{Harness-guided reasoning within an episode.}
In the first case, the agent must clean a kettle and place it on the dining
table. The trajectory shows a natural commit $\rightarrow$
recall $\rightarrow$ act $\rightarrow$ note loop. The agent
commits to a search subgoal, recalls task and object-location priors, verifies
them through interaction, and writes corrective evidence when the recalled
hint is stale. Rather than repeatedly searching the suggested countertops, it
finds the kettle on a stoveburner and records the contradiction for future
consolidation.

\definecolor{caseblue}{RGB}{28,54,98}
\definecolor{titlegray}{RGB}{245,245,245}
\definecolor{accentgold}{RGB}{165,115,25}
\definecolor{successgreen}{RGB}{0,110,35}
\definecolor{warnred}{RGB}{150,35,35}

\newcommand{\ha}[1]{\textcolor{accentgold}{\textbf{#1}}}

\begin{tcolorbox}[
    enhanced,
    breakable,
    colback=white,
    colframe=caseblue,
    boxrule=0.8pt,
    sharp corners,
    left=10pt,
    right=10pt,
    top=0pt,
    bottom=8pt,
    boxsep=0pt
]

\noindent
\colorbox{titlegray}{%
    \parbox{\dimexpr\linewidth-2\fboxsep\relax}{%
        \vspace{3pt}
        \textbf{\large Case Study: Skill Loop with Self-Correction}
        \vspace{3pt}
    }%
}

\vspace{0.8em}

\noindent
\textbf{Task:} clean some kettle and put it in diningtable.

\vspace{0.35em}

\noindent
\textbf{Type:} pick\_clean\_then\_place\_in\_recep
\hfill
\textcolor{successgreen}{\textbf{SUCCESS}}: score 1.0, 30 turns

\vspace{0.45em}
\hrule height 0.5pt
\vspace{0.35em}

\noindent
\begin{tabular}{@{}p{0.08\linewidth} p{0.86\linewidth}@{}}
\textbf{Skills}
&
\textbf{[cle\_001]} Phase-ordered plan: locate $\to$ clean $\to$ place;
\textbf{[cle\_006]} object-location priors.
\\
\textbf{Avoid}
&
\textbf{[err\_002]} Blindly trusting stale hints;
\textbf{[err\_003]} repeatedly searching the same locations.
\end{tabular}

\vspace{0.35em}
\hrule height 0.5pt
\vspace{1.0em}

\noindent
\colorbox{titlegray}{%
    \parbox{\dimexpr\linewidth-2\fboxsep\relax}{%
        \centering
        \vspace{3pt}
        \textbf{Agent Trajectory with Harness-Guided Reasoning}
        \vspace{3pt}
    }%
}

\vspace{0.7em}

\renewcommand{\arraystretch}{1.12}
\begin{tabular}{@{}p{0.10\linewidth} p{0.84\linewidth}@{}}

\textbf{Step 1}
&
\textbf{Obs:} Initial task received. \newline
\textbf{Think:} Register the first subgoal before acting; decompose the task into locate, clean, and place. \newline
\textbf{Harness:} \ha{commit} [find kettle]
\\[0.45em]
\midrule

\textbf{Step 2}
&
\textbf{Obs:} The agent needs a search strategy for the kettle. \newline
\textbf{Think:} Recall both the general clean-task procedure and likely kettle locations. \newline
\textbf{Harness:} \ha{recall} [how to do clean task] $\to$ \ha{recall} [where to find kettle] \newline
\textbf{Return:} kettle: countertop; avoid revisiting the same failed locations.
\\[0.45em]
\midrule

\textbf{Step 3}
&
\textbf{Obs:} countertop 1 empty; countertop 2 empty; stoveburner 1 has only a pan; stoveburner 2 has only a pot; stoveburner 3 contains the kettle. \newline
\textbf{Think:} The recalled countertop hint fails twice. Following \textbf{[err\_003]}, the agent avoids looping and falls back on the prior that kettles are often on stoveburners. \newline
\textbf{Action:} go to countertop 1 $\to$ go to countertop 2 $\to$ go to stoveburner 3.
\\[0.45em]
\midrule

\textbf{Step 4}
&
\textbf{Obs:} The kettle is found at stoveburner 3, contradicting the recalled countertop prior. \newline
\textbf{Think:} Use the current trajectory as corrective evidence instead of blindly trusting stale memory. \newline
\textbf{Action:} take kettle 1 from stoveburner 3. \newline
\textbf{Harness:} \ha{note} [kettle found at stoveburner 3; recalled hint said countertop].
\\[0.45em]
\midrule

\textbf{Step 5}
&
\textbf{Obs:} The kettle is held by the agent. \newline
\textbf{Think:} Continue the phase-ordered clean task: clean the kettle, then place it on the dining table. \newline
\textbf{Harness:} \ha{commit} [clean kettle with sinkbasin]. \newline
\textbf{Action:} go to sinkbasin 1 $\to$ clean kettle 1 with sinkbasin 1 $\to$ go to diningtable 1 $\to$ move kettle 1 to diningtable 1.
\\

\end{tabular}

\vspace{0.55em}
\hrule height 0.5pt
\vspace{0.45em}

\noindent
\textbf{Result:}
\textcolor{successgreen}{\textbf{SUCCESS}}.
The trajectory exhibits the full harness loop:
\ha{commit} $\to$ \ha{recall} $\to$ act $\to$ \ha{note}.
The recalled hint was wrong; the agent detected the contradiction, adapted, and wrote corrective evidence back to the skill bank.

\end{tcolorbox}

\paragraph{Experience evolution across episodes.}
The second case follows that corrective evidence beyond a single rollout. It
traces how an initially incorrect cooling procedure enters the Experience
bank, causes a failed trajectory, is corrected by the evolver, and is finally
abstracted into a reusable general workflow.

\newcommand{\evolabel}[1]{\textcolor{caseblue}{\textbf{#1}}}
\newcommand{\evoreason}[1]{{\footnotesize\color{black!65}#1}}
\newcommand{\evobad}[1]{{\small\color{warnred}\itshape #1}}
\newcommand{\evogood}[1]{{\small\color{successgreen}\itshape #1}}
\newcommand{\evoop}[1]{{\small\textcolor{accentgold}{\textbf{\textsc{#1}}}}}

\begin{tcolorbox}[
    enhanced,
    breakable,
    colback=white,
    colframe=caseblue,
    boxrule=0.8pt,
    sharp corners,
    left=10pt,
    right=10pt,
    top=0pt,
    bottom=8pt,
    boxsep=0pt
]

\noindent
\colorbox{titlegray}{%
    \parbox{\dimexpr\linewidth-2\fboxsep\relax}{%
        \vspace{3pt}
        \textbf{\large Case Study: Experience Evolution through Self-Correction}
        \vspace{3pt}
    }%
}

\vspace{0.8em}

\noindent
\textbf{Task family:} pick\_cool\_then\_place\_in\_recep
\hfill
\textbf{Entries:} \texttt{pic\_005} (task-specific), \texttt{gen\_015} (general)

\vspace{0.35em}

\noindent
{\footnotesize\color{black!65}\textbf{Bank} denotes persistent Experience
available to later rollouts; gray text summarizes the evidence and reason for
each edit.}

\vspace{0.35em}
\hrule height 0.5pt
\vspace{0.9em}

\noindent
\colorbox{titlegray}{%
    \parbox{\dimexpr\linewidth-2\fboxsep\relax}{%
        \centering
        \vspace{3pt}
        \textbf{End-to-End Skill-Evolution Trajectory}
        \vspace{3pt}
    }%
}

\vspace{0.7em}

\renewcommand{\arraystretch}{1.12}
\begin{tabular}{@{}p{0.10\linewidth} p{0.84\linewidth}@{}}

\textbf{Step 1}
&
\textbf{Cycle 1 --- Add an initial skill.} \newline
\evolabel{Bank before:} \emph{No cooling procedure.} \newline
\evoreason{Evidence contains only object-location notes; the bank lacks reusable
cooling guidance.} \newline
\evolabel{Operation:} \evoop{Add} \texttt{pic\_005}. \newline
\evolabel{Bank after:} \evobad{Take the object to a fridge, put it in, close
the fridge, then retrieve it.} \newline
\evoreason{The inferred rule is invalid but is retrieved 127 times by step 10.}
\\[0.45em]
\midrule

\textbf{Step 2}
&
\textbf{Cycle 11 --- Failure reveals the error.} \newline
\evolabel{Bank before:} \evobad{Put the object inside the fridge.} \newline
\evolabel{Rollout:} find plate $\rightarrow$ take plate $\rightarrow$ go to
fridge $\rightarrow$ \textcolor{warnred}{\textbf{put plate in fridge}}
$\rightarrow$ repeat fridge navigation $\rightarrow$
\textcolor{warnred}{\textbf{FAILURE}}. \newline
\evoreason{Insertion never triggers ALFWorld's cooling transition; the failed
trajectory is buffered as correction evidence.}
\\[0.45em]
\midrule

\textbf{Step 3}
&
\textbf{Cycle 11 --- Correct the persistent entry.} \newline
\evolabel{Bank before:} \evobad{Put it in, close the fridge, then retrieve it.}
\newline
\evolabel{Operation:} \evoop{Update} \texttt{pic\_005}. \newline
\evolabel{Bank after:} \evogood{Hold the object and issue
\texttt{cool object with fridge}; never put it inside the fridge.} \newline
\evoreason{The update preserves the entry's existing \texttt{usage\_count} of
145.}
\\[0.45em]
\midrule

\textbf{Step 4}
&
\textbf{Cycle 25 onward --- Generalize and reuse.} \newline
\evolabel{Bank before:} \emph{Correct, but task-specific \texttt{pic\_005}.}
\newline
\evoreason{The corrected mechanism applies across objects and destinations.}
\newline
\evolabel{Operation:} \evoop{Add} general entry \texttt{gen\_015}. \newline
\evolabel{Bank after:} \evogood{Take $\rightarrow$ cool while holding
$\rightarrow$ move to destination.} \newline
\evolabel{Reuse:} recall bread location $\rightarrow$ take bread $\rightarrow$
cool with fridge $\rightarrow$ move to diningtable
$\rightarrow$ \textcolor{successgreen}{\textbf{SUCCESS}}. \newline
\evoreason{The generalized entry reaches 882 retrievals by step 150.}
\\

\end{tabular}

\vspace{0.55em}
\hrule height 0.5pt
\vspace{0.45em}

\noindent
\textbf{Evolution loop:}
rollout evidence $\rightarrow$ \textsc{Add} $\rightarrow$ failure feedback
$\rightarrow$ \textsc{Update} $\rightarrow$ global abstraction
$\rightarrow$ successful reuse. The example shows that Experience evolution
changes both the content and scope of a skill rather than merely appending new
notes.

\end{tcolorbox}

\paragraph{What the cases show.}
Together, the cases reveal a closed feedback loop between policy behavior and
external Experience. Within an episode, the policy treats recalled knowledge
as a revisable prior: it grounds the suggestion against current observations
and emits note when the two disagree. Across episodes, the evolver turns
these notes and rollout outcomes into bank edits, correcting the faulty rule
while preserving its retrieval history and then lifting the corrected
task-specific procedure into a general skill. The updated skill subsequently
guides a shorter, successful trajectory. Thus, self-evolution is not merely
the accumulation of additional memory entries; it is an evidence-driven loop
in which the policy tests Experience, reports failures, and benefits from the
reconciled and abstracted Experience in later rollouts.
\section{Environment-Specific BPE Grounding}
\label{app:environment_grounding}

BPE specifies a shared functional interface, while each environment adapter
determines how the corresponding state is represented and maintained. In
\textsc{EvoHarness-Base}, the adapter updates BPE from agent-visible observations
and action outcomes and automatically renders the resulting compact workspace
at each step. The policy selects only environment actions, so access to this
always-on workspace consumes no additional interaction step. In
\textsc{EvoHarness-SFT} and \textsc{EvoHarness-RL}, the same environment-grounded
stores support adaptive coordination: BPE content is exposed or updated only
when the policy emits the corresponding meta-action, and each such call consumes
one interaction step. Belief and Progress are maintained within an episode,
whereas Experience persists and evolves across episodes. No variant exposes
privileged environment state; all policy-facing content is derived from normal
agent interaction.

\subsection{Environment-Specific State Grounding}

\paragraph{Belief (\Bstate) and \acttrack{}.}
Belief provides persistent within-episode state that would otherwise disappear
from the immediate observation context. In ALFWorld, we maintain a lightweight
world-state store from action-observation pairs, including object states,
locations, receptacle relations, and inventory information. In WebShop, the
adapter maintains shopping state derived from the user request and product
information observed during search and comparison. In WebArena, Belief is
implemented as a compact application-state representation constructed from the
agent-visible accessibility tree and interaction history. It preserves the
current page, navigation history, active controls, and task-relevant records
collected from earlier pages without accessing the raw DOM or environment
ground truth. The underlying Belief stores are updated after every environment
interaction. \textsc{EvoHarness-Base} renders their compact views automatically,
whereas in the adaptive setting \acttrack{} performs a selective read without
changing Belief:
\acttrack{}~\texttt{[object]} returns matching object states and relations in
ALFWorld, while WebShop routes \acttrack{}~\texttt{[price]},
\acttrack{}~\texttt{[candidates]}, and \acttrack{}~\texttt{[options]} to
budget, comparison, and variant-selection views. In WebArena, its arguments
select navigation history, previously observed values, objective-relevant text,
or a literal match from earlier pages. The returned slice is inserted into the
visible harness context for the next policy decision.

\paragraph{Progress (\Pstate) and \actcommit{}.}
Progress externalizes the execution state of the current task. In ALFWorld, it
is represented as a bounded subgoal record over household manipulation steps.
In WebShop, it tracks which product requirements have been resolved and which
remain to be verified before purchase. In WebArena, it maintains completed and
pending requirements across multi-page workflows. \textsc{EvoHarness-Base}
automatically maintains and displays this execution summary from observed
outcomes. In the adaptive setting, \actcommit{}~\texttt{[subgoal]} writes the
current subgoal to the bounded Progress record and returns the updated plan, so
a later decision can recover what has been completed and what remains without
reconstructing the full trajectory.

\paragraph{Experience (\Estate), \actrecall{}, and \actnote{}.}
Experience stores reusable cross-episode knowledge. We organize it into
\textbf{general experience}, which captures broadly reusable interaction
principles, and \textbf{task-specific experience}, which captures procedures
and failure patterns associated with particular task families. Individual
adapters may additionally maintain compact mistake-correction entries or
domain-specific search priors. \textsc{EvoHarness-Base} automatically retrieves
and displays relevant entries at each step. In the adaptive setting,
\actrecall{}~\texttt{[query]} instead performs this read on demand: the query
selects relevant general principles, task-specific procedures, mistakes, or
search hints and inserts them into the visible context. Conversely,
\actnote{}~\texttt{[insight]} is a write to a worker-local evidence buffer. It
records a reusable discovery or a correction to recalled Experience, but does
not modify the persistent bank during the rollout; buffered notes are validated
against outcomes and reconciled only during batched consolidation. Although the
consolidator also receives completed trajectories and the current Experience
snapshot, a reusable mismatch can be difficult to identify in a long trace: the
recalled recommendation and the later observation that confirms or contradicts
it may be separated by many unrelated interactions. At the moment of the
mismatch, the acting policy has both pieces in its visible context, so
\actnote{} provides a sparse, localized annotation of what may need to be added
or corrected and why. It is only a proposal rather than an authoritative memory
write; the consolidator checks it against trajectory outcomes and other batch
evidence before choosing \textsc{Add}, \textsc{Update}, \textsc{Archive}, or
\textsc{Skip}.

\subsection{Experience Evolution and Consolidation}

We separate online task execution from cross-episode Experience evolution.
During rollout collection, the policy acts as the \emph{solver} using a fixed
snapshot of the current Experience store. Completed trajectories, outcomes, and
note outputs are accumulated as evidence. The \emph{evolver} operates
outside the rollout loop: it abstracts reusable general and task-specific
Experience from this evidence and reconciles the resulting candidates with the
existing store through four operations: \textsc{Add}, \textsc{Update},
\textsc{Archive}, and \textsc{Skip}.

\paragraph{Consolidation trigger.}
During GRPO training, consolidation is explicitly decoupled from parallel
rollout generation. All workers use the same Experience snapshot within an
epoch, and completed episodes and notes are buffered without modifying the
persistent store. At the end of the rollout epoch, the collected evidence is
gathered and consolidation is triggered once over the complete batch. The
updated Experience store is then used in the following epoch. This keeps the
external knowledge fixed within each rollout batch and avoids introducing
additional non-stationarity into policy optimization.

\paragraph{Evolver operations.}
The evolver first abstracts accumulated trajectories into candidate Experience
at two levels. Task-specific consolidation extracts reusable procedures and
recurring failure patterns within a task family, while general consolidation
identifies principles that transfer across task families. Each candidate is
then reconciled with the current store. \textsc{Add} introduces genuinely new
Experience; \textsc{Update} refines or corrects an existing entry;
\textsc{Archive} removes stale, invalidated, or consistently unhelpful
Experience from active retrieval while retaining it for provenance; and
\textsc{Skip} ignores evidence that is redundant or overly instance-specific.
Capacity constraints and usage-based pruning further keep the active store
compact.

\begin{algorithm}[t]
\caption{Batched Experience Evolution}
\label{alg:experience_evolver}
\begin{algorithmic}[1]
\Require Experience store $\mathcal{E}^{(k)}$,
policy $\pi_\theta$, rollout batch $\mathcal{D}^{(k)}$,
workers ${w_i}_{i=1}^{W}$
\Ensure Updated Experience store $\mathcal{E}^{(k+1)}$

\State $\bar{\mathcal{E}} \gets \mathrm{Freeze}(\mathcal{E}^{(k)})$
\State $\mathcal{B} \gets \emptyset$

\ForAll{workers $w_i$ \textbf{in parallel}}
\State $\mathcal{B}_i \gets \emptyset$
\ForAll{$x \in \mathcal{D}^{(k)}*i$}
\State $(\tau, y, N) \gets
\mathrm{Rollout}(\pi*\theta, x, \bar{\mathcal{E}})$
\State $\mathcal{B}_i \gets
\mathcal{B}_i \cup {(x,\tau,y,N)}$
\EndFor
\EndFor

\State $\mathcal{B} \gets \bigcup_{i=1}^{W} \mathcal{B}*i$
\State $\mathcal{C}*{\mathrm{task}} \gets
\mathrm{AbstractTaskExperience}(\mathcal{B})$
\State $\mathcal{C}*{\mathrm{gen}} \gets
\mathrm{AbstractGeneralExperience}(\mathcal{B})$
\State $\mathcal{C} \gets
\mathcal{C}*{\mathrm{task}} \cup \mathcal{C}_{\mathrm{gen}}$

\State $\Delta \gets
\mathrm{Reconcile}(\mathcal{C}, \bar{\mathcal{E}})$
\Comment{\textsc{Add}, \textsc{Update}, \textsc{Archive}, or \textsc{Skip}}

\State $\mathcal{E}^{(k+1)} \gets
\mathrm{ApplyEdits}(\bar{\mathcal{E}}, \Delta)$
\State $\mathcal{E}^{(k+1)} \gets
\mathrm{Compact}(\mathcal{E}^{(k+1)})$

\State \Return $\mathcal{E}^{(k+1)}$
\end{algorithmic}
\end{algorithm}

Algorithm~\ref{alg:experience_evolver} summarizes the training-time procedure.
Each worker independently collects trajectory evidence using the same frozen
Experience snapshot. The worker-local buffers are gathered only after rollout
collection finishes, after which the evolver performs a single consolidation
step. Therefore, no worker observes an Experience update generated by another
worker within the same rollout epoch.

\subsection{Policy-Facing BPE Examples}

The following examples show representative mid-trajectory BPE views in the
field structure used by our environment adapters. They are not complete agent
prompts. \textsc{EvoHarness-Base} receives these compact views automatically at
each step. In \textsc{EvoHarness-SFT} and \textsc{EvoHarness-RL}, the same
content becomes visible only after the corresponding \acttrack{}, \actcommit{},
or \actrecall{} action; the boxes combine these returned views to illustrate the
external state available to subsequent decisions. Across all three environments,
Experience follows the same
two-level organization into general principles and task-specific procedures or
failure patterns. Unlike a summary of the latest observation, these views retain
facts, action outcomes, constraints, and unfinished requirements accumulated
over a long interaction.

\paragraph{ALFWorld example.}
Consider a task that requires placing two tissue boxes on a dining table. After
the agent has searched several receptacles, placed the first box, and picked up
the second, the external state can expose:

\begin{center}
\fbox{
\begin{minipage}{0.94\linewidth}
\footnotesize
\raggedright
\textbf{Belief (episode-local)}\par
Access: returned by \acttrack{}~\texttt{[tissuebox]}.\par
Last successful action: pick up \texttt{tissuebox 2} from
\texttt{coffeetable 1}.\par
Inventory: \texttt{tissuebox 2}.\par
Relations retained from earlier observations:\par
\quad \texttt{tissuebox 1 on diningtable 1}\par
\quad \texttt{tissuebox 2 held\_by agent}\par
\quad \texttt{tissuebox 3 on coffeetable 1}\par
Searched receptacles: \texttt{diningtable 1}, \texttt{coffeetable 1}.

\vspace{4pt}
\textbf{Progress (episode-local)}\par
$\checkmark$ Find and place the first tissue box on the dining table.\par
$\checkmark$ Locate and pick up the second tissue box.\par
$\square$ Place the held tissue box on \texttt{diningtable 1}; then verify that
two distinct boxes are on the target receptacle.

\vspace{4pt}
\textbf{Experience (cross-episode)}\par
\textbf{General:} Track repeated entities separately and verify each state
transition before advancing a multi-object task.\par
\textbf{Task-specific:} Tissue boxes have frequently appeared on coffee tables;
after each placement, verify the new relation before searching for the next
instance. Do not treat the first placement as completion of a two-object goal.
\end{minipage}
}
\end{center}

The immediate observation need not repeat where the first box was placed or
which receptacles were already searched. Belief preserves those facts,
Progress keeps the remaining cardinality constraint explicit, and Experience
supplies a reusable procedure without replacing instance-specific evidence. If
a recalled location prior is contradicted, \actnote{} records the correction
for later consolidation rather than overwriting the current bank online.

\paragraph{WebShop example.}
For a shopping task, the agent may inspect several result pages and product
variants before purchasing. The accumulated BPE state can render:

\begin{center}
\fbox{
\begin{minipage}{0.94\linewidth}
\footnotesize
\raggedright
\textbf{Belief (episode-local)}\par
Access: assembled from targeted \acttrack{} calls.\par
Request constraints: navy; long sleeve; button-down; size L; price $<\$40$.\par
Compared candidates retained across search pages:\par
\quad \texttt{B09A}: \$29.99, navy button-down, long sleeve, size L available.\par
\quad \texttt{B07C}: \$35.50, navy button-down, short sleeve---reject.\par
Current item: \texttt{B09A}; selected options: navy, L; displayed price:
\$31.99 after variant selection (within budget).

\vspace{4pt}
\textbf{Progress (episode-local)}\par
$\checkmark$ Match color, sleeve style, and product type.\par
$\checkmark$ Select the requested size and re-check the variant price.\par
$\square$ Verify the final option summary and availability before purchase.

\vspace{4pt}
\textbf{Experience (cross-episode)}\par
\textbf{General:} Treat every requested attribute as a hard constraint and
verify all selected options immediately before an irreversible action.\par
\textbf{Task-specific:} Search-result prices may differ from the price of the
selected color/size variant; compare the post-selection price against the
budget before \emph{Buy Now}.
\end{minipage}
}
\end{center}

Belief functions as an external comparison table over pages that are no longer
visible, while Progress prevents a plausible candidate from being purchased
before every hard constraint is checked. Experience contributes only the
transferable verification procedure, keeping it distinct from the current
product facts.

\paragraph{WebArena example.}
For a multi-page task requesting monthly successful-order counts from an admin
report, the policy may receive the following view after navigating through the
dashboard and configuring several controls:

\begin{center}
\fbox{
\begin{minipage}{0.94\linewidth}
\footnotesize
\raggedright
\textbf{Belief (episode-local)}\par
Access: rendered automatically in \textsc{EvoHarness-Base} and exposed through
BPE meta-actions in \textsc{EvoHarness-SFT/RL}.\par
Current page: Orders Report.\par
Navigation trail: Dashboard $\rightarrow$ Sales $\rightarrow$ Reports
$\rightarrow$ Orders Report.\par
Active controls: From = 01/2023; To = 05/2023; Period = Month.\par
Recent outcomes: \emph{Show Report} produced an error; repeated clicks on node
\texttt{16130} failed because that element is no longer present in the current
accessibility tree.

\vspace{4pt}
\textbf{Progress (episode-local)}\par
$\checkmark$ Reach the sales report and set the date range and monthly period.\par
$\square$ Refresh the accessibility tree and submit the configured report.\par
$\square$ Collect five monthly counts and return them in \texttt{MM:COUNT}
format.

\vspace{4pt}
\textbf{Experience (cross-episode)}\par
\textbf{General:} After a state-changing action or failed submission, re-observe
the interface rather than reusing stale action references.\par
\textbf{Task-specific:} In admin reports, replace existing filter values and
verify the rendered date and period controls before submitting; do not append
to an old value.
\end{minipage}
}
\end{center}

Here, the current page alone does not explain how the agent arrived there,
which filters were successfully set, or why the last action failed. Belief
externalizes that execution history, Progress preserves the output contract,
and Experience proposes a recovery rule grounded in the current failure. Thus,
the adapter changes the concrete representation and update logic while
preserving the same policy-facing BPE semantics.

\section{Baseline Details}
\label{app:baseline_details}

\paragraph{Implementation and reporting protocol.}
For baselines evaluated by us, we adapt the authors' official implementations
to our common environment interfaces and evaluation budgets whenever an
official implementation is available; otherwise, we reproduce the method from
the algorithm and prompts in the corresponding paper.  Because SkillOS adopts
the same evaluation setting and Qwen3-8B backbone as ours, the ALFWorld and
WebShop results for ReasoningBank, MemP, SkillOS-base, and SkillOS in
Table~\ref{tab:main_results} are transcribed from the SkillOS
paper~\citep{ouyang2026skillos}.  Results on WebArena and all remaining entries
are from our evaluations unless stated otherwise. For these transcribed rows,
we recompute ALFWorld Avg. from the reported per-category means using our
140-task category composition.

\paragraph{ReAct.}
ReAct~\citep{react} interleaves free-form reasoning traces with executable
environment actions.  After every action, the resulting observation is added
to the trajectory and conditions the next thought--action decision.  It does
not maintain cross-episode memory or update model parameters, and therefore
serves as the memory-free agent backbone for both the frontier and open-source
comparisons.

\paragraph{ExpeL.}
ExpeL~\citep{zhao2024expel} learns from a pool of self-generated experience
without gradient updates.  It first collects successful and failed trajectories
through trial, reflection, and retry; then contrasts failure--success pairs and
groups of successful runs to induce reusable natural-language insights.  The
insight store is maintained through \textsc{add}, \textsc{edit},
\textsc{upvote}, and \textsc{downvote} operations.  At evaluation time, the
agent is prompted with these global insights together with retrieved successful
trajectories similar to the current task.

\paragraph{ReasoningBank.}
ReasoningBank~\citep{ouyang2025reasoningbank} distills both successful and
failed trajectories into structured memory items containing a title, a short
description, and reusable reasoning content.  An LLM judge supplies a proxy
success signal: successes contribute validated strategies, whereas failures
contribute counterfactual lessons and pitfalls.  For a new task, embedding
similarity retrieves the most relevant items and injects them into the agent's
instructions; newly extracted items are subsequently consolidated into the
bank.  We use the ReasoningBank memory loop, not its optional memory-aware
test-time-scaling extension.

\paragraph{MemP.}
MemP~\citep{fang2026memp} models procedural memory through three stages:
building memory from completed trajectories, retrieving task-relevant memory,
and updating the store from new execution feedback.  Its proceduralization
representation combines concrete trajectories with
abstract scripts, so the agent receives both an executable example and a
general procedure.  Retrieval uses semantic task keys, while online management
can append new procedures, retain only validated successes, or revise a
retrieved procedure when it leads to failure.

\paragraph{Dynamic Cheatsheet.}
Dynamic Cheatsheet~\citep{suzgun2026dynamic} equips a frozen black-box model
with an evolving non-parametric memory.  A generator answers each query using
the current cheatsheet, and a separately prompted curator judges the generated
solution and adds, revises, merges, or removes compact strategies, solution
sketches, and code snippets.  Its retrieval-and-synthesis variant additionally
retrieves similar past input--output pairs before curation.  Thus, adaptation
occurs entirely in context and does not require labels or parameter updates.

\paragraph{ACE.}
ACE~\citep{zhang2025agentic} treats context as an evolving, structured
playbook.  A generator produces task trajectories, a reflector extracts lessons
from successes and errors, and a curator converts those lessons into localized
delta updates.  Playbook entries are itemized bullets with usefulness and
harmfulness statistics; new bullets are appended, existing ones are updated in
place, and semantic deduplication periodically removes redundancy.  This
grow-and-refine design preserves prior knowledge while avoiding repeated
monolithic prompt rewrites.

\paragraph{GRPO.}
Group Relative Policy Optimization (GRPO)~\citep{grpo} is the memory-free
training baseline.  For each task, it samples a group of trajectories and
standardizes their rewards within the group to obtain relative advantages,
thereby avoiding a separately trained critic.  The policy is optimized with a
PPO-style clipped surrogate and KL regularization against a reference policy.
Unlike the skill-based methods below, vanilla GRPO neither retrieves nor
updates an external experience store.

\paragraph{SkillOS.}
SkillOS~\citep{ouyang2026skillos} uses a frozen executor that retrieves relevant
\texttt{SKILL.md}-style files from an external SkillRepo with BM25 and a
separate curator that issues structured \textsc{insert}, \textsc{update}, or
\textsc{delete} operations after each task.  SkillOS trains this curator with
GRPO on streams of related tasks: experiences from early tasks update
SkillRepo, and performance on later related tasks measures whether those
curation decisions were useful.  Its composite reward combines downstream task
outcomes with operation-validity, skill-quality, and conciseness signals.
SkillOS-base is the corresponding variant that uses the initial curator without
RL training, thereby isolating the effect of the architecture and prompting
from learned skill curation.

\paragraph{SkillRL.}
SkillRL~\citep{xia2026skillrl} jointly improves a skill-augmented policy and its
external SkillBank.  A teacher first distills successful trajectories into
reusable strategies and failed trajectories into concise counterfactual
lessons, then organizes them into always-available general skills and
semantically retrieved task-specific skills.  Skill-augmented teacher traces
provide a cold-start SFT stage, after which GRPO optimizes the agent policy.
Following validation epochs, failure patterns trigger additions or revisions
to SkillBank, so the policy and hierarchical skill library evolve recursively.
Because access to the original teacher model is constrained by API
availability, we initialize SkillRL with the warm-up SkillBank released by its
authors and use Claude Opus 4.6 as the teacher during online skill evolution.

\section{Implementation Details}
\label{app:implementation_details}

This section records the implementation choices used on ALFWorld, WebShop,
and WebArena. We follow the shared BPE interface and cost-aware objective in
Section~\ref{sec:method}; Table~\ref{tab:implementation_hyperparams} highlights
the benchmark-specific settings that are most relevant for reproduction.

\paragraph{Model roles.}
We use Qwen3-8B and Qwen3-4B as trainable policy backbones; the headline
results use Qwen3-8B, while Qwen3-4B is included in the cross-backbone study.
Only the policy is updated by SFT and GRPO. We use Claude Opus 4.6 to collect
successful SFT trajectories. To align the comparison with SkillRL, whose
pipeline also uses a strong teacher model for skill construction and evolution,
the runs reported in Table~\ref{tab:main_results} use Claude Opus 4.6 as the
Experience consolidator. The consolidator operates only between rollout
batches, never inside an environment step, and does not replace the Qwen3-8B
policy that interacts with the environment. We further test whether our method
depends on a frontier consolidation model. In the Self-Consolidate variant of
Table~\ref{tab:skill_evolution}, Qwen3-8B itself performs consolidation without
assistance from any frontier model. This variant reaches 94.3\% on seen tasks
and 88.1\% on unseen tasks, still outperforming all baselines while remaining
close to the 95.0\% and 94.8\% obtained with Opus-based consolidation. These
results reveal that the consolidation-model gap is concentrated on unseen
tasks. We hypothesize that Qwen3-8B tends to extract more direct and homogeneous
Experience tied to observed task patterns, whereas the frontier model more
readily abstracts transferable principles. This distinction is expected because
Qwen3-8B has not been specifically trained for Experience consolidation. Yet
its strong self-consolidation performance shows that the effectiveness of our
paradigm does not come merely from outsourcing reasoning to a frontier model:
even an uncustomized 8B policy can construct a useful evolving Experience store,
while a stronger consolidator primarily improves the generality of the
knowledge it produces.

\paragraph{Action parsing and execution.}
The environment adapters parse each response into either an environment action
or one of track, commit, recall, and note.
ALFWorld and WebShop use the \texttt{<think>...</think><action>...</action>}
format, while
WebArena follows its native fenced-action format. Malformed or out-of-space
actions receive failure feedback. Fixed penalties for invalid and degenerate
repeated actions are included in the base reward
$R_{\mathrm{GRPO}}$ of Equation~\ref{eq:reward}, rather than introduced as
additional shaping terms. We disable the trainer's built-in invalid-action
penalty to avoid counting the same failure twice.

\paragraph{Harness grounding.}
All harness state is constructed from agent-visible observations and action
outcomes. ALFWorld uses a rule-based object-state and location tracker; WebShop
stores request constraints and product information; and WebArena maintains a
compact application state from the accessibility tree and interaction history.
Progress and Belief reset each episode, while Experience persists across
episodes. Each Experience category has a maximum capacity of 80 entries; this
limit applies separately to every task-family-specific category rather than to
their aggregate. Note outputs are buffered for later consolidation.
Evaluation uses two Experience protocols. For the training-based
\textsc{EvoHarness-RL}, we freeze the Experience bank saved at the selected
training checkpoint: evaluation trajectories may retrieve from this bank, but
buffered notes are discarded and no consolidation is performed. For the
frontier-model \textsc{EvoHarness-Base} results, the Experience bank starts
empty and evolves across episodes on the evaluation split through the same
batched consolidation mechanism between evaluation batches; the model
parameters remain frozen. Experience is updated only from completed preceding
evaluation episodes; no information from future episodes or unrevealed task
instances is used, consistent with the other frozen self-evolving baselines.
Thus, the former measures generalization with training-acquired Experience,
whereas the latter measures an inference-only scaffold with online Experience
evolution.

\paragraph{SFT data construction.}
For each benchmark, the teacher acts with the same BPE interface used by the
policy, and we retain only successful episodes before flattening trajectories
into next-action examples. This yields 87 trajectories and 1{,}153 next-action
conversation pairs on ALFWorld, 423 trajectories on WebShop, and 187
trajectories and 1{,}327 next-action conversation pairs on WebArena. SFT uses
LoRA with rank 64 and scaling 128. We
optimize with AdamW, a cosine schedule, 10\% warmup, and gradient clipping at
1.0. The resulting merged checkpoint initializes full-parameter GRPO.

\paragraph{GRPO training.}
GRPO samples 128 trajectories per update on every benchmark, normalizes rewards
within each prompt group, and regularizes the policy against the SFT reference.
We use PPO clipping 0.2, entropy coefficient 0.001, weight decay 0.01, and
gradient clipping 1.0. Diversity shaping is cosine-annealed over training. For
rollout construction, we follow the interactive setup of SkillRL~\citep{xia2026skillrl}
for ALFWorld and WebShop, with prompt and response token limits applied
per generation turn. For WebArena, we follow AgentGym-RL's ScalingInter
setup~\citep{xi2025agentgymrl}, which serializes multi-turn interactions into
trajectory-level sequences for optimization. All three environments
use on-policy GRPO rollouts. ALFWorld training episodes allow up to 70 steps, providing additional
interaction budget for early exploration of the newly introduced meta-actions;
WebShop training episodes allow up to 20 steps. For a consistent comparison,
all trainable WebArena methods use the AgentGym-RL setup, running for 300 updates
with an interaction-horizon curriculum of 8, 10, and 15 turns. At evaluation,
all methods share the same interaction budgets: 50 steps on ALFWorld, 20 on
WebShop, and 15 on WebArena.
Both environment actions and harness operations count toward these budgets.

\begin{table*}[t]
\centering
\footnotesize
\setlength{\tabcolsep}{4pt}
\renewcommand{\arraystretch}{1.06}
\begin{tabular}{lccc}
\toprule
\textbf{Setting} & \textbf{ALFWorld} & \textbf{WebShop} & \textbf{WebArena} \\
\midrule
\multicolumn{4}{l}{\emph{Model and SFT}} \\
Policy backbones & \multicolumn{3}{c}{Qwen3-8B and Qwen3-4B} \\
SFT teacher & \multicolumn{3}{c}{Claude Opus 4.6} \\
LoRA rank $r$ and scaling $\alpha$ & \multicolumn{3}{c}{64 and 128} \\
SFT learning rate & \multicolumn{3}{c}{$1\times10^{-5}$} \\
SFT epochs & 4 & 4 & 3 \\
\midrule
\multicolumn{4}{l}{\emph{GRPO}} \\
Group size $G$ & 8 & 8 & 4 \\
Prompts per step & 16 & 16 & 32 \\
Trajectories per update & \multicolumn{3}{c}{128} \\
PPO mini-batch & 128 & 32 & 4 \\
PPO inner epochs & 1 & 1 & 2 \\
Training updates & 150 & 150 & 300 \\
Training interaction horizon & 70 & 20 & $8\!\rightarrow\!10\!\rightarrow\!15$ \\
Actor learning rate & \multicolumn{3}{c}{$1\times10^{-6}$} \\
KL coefficient & \multicolumn{3}{c}{0.01} \\
Maximum prompt tokens & 12{,}288 & 8{,}192 & 2{,}048 \\
Maximum response tokens & 512 & 1{,}024 & 8{,}192$^\dagger$ \\
vLLM tensor parallelism & 4 & 2 & 1 \\
RL hardware & 8$\times$H200 & 8$\times$H200 & 16$\times$H200 \\
\midrule
\multicolumn{4}{l}{\emph{Harness, evaluation, and reward}} \\
Maximum evaluation steps & 50 & 20 & 15 \\
Recall top-$k_{\mathrm{rec}}$ & 3 & 6 & 3 \\
Maximum Experience entries per category & \multicolumn{3}{c}{80} \\
Success scale in $R_{\mathrm{GRPO}}$ & 10.0 & 10.0 & 1.0 \\
Efficiency weight $\lambda_{\mathrm{eff}}$ & 1.0 & 1.0 & 0.02 \\
Max diversity weight $\lambda_{\mathrm{div}}^{\max}$ & 1.0 & 1.0 & 0.02 \\
Repeated-action penalty in $R_{\mathrm{GRPO}}$ & 0.1 & 0.1 & 0.01 \\
Invalid-action penalty in $R_{\mathrm{GRPO}}$ & 0.1 & 0.1 & 0.01 \\
\bottomrule
\end{tabular}
\caption{Implementation and training hyperparameters for \methodname{} across
the three benchmarks. Shared optimizer defaults are given in the text.
$^\dagger$WebArena limits each generation turn to 512 tokens and the complete
trajectory response to 8{,}192 tokens.}
\label{tab:implementation_hyperparams}
\end{table*}

\section{Prompts}
\label{app:prompts}

We reproduce the key prompts that instantiate the BPE harness on ALFWorld,
WebShop, and WebArena. \textsc{EvoHarness-Base} receives automatically rendered
BPE views, whereas the adaptive prompts used for \textsc{EvoHarness-SFT} and
\textsc{EvoHarness-RL} expose the same four meta-actions with
environment-specific action semantics and no prescribed invocation cadence.
The consolidation prompts
operate outside the rollout loop and turn trajectory evidence into structured
Experience updates. WebArena retains its native browser-agent conversation and
action format. The prompt text is lightly abridged for readability.

\providecolor{accentgold}{RGB}{165,115,25}
\providecommand{\ha}[1]{\textcolor{accentgold}{\textbf{#1}}}
\providecommand{\lt}{\textless{}}
\providecommand{\gt}{\textgreater{}}
\providecommand{\pipe}{\textbar{}}

\definecolor{promptbluebg}{RGB}{250,252,254}   \definecolor{promptbluefr}{RGB}{175,200,225}
\definecolor{promptgreybg}{RGB}{252,252,252}   \definecolor{promptgreyfr}{RGB}{200,200,200}
\definecolor{promptyellowbg}{RGB}{254,253,248}  \definecolor{promptyellowfr}{RGB}{220,205,150}
\definecolor{promptredbg}{RGB}{254,251,251}    \definecolor{promptredfr}{RGB}{220,180,180}

\newtcolorbox{promptbox}[3]{
    enhanced, breakable,
    colback=#2, colframe=#3,
    boxrule=0.6pt, arc=2pt,
    left=8pt, right=8pt, top=6pt, bottom=8pt,
    fonttitle=\bfseries, coltitle=black,
    title={#1},
    before upper={\footnotesize\setlength{\parskip}{2pt}}
}

\subsection{Runtime Harness Prompts}
\label{app:prompt_runtime}

\paragraph{Shared structure.}
Under \textsc{EvoHarness-Base}, each environment adapter automatically appends
the current BPE view to the policy context without requiring an additional
action. The adaptive prompts used for \textsc{EvoHarness-SFT} and
\textsc{EvoHarness-RL} instead define the environment actions, expose
$\mathcal{A}_{\mathrm{bpe}}=\{\ha{commit},\ha{track},\ha{recall},\ha{note}\}$,
and specify the response syntax. They describe action semantics without
prescribing an invocation order or cadence.
Across environments, \ha{commit} externalizes the current subgoal,
\ha{track} queries episode-local Belief, \ha{recall} retrieves cross-episode
Experience, and \ha{note} records reusable evidence. The environment-specific
text changes what can be tracked and recalled: object locations in ALFWorld,
product constraints in WebShop, and previously observed pages and values in
WebArena.

\subsubsection{ALFWorld}

The ALFWorld prompt grounds the four actions in household manipulation.

\begin{promptbox}{Harness system prompt (policy-facing)}{promptbluebg}{promptbluefr}
You are an autonomous intelligent agent operating in the ALFWorld text-based household environment. The objective is task completion through interactions with objects in rooms.

You will receive: OBJECTIVE, OBSERVATION, ADMISSIBLE COMMANDS, PREVIOUS ACTION(S).

\textbf{\#\# Environment Actions} (listed in ADMISSIBLE COMMANDS)\\
- go to \lt recep\gt, take \lt obj\gt{} from \lt recep\gt, put \lt obj\gt{} in/on \lt recep\gt\\
- open/close \lt recep\gt, heat/cool/clean \lt obj\gt{} with \lt appliance\gt, use \lt tool\gt\\
- examine/look/inventory

\textbf{\#\# Harness Actions} (cognitive tools; each call costs one step)\\
\textit{\#\#\# Plan}\\
- \ha{commit} [subgoal]: Register the current subgoal. e.g. \ha{commit} [find egg]\\
\textit{\#\#\# Perception}\\
- \ha{track} [object]: Query a specific object's last location, state, and visited locations. e.g. \ha{track} [egg]\\
\textit{\#\#\# Experience}\\
- \ha{recall} [query]: Retrieve past experience. What you get depends on the query:\\
\hspace*{1em}\ha{recall} [where to find X] $\rightarrow$ search hints (object$\rightarrow$location)\\
\hspace*{1em}\ha{recall} [how to do Y task] $\rightarrow$ procedures, general skills, common mistakes\\
\hspace*{1em}\ha{recall} [mistakes to avoid] $\rightarrow$ common pitfalls and fixes\\
- \ha{note} [insight]: Record a generalizable discovery for future episodes. e.g. \ha{note} [food usually in fridge or countertop]

\textbf{\#\# Response Syntax}\\
\lt think\gt Brief reasoning in 1-2 sentences.\lt/think\gt\\
\lt action\gt your action here\lt/action\gt

Each response contains one action. Environment-action strings come from
ADMISSIBLE COMMANDS.
\end{promptbox}

At each step the policy additionally receives a per-turn user message carrying the
current context $x_t$ and the rendered harness views $h_t$ (the committed plan,
the last recalled hints, and any \ha{track}/\ha{recall} results):

\begin{promptbox}{Per-turn user template}{promptyellowbg}{promptyellowfr}
OBJECTIVE: \{task\_description\}\\
OBSERVATION: \{current\_observation\}\\
ADMISSIBLE COMMANDS: \{admissible\_commands\}\\
PREVIOUS ACTION(S): \{action\_history\}

\{harness\_views\}\hfill\textrm{\# PLAN, RECALLED HINTS, track/recall results}
\end{promptbox}

\subsubsection{WebShop}

The WebShop prompt preserves the same interface but grounds Belief in the
shopping state: price, candidate products, and selected options. The block
below is the RL policy prompt.

\begin{promptbox}{WebShop harness prompt (RL policy)}{promptbluebg}{promptbluefr}
You are an autonomous intelligent agent operating in the WebShop e-commerce
environment. The objective is to buy the product that best matches the
instruction, including its options and price ceiling.

You will receive: OBJECTIVE, OBSERVATION, AVAILABLE ACTIONS, PREVIOUS ACTION(S).

\textbf{\#\# Environment Actions} (listed in AVAILABLE ACTIONS)\\
- \texttt{search[\lt query\gt]} on the search page.\\
- \texttt{click[\lt button\gt]} for a product, option, navigation control, or
\texttt{Buy Now}.

\textbf{\#\# Harness Actions} (each call costs one step)\\
- \ha{commit} [subgoal]: register the current shopping subgoal.\\
- \ha{track} [price]: compare the current price with the budget.\\
- \ha{track} [candidates]: list previously seen products and prices.\\
- \ha{track} [options]: show selected options on the current item.\\
- \ha{recall} [query]: retrieve effective searches, purchase procedures, or
mistakes to avoid.\\
- \ha{note} [insight]: save a reusable search phrasing or a shopping pitfall.

\textbf{\#\# Response Syntax}\\
\lt think\gt Brief reasoning in 1--2 sentences.\lt/think\gt\\
\lt action\gt your action here\lt/action\gt

Each response contains one action. Environment-action strings come from
AVAILABLE ACTIONS.
\end{promptbox}

During SFT, WebShop uses the same action names and semantics.

\begin{promptbox}{WebShop per-turn user template}{promptyellowbg}{promptyellowfr}
OBJECTIVE: \{task\_description\}\\
OBSERVATION: \{current\_observation\}\\
AVAILABLE ACTIONS: \{available\_actions\}\\
PREVIOUS ACTION(S): \{action\_history\}

\{harness\_views\}
\end{promptbox}

\subsubsection{WebArena}

In \textsc{EvoHarness-Base}, WebArena automatically appends the current compact
BPE view to the native browser-agent context, and the policy emits only browser
actions. The prompt below instantiates the adaptive interface used by
\textsc{EvoHarness-SFT} and \textsc{EvoHarness-RL}. The policy observes an
accessibility tree, URL, open tabs, and prior actions and explicitly chooses
between browser and BPE meta-actions. The full conversation is retained across
rounds, and both action types are emitted in triple-backtick blocks rather than
\texttt{<action>} tags.

\begin{promptbox}{WebArena browser and cognitive-tools prompt}{promptredbg}{promptredfr}
You are an autonomous intelligent agent tasked with navigating a web browser.
You receive the user's objective, the current page's accessibility tree and
URL, the open tabs, and the previous action.

\textbf{\#\# Browser Actions}\\
- Page operations: \texttt{click [id]}, \texttt{type [id] [content] [0|1]},
\texttt{hover [id]}, \texttt{press [key]}, and \texttt{scroll [down|up]}.\\
- Navigation: \texttt{new\_tab}, \texttt{tab\_focus [i]}, \texttt{close\_tab},
\texttt{goto [url]}, \texttt{go\_back}, and \texttt{go\_forward}.\\
- Completion: \texttt{stop [answer]}.

\textbf{\#\# Cognitive Tools} (page-independent; each call consumes one interaction step)\\
- \ha{commit} [subgoal]: declare the goal you are about to pursue, not a result.\\
- \ha{track} [visited]: list previously opened pages and URLs.\\
- \ha{track} [values]: recover numbers seen on earlier pages.\\
- \ha{track} [objective]: recover earlier text related to the objective; any
other argument is matched literally against previously observed page text.\\
- \ha{recall} [how to do \lt task type\gt]: retrieve procedures and principles.\\
- \ha{recall} [mistakes to avoid]: retrieve common pitfalls.\\
- \ha{note} [insight]: save a reusable procedure, feature location, or pitfall;
its content represents reusable experience rather than the current task's
answer.

\textbf{\#\# Response Syntax}\\
A response consists of step-by-step reasoning followed by one browser or
cognitive-tool action inside a triple-backtick block. \texttt{stop [answer]}
represents task completion.
\end{promptbox}

\begin{promptbox}{WebArena conversation layout}{promptyellowbg}{promptyellowfr}
SYSTEM: You are Qwen, created by Alibaba Cloud. You are a helpful assistant.\\
USER: \{browser instruction + cognitive-tools block\}\\
ASSISTANT: Ok.\\
USER: \{round-1 observation\}\\
ASSISTANT: \{round-1 reasoning and action\}\\
\ldots\quad\textrm{the full user/assistant history is retained}
\end{promptbox}

\subsection{Experience-Store Consolidation Prompts}
\label{app:prompt_consolidation}

Experience evolution uses three prompt blocks: task-specific abstraction,
general abstraction, and reconciliation. The main experiments initialize the
Experience bank as empty; teacher trajectories supervise use of the BPE
interface while the bank used by the main RL runs remains unseeded. During RL, the bank
remains frozen within each rollout batch and is updated only between batches.
The two abstractors distill evidence from trajectories and policy-written notes,
after which the reconciler folds supported candidates into the current bank.

\paragraph{Shared system message.}
All three calls use the same system message. Environment-specific slots retain
their benchmark-specific bindings.

\begin{promptbox}{Shared consolidation system message}{promptgreybg}{promptgreyfr}
You are a helpful assistant. Reasoning may precede the FINAL output. The FINAL
output format is a single valid JSON value, with no prose or markdown fences.
\end{promptbox}

\paragraph{Task-specific abstraction.}
This prompt is applied to each task type represented in the rollout batch.

\begin{promptbox}{Task-specific Experience abstraction}{promptgreybg}{promptgreyfr}
You are the task-experience abstractor for \{agent\_description\}. You are given
the episodes produced for one \{task\_specific\_noun\}. The output is candidate
Experience distilled from this evidence alone; the store remains unchanged.

\{task\_label\}: \{task\_type\}\\
SUCCESSFUL EPISODES: \{success\_data\}\\
FAILED EPISODES: \{failure\_data\}\\
AGENT NOTES: \{notes\_text\}

Return a JSON object containing candidate skills, mistakes, and search
priorities. Each candidate contains a unique ID, a short actionable principle,
its trigger condition, and supporting evidence. Candidates are specific to
\{task\_type\}; a compact set of high-value entries is preferred, with empty
lists representing a lack of supported candidates.
\end{promptbox}

\paragraph{General abstraction.}
This prompt operates once over the pooled batch.

\begin{promptbox}{General Experience abstraction}{promptgreybg}{promptgreyfr}
You are the general-experience abstractor for \{agent\_description\}. You are
given episodes produced across several \{task\_specific\_noun\}s. The output is
candidate Experience that transfers across them; the store remains unchanged.

\{task\_label\}S PRESENT IN THIS BATCH: \{task\_type\_list\}\\
SUCCESSFUL EPISODES: \{success\_data\}\\
FAILED EPISODES: \{failure\_data\}\\
AGENT NOTES: \{notes\_text\}

Return a JSON object containing general candidate skills and mistakes. Retain a
candidate only when at least two task types support it, name those task types,
and cite the supporting trajectories or notes. Omit task-specific observations
and search priorities. Prefer a few short, high-value candidates, and return
empty lists when appropriate.
\end{promptbox}

\paragraph{Reconciliation.}
Task-specific and general candidates are reconciled with the frozen bank.

\begin{promptbox}{Experience reconciliation}{promptgreybg}{promptgreyfr}
You are the reconciler for \{agent\_description\}'s experience store. You are
given candidates distilled from the latest rollout batch and the frozen store.
The output represents incremental edits that fold supported candidates into the
store rather than restating the full store.

STAGE: \{stage\}\\
\{stage\_rules\}\\
CANDIDATE SKILLS: \{candidate\_skills\}\\
CANDIDATE MISTAKES: \{candidate\_mistakes\}\\
CANDIDATE SEARCH PRIORITIES: \{candidate\_search\_priorities\}\\
EXISTING SKILLS: \{existing\_skills\}\\
EXISTING MISTAKES: \{existing\_mistakes\}

Return a JSON object containing Experience operations and supported search
priorities. Each candidate maps to one of \textsc{Add}, \textsc{Update},
\textsc{Archive}, or \textsc{Skip}. \textsc{Skip} represents a redundant
candidate, \textsc{Update} a refinement or correction, \textsc{Archive} an
entry that is misleading beyond repair, and \textsc{Add} genuinely new
Experience. Update and archive operations reference existing IDs; evidence
references come from the provided inputs, and fields are short.
\end{promptbox}

Global reconciliation edits general Experience, while local reconciliation may
also edit the current task type and its search priorities. The implementation
may fuse abstraction and reconciliation into a single batched call, while
preserving the same \textsc{Add}, \textsc{Update}, \textsc{Archive}, and
\textsc{Skip} semantics. \textsc{Archive} removes an entry from active retrieval
while retaining it for provenance; applying edits and enforcing the capacity
limit are deterministic operations.

\end{document}